\documentclass{article}
\usepackage{iclr2027_conference,times}

\usepackage{amsmath,amsfonts,bm}

\def\eqref#1{equation~\ref{#1}}

\def\1{\bm{1}}

\DeclareMathAlphabet{\mathsfit}{\encodingdefault}{\sfdefault}{m}{sl}
\SetMathAlphabet{\mathsfit}{bold}{\encodingdefault}{\sfdefault}{bx}{n}

\usepackage{placeins}

\newcommand{\cacheQwenPointsixPerTokenKibValue}{112}

\newcommand{\weightsQwenPointsixGibValue}{1.1}

\newcommand{\cacheQwenPointsixTwohundredthousandTokensGibValue}{21}

\newcommand{\cacheOverWeightsQwenPointsixTwohundredthousandTokensValue}{19}

\newcommand{\tokensQwenPointsixSixteenGibThousandsValue}{139}

\newcommand{\cacheQwenOnesevenPerTokenKibValue}{112}

\newcommand{\weightsQwenOnesevenGibValue}{3.2}

\newcommand{\cacheQwenOnesevenTwohundredthousandTokensGibValue}{21}

\newcommand{\cacheOverWeightsQwenOnesevenTwohundredthousandTokensValue}{7}

\newcommand{\tokensQwenOnesevenSixteenGibThousandsValue}{120}

\newcommand{\cacheQwenEightBPerTokenKibValue}{144}

\newcommand{\weightsQwenEightBGibValue}{15.3}

\newcommand{\cacheQwenEightBTwohundredthousandTokensGibValue}{27}

\newcommand{\cacheOverWeightsQwenEightBTwohundredthousandTokensValue}{2}

\newcommand{\tokensQwenEightBSixteenGibThousandsValue}{5}

\newcommand{\govreportQwenEightBUncompressedValue}{16.9}

\newcommand{\govreportQwenEightBFsqThirtytwoxValue}{12.5}

\newcommand{\govreportQwenEightBFsqNinetyeightxValue}{11.4}
\newcommand{\govreportQwenEightBPaluBfortyValue}{7.0}

\newcommand{\trecQwenEightBUncompressedValue}{63.0}

\newcommand{\trecQwenEightBFsqThirtytwoxValue}{38.0}

\newcommand{\trecQwenEightBFsqNinetyeightxValue}{3.5}

\newcommand{\trecQwenEightBPaluBonehundredtwelveValue}{2.0}

\newcommand{\trecQwenEightBPaluBfortyValue}{2.0}

\newcommand{\trecQwenEightBKiviBonethousandtwentyfourValue}{58.0}

\newcommand{\compressionQwenEightBKiviBonethousandtwentyfourTrecValue}{4.7}

\newcommand{\trecQwenEightBVecinferBninehundredfourValue}{59.0}

\newcommand{\compressionQwenEightBVecinferBninehundredfourTrecValue}{5.5}

\newcommand{\narrativeqaQwenEightBUncompressedValue}{17.4}

\newcommand{\narrativeqaQwenEightBFsqThirtytwoxValue}{8.5}

\newcommand{\narrativeqaQwenEightBFsqNinetyeightxValue}{6.5}

\newcommand{\narrativeqaQwenEightBPaluBonehundredtwelveValue}{3.0}

\newcommand{\narrativeqaQwenEightBPaluBfortyValue}{2.2}

\newcommand{\narrativeqaQwenEightBPyramidkvBonethousandtwentythreeValue}{18.8}

\newcommand{\compressionQwenEightBPyramidkvBonethousandtwentythreeNarrativeqaValue}{26.6}

\newcommand{\narrativeqaQwenEightBPyramidkvBthreehundredfortyValue}{11.5}

\newcommand{\compressionQwenEightBPyramidkvBthreehundredfortyNarrativeqaValue}{197.8}

\newcommand{\rulerRelQwenPointsixVecinferBninehundredsixtyfourCtxeightkValue}{93}

\newcommand{\rulerLeadQwenPointsixFsqThirtytwoxOverPyramidkvBonehundredfourCtxeightkBootMean}{0.10}

\newcommand{\rulerLeadQwenPointsixFsqTwelvexOverPyramidkvBthreehundredtwentysixCtxeightkBootMean}{0.12}

\newcommand{\rulerRelQwenOnesevenPaluBsevenhundredeightyfourCtxeightkValue}{94}

\newcommand{\rulerRelQwenOnesevenVecinferBninehundredsixtyfourCtxeightkValue}{97}

\newcommand{\rulerRelQwenOnesevenFsqThirtytwoxCtxeightkValue}{34}

\newcommand{\rulerLeadQwenOnesevenFsqTwelvexOverPyramidkvBthreehundredtwentyfourCtxeightkBootMean}{0.14}

\newcommand{\rulerRelQwenEightBVecinferBninehundredsixtyfourCtxeightkValue}{96}

\newcommand{\rulerRelQwenEightBFsqNinetyeightxCtxeightkValue}{25}

\newcommand{\rulerQwenEightBFsqThirtytwoxCtxeightkBootMean}{0.49}

\newcommand{\rulerRelQwenEightBFsqThirtytwoxCtxeightkValue}{54}

\newcommand{\rulerLeadQwenEightBFsqThirtytwoxOverPyramidkvBonehundredfourCtxeightkBootMean}{0.25}

\newcommand{\rulerQwenPointsixFsqEightxCtxeightkBootMean}{0.56}

\newcommand{\rulerRelQwenPointsixFsqEightxCtxeightkValue}{78}

\newcommand{\rulerRelQwenPointsixFsqFourxCtxeightkValue}{79}

\newcommand{\rulerRelQwenPointsixPaluBtwohundredseventytwoCtxeightkValue}{15}

\newcommand{\rulerRelQwenPointsixTransmlaEightxCtxeightkValue}{26}

\newcommand{\rulerRelQwenPointsixPyramidkvBonethousandthirtyfourCtxeightkValue}{56}

\newcommand{\rulerRelQwenPointsixVecinferBthreehundredtwentyfourCtxeightkValue}{6}

\newcommand{\rulerRelQwenOnesevenFsqEightxCtxeightkValue}{67}

\newcommand{\rulerRelQwenOnesevenPaluBtwohundredeightCtxeightkValue}{5}

\newcommand{\rulerRelQwenOnesevenTransmlaEightxCtxeightkValue}{21}

\newcommand{\rulerRelQwenOnesevenPyramidkvBonethousandthirtytwoCtxeightkValue}{62}

\newcommand{\rulerRelQwenOnesevenVecinferBthreehundredtwentyfourCtxeightkValue}{9}

\newcommand{\rulerQwenPointsixFsqWindowBfivehundredseventyeightCtxeightkBootMean}{0.56}

\newcommand{\rulerQwenEightBFsqWindowBonehundredfortyfourCtxeightkBootMean}{0.53}

\newcommand{\reachRulerEightkThresholdValue}{0.30}

\newcommand{\reachQwenPointsixFsqPyramidkvThirtytwoxSixteenGibMillionsValue}{3.8}

\newcommand{\reachQwenPointsixFsqSixteenxSixteenGibMillionsValue}{2.21}

\newcommand{\reachFactorValue}{1.7}

\newcommand{\reachFactorLow}{1.0}

\newcommand{\reachFactorHigh}{2.3}

\newcommand{\capLeftGib}{0.74}
\newcommand{\capOne}{5.4k}
\newcommand{\capSixteen}{87k}
\newcommand{\capSixtyfour}{347k}
\newcommand{\capMaxTokensM}{1.39M}
\newcommand{\byDense}{4096}
\newcommand{\byFsqThirtytwo}{128}
\newcommand{\rungThirtytwo}{32}
\newcommand{\byPosBytes}{4}
\newcommand{\byResidualWindow}{128}
\newcommand{\vecinferSinks}{4}
\newcommand{\fsqBreakevenSix}{683}
\newcommand{\fsqBreakevenSixRatio}{12}
\newcommand{\fsqBreakevenSeventeen}{1{,}024}
\newcommand{\fsqBreakevenSeventeenRatio}{8}
\newcommand{\fsqBreakevenEight}{1{,}365}
\newcommand{\fsqBreakevenEightRatio}{6}

\newcommand{\reachTabPyramidKV}{552k}
\newcommand{\reachTabKIVI}{684k}
\newcommand{\reachTabVecInfer}{592k}
\newcommand{\reachTabPalu}{1.42M}
\newcommand{\reachTabTransMLA}{542k}

\newcommand{\gpuHoursProjectTotalValue}{5{,}620}
\newcommand{\gpuHoursProjectTrainValue}{4{,}180}
\newcommand{\gpuHoursProjectEvalValue}{1{,}030}
\newcommand{\gpuHoursProjectThroughputValue}{410}
\newcommand{\gpuHoursFinalEvalValue}{235}
\newcommand{\gpuHoursFinalEvalFailedValue}{110}
\newcommand{\gpuHoursQwenTotalValue}{1{,}867}

\newcommand{\qatFsqB}{0.177}
\newcommand{\qatFsqBsd}{0.017}

\newcommand{\qatMinmaxB}{0.151}
\newcommand{\qatMinmaxBsd}{0.021}

\newcommand{\qatPosthocB}{0.136}
\newcommand{\qatPosthocBsd}{0.014}

\newcommand{\qatRankB}{0.001}
\newcommand{\qatRankBsd}{0.001}

\newcommand{\stackEightTwoSix}{0.39}
\newcommand{\stackSixteenTwoSix}{0.22}
\newcommand{\stackSixteenFourSix}{0.22}
\newcommand{\stackThirtytwoThirtytwoSix}{0.01}
\newcommand{\pkvMatchedSixteenSix}{0.30}
\newcommand{\pkvMatchedThirtytwoSix}{0.10}
\newcommand{\stackEightTwoSeventeen}{0.39}
\newcommand{\stackEightTwoSeventeenHalf}{0.02}
\newcommand{\pkvMatchedSixteenSeventeen}{0.32}

\newcommand{\memDenseBytes}{900{,}874{,}240}
\newcommand{\memDenseMB}{900.87}
\newcommand{\memDenseFullMB}{939.52}
\newcommand{\memFsqThirtytwoBytes}{28{,}152{,}320}
\newcommand{\memFsqThirtytwoMB}{28.15}
\newcommand{\memFsqThirtytwoFullMB}{29.36}
\newcommand{\memFsqThirtytwoLayerMB}{1.005}
\newcommand{\memPromptTokens}{7{,}855}
\newcommand{\memFsqEightBytes}{112{,}609{,}280}
\newcommand{\memFsqEightMB}{112.61}
\newcommand{\memFsqEightFullMB}{117.44}

\newcommand{\tpSixThirtytwo}{1.38}

\newcommand{\tpSixFourhundred}{1.98}

\newcommand{\tpSixFourhundredKv}{5.86}

\newcommand{\tpMatchKiviSix}{15}

\newcommand{\tpSeventeenThirtytwo}{1.34}

\newcommand{\tpSeventeenFourhundred}{2.20}

\newcommand{\tpMatchPaluSeventeen}{88}

\newcommand{\tpEightThirtytwo}{1.01}

\newcommand{\tpEightThirtytwoLo}{0.98}
\newcommand{\tpEightThirtytwoHi}{1.05}
\newcommand{\tpEightNinetyseven}{1.54}

\newcommand{\tpEightFourhundred}{1.63}

\newcommand{\tpMatchPaluEight}{100}

\newcommand{\tpNativeCtxSixHi}{24{,}555}

\newcommand{\tpReachFactorLo}{6.1}

\newcommand{\tpWarmups}{2}
\newcommand{\tpReps}{5}
\newcommand{\tpSteps}{32}
\newcommand{\tpBatches}{1, 4, 16 and 64}
\newcommand{\tpContextLo}{2{,}048}
\newcommand{\tpContextHi}{150{,}000}
\newcommand{\tpContexts}{8}
\newcommand{\tpPriceLen}{8{,}192}
\newcommand{\tpLoadGate}{5.0}
\newcommand{\tpGpus}{4}
\newcommand{\tpCalibPairs}{6}
\newcommand{\tpCalibBandLo}{0.97}
\newcommand{\tpCalibBandHi}{1.03}
\newcommand{\tpSiblings}{three}
\newcommand{\tpBootstrapN}{10{,}000}
\newcommand{\tpMinBatchTokens}{100{,}000}

\newcommand{\bfMargin}{0.02}

\newcommand{\bfOutlierBound}{0.05}

\newcommand{\bfLbMargin}{2.0}

\newcommand{\bfLbOutlierBound}{8}

\newcommand{\bfLbVerdict}{equivalent in 5 of 6 (model, method) groups, different in 0, inconclusive in 1}

\newcommand{\nfGroup}{64}

\newcommand{\nfMargin}{0.05}

\newcommand{\nfLinears}{252}

\newcommand{\NFfourFSQ}{0.382}

\newcommand{\NFfourDense}{0.913}

\newcommand{\NFfourFSQbf}{0.493}

\newcommand{\NFfourDenseBf}{0.922}

\newcommand{\NFfourDeltaFSQ}{0.111}

\newcommand{\NFfourDeltaDense}{0.010}

\newcommand{\NFfourQATFSQ}{0.464}
\newcommand{\NFfourQATFSQCI}{[0.444, 0.485]}

\newcommand{\NFfourQATDelta}{0.029}

\newcommand{\NFfourQATRecovery}{74\%}

\usepackage[hypertexnames=false]{hyperref}
\hypersetup{pdftitle={Transformer-FSQ: Discrete Latents for Extreme KV-Cache Compression}, pdfauthor={Andras Szecsenyi, Thomas L. Lee, Vaisakh Shaj, Elliot J. Crowley, Amos Storkey}}
\usepackage{url}
\usepackage{graphicx}
\usepackage{float}
\usepackage{wrapfig}
\usepackage{tikz}
\usepackage{booktabs}
\usepackage{colortbl}
\usepackage{array}
\usepackage{rotating}
\usepackage{etoc}
\usepackage{capt-of}
\usetikzlibrary{shapes.geometric,arrows.meta,positioning,calc,fit,backgrounds,decorations.pathreplacing}
\newlength{\kvPlotWidth}

\colorlet{cellline}{black!35}
\colorlet{cw}{red!65!black}
\colorlet{inp}{green!55!black}
\colorlet{assigncol}{black!70}
\colorlet{fsqpt}{blue!65!black}
\colorlet{cwfill}{red!12}
\colorlet{cellfill}{black!6}

\newcommand{\tikzVqFsq}{%
\begin{tikzpicture}[x=1cm,y=1cm,
  every node/.style={font=\small},
  assign/.style={-{Stealth[length=3.2pt,width=2.6pt]},assigncol,line width=0.55pt,
                 shorten <=2.2pt,shorten >=2.6pt},
  input/.style={diamond,fill=inp,draw=white,line width=0.4pt,inner sep=0pt,minimum size=5.2pt}]

\begin{scope}
\node[font=\small\bfseries] at (1.95,3.22) {(a) Vector quantisation (VQ)};
\draw[black!70,line width=0.5pt] (0,0) rectangle (3.9,2.9);

\draw[cellline,line width=0.45pt] (0,1.43) -- (0.985,1.634);
\draw[cellline,line width=0.45pt] (0.985,1.634) -- (1.459,2.9);
\draw[cellline,line width=0.45pt] (0.985,1.634) -- (1.59,1.089);
\draw[cellline,line width=0.45pt] (1.338,0) -- (1.59,1.089);
\draw[cellline,line width=0.45pt] (1.59,1.089) -- (2.351,1.323);
\draw[cellline,line width=0.45pt] (1.646,2.9) -- (2.488,1.707);
\draw[cellline,line width=0.45pt] (2.351,1.323) -- (2.488,1.707);
\draw[cellline,line width=0.45pt] (2.351,1.323) -- (2.809,0.752);
\draw[cellline,line width=0.45pt] (2.488,1.707) -- (3.9,2.412);
\draw[cellline,line width=0.45pt] (2.78,0) -- (2.809,0.752);
\draw[cellline,line width=0.45pt] (2.809,0.752) -- (3.9,1.137);

\foreach \x/\y in {0.55/2.30,0.85/0.85,1.75/1.85,2.15/0.55,2.60/2.45,3.15/1.35,3.45/0.50}
    \fill[cw] (\x,\y) circle (1.9pt);
\draw[cw,line width=0.7pt] (3.15,1.35) circle (3.4pt);

\node[input] (za) at (2.72,1.72) {};
\draw[assign] (za.center) -- (3.15,1.35);
\node[inp,anchor=south east,inner sep=1pt] at (2.70,1.78) {$\mathbf z$};
\node[cw,anchor=west,inner sep=1.5pt] at (3.28,1.27) {$\mathbf c_{k^\star}$};

\node[anchor=east,inner sep=1pt] at (0.68,-0.72) {$\mathbf C=$};
\begin{scope}[shift={(0.72,-0.94)}]
    \fill[cellfill] (0,0) rectangle (2.75,0.44);
    \fill[cwfill]   (1.10,0) rectangle (1.65,0.44);
    \draw[black!70,line width=0.5pt] (0,0) rectangle (2.75,0.44);
    \foreach \x in {0.55,1.10,1.65,2.20} \draw[black!40,line width=0.4pt] (\x,0) -- (\x,0.44);
    \node[font=\small] at (0.275,0.22) {$\mathbf c_1$};
    \node[font=\small] at (0.825,0.22) {$\mathbf c_2$};
    \node[font=\small,cw] at (1.375,0.22) {$\mathbf c_{k^\star}$};
    \node[font=\small] at (1.925,0.22) {$\cdots$};
    \node[font=\small] at (2.475,0.22) {$\mathbf c_K$};
    \draw[cw,line width=0.8pt] (1.10,0) rectangle (1.65,0.44);
\end{scope}
\node[align=center,font=\small] at (1.95,-1.66)
    {\textbf{explicit learned codebook}\\[-1pt] stored and searched};
\end{scope}

\begin{scope}[shift={(5.9,0)}]
\node[font=\small\bfseries] at (1.95,3.22) {(b) Finite scalar quantisation (FSQ)};
\draw[black!70,line width=0.5pt] (0,0) rectangle (3.9,2.9);

\foreach \x in {1.3,2.6} \draw[cellline,line width=0.45pt] (\x,0) -- (\x,2.9);
\foreach \y in {0.9667,1.9333} \draw[cellline,line width=0.45pt] (0,\y) -- (3.9,\y);

\foreach \x in {0.65,1.95,3.25}
    \foreach \y in {0.4833,1.45,2.4167}
        \draw[fill=white,draw=fsqpt,line width=0.7pt] (\x,\y) circle (1.7pt);
\fill[fsqpt] (3.25,1.45) circle (1.7pt);

\foreach \x/\l in {0.65/-1,1.95/0,3.25/1}{
    \draw[black!70,line width=0.5pt] (\x,0) -- (\x,-0.07);
    \node[font=\small,anchor=north,inner sep=1.5pt] at (\x,-0.09) {$\l$};
}
\draw[decorate,decoration={brace,mirror,amplitude=2.5pt},black!70,line width=0.5pt]
    (0.48,-0.36) -- (3.42,-0.36);
\node[font=\small,anchor=north,inner sep=1pt] at (1.95,-0.50)
    {$\mathcal L_1=\{-1,0,1\}$:\; $L_1{=}3$ levels};
\foreach \y/\l in {0.4833/-1,1.45/0,2.4167/1}{
    \draw[black!70,line width=0.5pt] (0,\y) -- (-0.07,\y);
    \node[font=\small,anchor=east,inner sep=1.5pt] at (-0.09,\y) {$\l$};
}
\node[font=\small,anchor=east,inner sep=1.5pt] at (-0.09,2.72) {$\mathcal L_2$};

\node[input] (zb) at (2.78,1.05) {};
\draw[assign,rounded corners=1.2pt] (zb.center) -- (3.25,1.05) -- (3.25,1.45);
\node[inp,anchor=north east,inner sep=1pt] at (2.74,1.00) {$\mathbf z$};
\node[fsqpt,anchor=south,inner sep=2pt] at (3.22,1.56) {$Q(\mathbf z)$};

\node at (1.95,-0.98) {$\mathcal C_{\mathrm{FSQ}}=\mathcal L_1\times\mathcal L_2$\; ($3{\times}3{=}9$ codes)};
\node[align=center,font=\small] at (1.95,-1.66)
    {\textbf{implicit fixed codebook}\\[-1pt] not materialised or searched};
\end{scope}

\end{tikzpicture}%
}

\title{Transformer-FSQ: Discrete Latents for\\Extreme KV-Cache Compression}

\author{
  Andras Szecsenyi\textsuperscript{1}, 
  Thomas L. Lee\textsuperscript{1}, 
  Vaisakh Shaj\textsuperscript{1}, 
  Elliot J. Crowley\textsuperscript{2}, 
  Amos Storkey\textsuperscript{1} \\
  \textsuperscript{1}School of Informatics \quad 
  \textsuperscript{2}School of Engineering \\
  The University of Edinburgh \\
  \texttt{\{andras.szecsenyi, a.storkey\}@ed.ac.uk}
}

\definecolor{cFSQ}{HTML}{DD7D0B}
\definecolor{cFSQsmall}{HTML}{EE0044}
\definecolor{cMLAFSQ}{HTML}{0000DD}
\definecolor{cPalu}{HTML}{0B5588}
\definecolor{cTransMLA}{HTML}{938EE9}
\definecolor{cKIVI}{HTML}{22AA99}
\definecolor{cPyramidKV}{HTML}{993311}
\definecolor{cVecInfer}{HTML}{832FFF}
\definecolor{cTurboQuant}{HTML}{667799}
\definecolor{cUncompressed}{HTML}{4D4D4D}
\newcommand{\swatch}[1]{\textcolor{#1}{\rule[0.1ex]{1.6ex}{1.6ex}}}

\newcommand{\deepPyramidKV}{174}
\newcommand{\deepKIVI}{5.2}
\newcommand{\deepVecInfer}{11.0}
\newcommand{\deepPalu}{10.2}
\newcommand{\deepTransMLA}{14.2}
\newcommand{\deepFSQ}{410}

\iclrfinalcopy

\makeatletter
\let\iclrmakecaption\@makecaption
\renewcommand{\@makecaption}[2]{\setlength{\parfillskip}{0pt plus 0.7\textwidth}\iclrmakecaption{#1}{#2}}
\makeatother
\begin{document}
\etocdepthtag.toc{main}
\etocsettagdepth{main}{none}
\etocsettagdepth{appendix}{subsection}
\raggedbottom
\setlength{\parfillskip}{0pt plus 0.7\textwidth}
\setlength{\emergencystretch}{3em}
\setlength{\textfloatsep}{10pt plus 1.0pt minus 2.0pt}
\setlength{\floatsep}{8pt plus 1.0pt minus 2.0pt}
\setlength{\intextsep}{8pt plus 1.0pt minus 2.0pt}
\setlength{\abovecaptionskip}{4pt plus 1.0pt minus 1.0pt}
\setlength{\belowcaptionskip}{2pt plus 0.5pt minus 0.5pt}
\setlength{\parskip}{3.5pt plus 1.0pt minus 1.0pt}
\setlength{\abovedisplayskip}{5pt plus 1pt minus 2pt}
\setlength{\belowdisplayskip}{5pt plus 1pt minus 2pt}
\setlength{\abovedisplayshortskip}{2pt plus 1pt}
\setlength{\belowdisplayshortskip}{2pt plus 1pt}

\maketitle
\lhead{Preprint}

\begin{abstract}
Private, personalised language models must run on local consumer devices, where large, frontier models cannot fit, and retaining long contexts is non-trivial. Sequential memory---the key-value (KV) cache---grows linearly with context length, quickly dwarfing the model itself: for Qwen3-1.7B-Base, 200,000 retained entries require 21 GiB, 6.6 times more than the model weights. Yet the KV-cache compression methods we measured fail beyond moderate compression ratios. We introduce \emph{Transformer-FSQ}, addressing extreme KV-cache compression by replacing standard key and value projections with a finite scalar quantisation bottleneck. Our cache stores a packed vector of discrete integer codes per token, with no per-token quantisation metadata. Converting a pretrained transformer trains only these projections on 0.25 tokens per original parameter. 
We evaluate Transformer-FSQ against five baseline families on models from 0.6B to 8B parameters. On long-context retrieval (RULER), from 12$\times$ to 97.5$\times$ Transformer-FSQ is the most accurate method we evaluate, across scales. 
A fused Triton kernel decodes directly from the integer codes; with baselines given their authors' kernels where available, it decodes at $\tpEightThirtytwo\times$ (95\% interval $\tpEightThirtytwoLo$--$\tpEightThirtytwoHi$) the throughput of native SDPA attention on the uncompressed Qwen3-8B-Base at $32\times$ and $\tpEightNinetyseven\times$ at $97.5\times$.
Extreme KV-cache compression is therefore a practical way to extend how much context a fixed device can hold.
\end{abstract}

\section{Introduction}

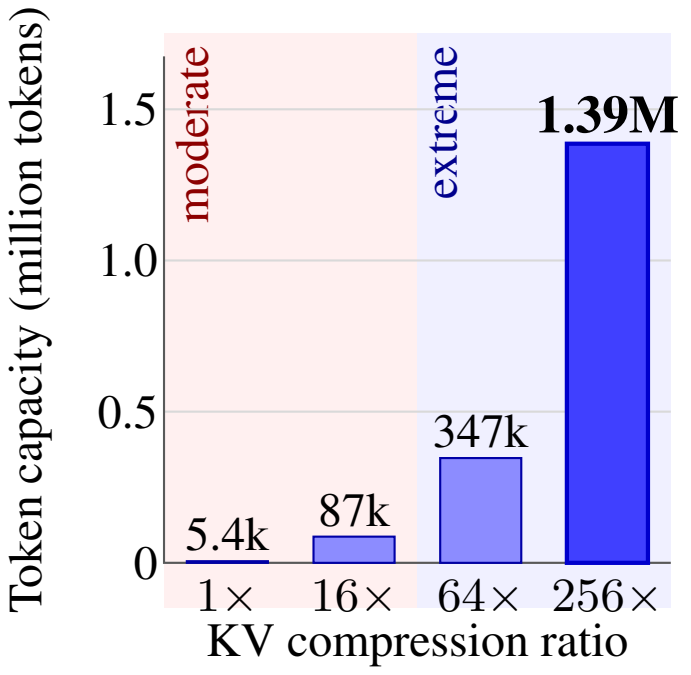
\begin{wrapfigure}{r}{0.333\textwidth}
\vspace{-1em}
  \centering
  \setlength{\kvPlotWidth}{\linewidth}
  \addtolength{\kvPlotWidth}{-13mm}

  \begin{tikzpicture}[
    x=\kvPlotWidth,
    y=1cm,
    font=\fontsize{9}{11}\selectfont,
    every node/.style={inner sep=1pt},
    bar/.style={draw=blue!65!black, fill=blue!45}
  ]
    \path[use as bounding box]
      ([xshift=-11mm]0,-0.65) rectangle
      ([xshift=2mm]1,3.60);

    \node[rotate=90, anchor=south]
      at ([xshift=-7mm]0,1.675)
      {Token capacity (million tokens)};

    \path[fill=red!6, draw=none]
      (0,-0.30) rectangle (0.5,3.50);
    \path[fill=blue!6, draw=none]
      (0.5,-0.30) rectangle (1,3.50);

    \node[
      rotate=90,
      anchor=north east,
      inner sep=0pt,
      text=red!55!black
    ] at ([xshift=2pt,yshift=-2pt]0,3.50) {moderate};

    \node[
      rotate=90,
      anchor=north east,
      inner sep=0pt,
      text=blue!55!black
    ] at ([xshift=2pt,yshift=-2pt]0.5,3.50) {extreme};

    \foreach \yy/\lab in {
      0/0,
      1/0.5,
      2/1.0,
      3/1.5
    } {
      \draw[black!15] (0,\yy) -- (1,\yy);
      \node[anchor=east] at (0,\yy) {\lab};
    }

    \draw[black!65] (0,3.35) -- (0,0) -- (1,0);

    \path[bar] (0.045,0) rectangle (0.205,0.01083);
    \path[bar] (0.295,0) rectangle (0.455,0.17326);
    \path[bar] (0.545,0) rectangle (0.705,0.69305);
    \path[
      bar,
      fill=blue!75,
      draw=blue!80!black,
      line width=0.8pt
    ] (0.795,0) rectangle (0.955,2.77222);

    \node[anchor=south] at (0.125,0.04) {\capOne};
    \node[anchor=south] at (0.375,0.20) {\capSixteen};
    \node[anchor=south] at (0.625,0.72) {\capSixtyfour};
    \node[anchor=south, font=\bfseries]
      at (0.875,2.80) {\capMaxTokensM};

    \foreach \xx/\lab in {
      0.125/1,
      0.375/16,
      0.625/64,
      0.875/256
    } {
      \node[anchor=north] at (\xx,-0.07) {$ \lab\times $};
    }

    \node[anchor=north] at (0.5,-0.36)
      {KV compression ratio};
  \end{tikzpicture}

  \begingroup
    \let\small\relax
    \let\footnotesize\relax
    \fontsize{9}{11}\selectfont
    \caption{\textbf{KV-cache capacity on a 16\,GiB GPU.} With ${\approx}\capLeftGib$\,GiB left for cache after Qwen3-8B model weights, extreme compression ($256\times$) expands token capacity from \capOne{} to \capMaxTokensM{}, enabling million-token contexts on consumer hardware.}
    \label{fig:qwen-kv-capacity}
  \endgroup
    \vspace{-0.4\intextsep}
\end{wrapfigure}

A language model's key-value (KV) cache grows with every token it processes, and on local hardware, the cache competes directly with model weights for a fixed memory budget. Unlike the model weights, whose size does not depend on context, the KV cache grows with every retained entry and soon outgrows the model itself. Local deployments matter: prompts stay on device within user control \citep{nissenbaum2010privacy,stallman2015free,zittrain2008future}, and processing spreads beyond centralised intermediaries \citep{benkler2006wealth,seger2023democratising}; releasing weights alone does not achieve this \citep{liesenfeld2024openwashing,widder2024open}. For Qwen3-8B-Base, caching $\cacheQwenEightBPerTokenKibValue$\,KiB per token in half precision (36 layers, 8 key-value heads of width 128) \citep{yang2025qwen3technicalreport} alongside $\weightsQwenEightBGibValue$\,GiB of weights on a 16\,GiB GPU leaves room for only \tokensQwenEightBSixteenGibThousandsValue{} thousand uncompressed tokens before runtime activations, reaching \capMaxTokensM{} retained entries only at $256\times$ compression (Figure~\ref{fig:qwen-kv-capacity} highlights the difference in magnitudes). Large capacity is essential: in-context learning, multi-step reasoning, and concurrent rollouts all depend on holding many tokens in memory simultaneously \citep{li2025survey}.

KV-cache compression trades accuracy for context. Existing methods shrink the cache along four axes (\S\ref{sec:related}). Token eviction keeps fewer tokens. Scalar quantisation stores each coordinate in fewer bits. Low-rank projection and latent attention store fewer coordinates per token. Vector quantisation replaces groups of coordinates with indices into a learned codebook. Four of the five baselines we compare report accuracy at compression ratios of at most about $14\times$ (Table~\ref{tab:baselines}), whereas holding several million tokens beside a model's weights on a 16\,GiB GPU needs ratios of tens for the smaller models studied here, and deeper compression for the 8B model (Figure~\ref{fig:qwen-kv-capacity}; Appendix~\ref{app:stack}), the extreme regime (above $16\times$) this paper targets.

\begin{figure}[!t]
\centering
\includegraphics{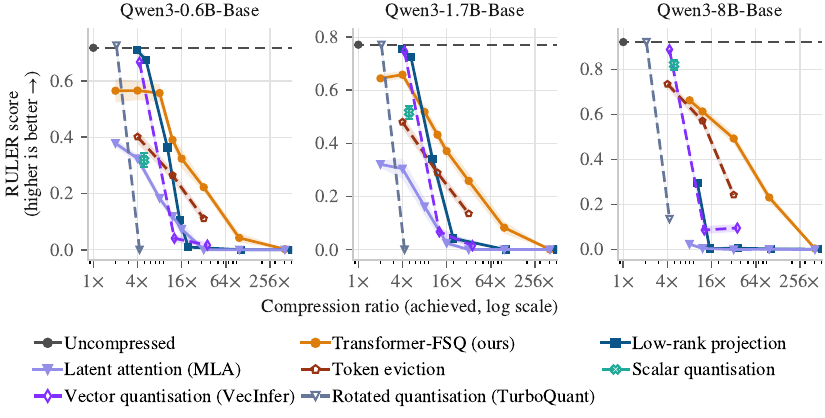}

\caption{\textbf{RULER at $8{,}192$ tokens against achieved compression} on Qwen3-0.6B, 1.7B and 8B. Solid: finetuned; dashed: training-free; grey horizontal line: uncompressed. Points average up to three seeds (one on 8B), with 95\% bootstrap intervals over seeds and then items stratified by task. Beyond about $12\times$, Transformer-FSQ scores highest on all three models.}
\label{fig:ruler}
\end{figure}

To reach extreme compression we use finite scalar quantisation (FSQ) \citep{mentzer2024finite}. FSQ projects a continuous representation into a low-dimensional space and rounds each coordinate to a fixed number of levels. Unlike learned-codebook vector quantisation \citep{oord2017neural,esser2021vqgan}, FSQ requires no stored codebook: the discrete grid is defined implicitly by the bounding levels, eliminating codebook search, commitment losses and learned codebook collapse. Transformer-FSQ replaces the key and value projections of attention with an FSQ encoder and a linear decoder, storing bit-packed integer codes in the KV cache rather than floating-point vectors. Conversion inserts these projections into a pretrained transformer in closed form, and finetuning trains only the new parameters on $0.25$ tokens per original parameter, leaving the backbone model frozen. The inserted projections are lightweight at extreme compression: at $97.5\times$ on Qwen3-1.7B-Base they hold under $10$M parameters in total, fewer than the key and value projections they replace, though at $2\times$ on Qwen3-0.6B-Base the inserted bottlenecks hold $352$M parameters and replace $59$M original key and value projection parameters, a net increase of $294$M ($\approx 49\%$ of the model), as the code approaches the residual dimension. At extreme compression, where the bottleneck is narrow, Transformer-FSQ reduces the KV cache by large factors with a small parameter overhead (\S\ref{sec:operator}).

Figure~\ref{fig:ruler} evaluates Transformer-FSQ against five baselines, one from each family in Table~\ref{tab:baselines}, plus a training-free TurboQuant reference, on RULER \citep{hsieh2024ruler} (13 official tasks, 50 items each) at $8{,}192$ tokens across three Qwen3 models. Up to about $5\times$, training-free vector quantisation and low-rank projection preserve accuracy with less overhead. At these ratios Transformer-FSQ's code approaches or exceeds the model's hidden dimension, and beyond it the closed-form initialisation repeats principal directions, so part of the cache holds duplicate information (\S\ref{sec:operator}). From $12\times$ to $97.5\times$, Transformer-FSQ retains substantially higher retrieval accuracy than every baseline, outscoring token eviction by $\rulerLeadQwenPointsixFsqTwelvexOverPyramidkvBthreehundredtwentysixCtxeightkBootMean$ and $\rulerLeadQwenOnesevenFsqTwelvexOverPyramidkvBthreehundredtwentyfourCtxeightkBootMean$ at $12\times$ on Qwen3-0.6B and 1.7B, and retaining the highest share of uncompressed accuracy through $97.5\times$. Because Transformer-FSQ compresses channel width while token eviction reduces sequence length, the two compose: on a 16\,GiB GPU, Qwen3-0.6B-Base with both yields an aggregate projected capacity of \reachQwenPointsixFsqPyramidkvThirtytwoxSixteenGibMillionsValue{}M tokens across concurrent sequences at RULER $\ge \reachRulerEightkThresholdValue$, $\reachFactorValue\times$ the best baseline, our head-shared PyramidKV (95\% interval \reachFactorLow--\reachFactorHigh). Because the cached codes are small, decoding from them is fast: a fused Triton kernel matches native SDPA decoding of the uncompressed model at $32\times$ and exceeds it from $97.5\times$ (\S\ref{sec:results}).

Our contributions are: (1) Transformer-FSQ, a discrete latent compression method that stores packed integer codes in the KV cache (\S\ref{sec:bottleneck}); (2) a closed-form conversion and finetuning operator requiring only $0.25$ tokens per original parameter on ClimbMix with the backbone frozen (\S\ref{sec:operator}); (3) an empirical evaluation showing that beyond $12\times$, Transformer-FSQ achieves the highest RULER accuracy at $8{,}192$ tokens of the methods we evaluate, stacking with eviction to expand aggregate projected capacity by $\reachFactorValue\times$ (95\% interval \reachFactorLow--\reachFactorHigh) on a 16\,GiB GPU (\S\ref{sec:experiments}); and (4) a fused Triton decode kernel that reads the packed codes directly, timed against native SDPA under strict dispatch, with KIVI, VecInfer and Palu on their authors' released kernels (\S\ref{sec:kernel}).
\section{Related work}
\label{sec:related}

Attention remains the dominant sequence mixer in large language models, despite alternatives (sparse, linear, recurrent, state-space) that alter sequence processing to reduce long-context costs. We study a complementary problem: \emph{retaining Transformer attention while reducing the state stored during inference}. In autoregressive decoding, persistent key-value caches grow linearly with context length; FlashAttention \citep{dao2022flashattention} optimises working memory during execution, but leaves this persistent cache uncompressed.

KV-cache compression mitigates this bottleneck while preserving existing model weights and architectures. Prior methods compress along four axes: token count, coordinate precision, coordinate dimension, or discrete codebooks. We review these below (extended review: Appendix~\ref{app:extended-bg}; baseline details: Appendix~\ref{app:implementation}).

\textbf{Token eviction: fewer stored tokens.}
Token eviction discards less important key-value pairs along the sequence dimension, keeping full precision for surviving tokens. StreamingLLM \citep{xiao2024streamingllm} retains attention sinks and a local sliding window; \swatch{cPyramidKV}~PyramidKV \citep{cai2024pyramidkv}, SnapKV \citep{li2024snapkv}, and H$_2$O \citep{zhang2023h2o} allocate cache budgets by attention patterns. While eviction lowers both memory and attention arithmetic, discarded token context is permanently lost.

\textbf{Scalar quantisation: fewer bits per coordinate.}
Scalar quantisation preserves key-value dimensionality while reducing numerical precision. \swatch{cKIVI}~KIVI \citep{liu2024kivi} quantises keys per channel and values per token to 2 bits alongside a full-precision window; KVQuant \citep{hooper2024kvquant} and GEAR \citep{kang2024gear} employ non-uniform grids or error-correcting residuals. \swatch{cTurboQuant}~TurboQuant \citep{zandieh2026turboquant} rotates activations before scalar quantisation; it appears in Figure~\ref{fig:ruler} evaluated training-free at moderate compression, without recovery finetuning or extreme-budget adaptation (Appendix~\ref{app:implementation}).

\textbf{Latent projection: fewer coordinates per token.}
Latent projection reduces key-value dimensionality into narrower continuous vectors. Multi-Head Latent Attention (MLA) \citep{deepseek2024v2} replaces per-head storage with a shared continuous latent and decoupled rotary key; \swatch{cTransMLA}~TransMLA \citep{meng2025transmla} adapts models to MLA via finetuning, while \swatch{cPalu}~Palu \citep{chang2025palu} factorises key-value projections via SVD. These latents remain floating-point vectors unless subsequently quantised.

\textbf{Vector and discrete quantisation: discrete representations per token.}
Vector quantisation replaces continuous representations with discrete codes. VQ-VAE \citep{oord2017neural} maps inputs to their nearest codebook entries; product quantisation codes subspaces independently \citep{jegou2011product}, and residual quantisation codes successive error stages \citep{lee2022rqvae,zeghidour2021soundstream}. For KV caches, coupled quantisation \citep{zhang2024coupled} and CommVQ \citep{li2025commvq} learn codebooks; \swatch{cVecInfer}~VecInfer \citep{yao2026vecinfer} suppresses outliers before VQ; and PQCache \citep{zhang2025pqcache}, \citet{kumar2024rvqkv}, VQKV \citep{wang2026vqkv}, and GSRQ \citep{kim2026gsrq} use product or residual variants. While VQ enables discrete representations, codebook memory, nearest-neighbour search latency, and codebook collapse during training constrain its scalability under extreme compression.

Transformer-FSQ combines dimensionality reduction with a discrete representation: it maps the residual stream to a compact latent, rounding coordinates onto a finite grid. Unlike conventional VQ, this Cartesian grid is implicit and unlearned, eliminating stored codebooks and nearest-neighbour search.

\section{Discrete latents for KV-cache compression}
\label{sec:background_quantization}

Our goal is extreme KV-cache compression, where reducing precision or width alone is not enough. We seek a representation reducing both coordinates and bits per coordinate without auxiliary tables: a wide, low-bit latent narrower than the full KV cache but wider than an equal-byte floating-point vector. We examine VQ and FSQ as two routes to this discrete bottleneck.

\begin{figure}[t]
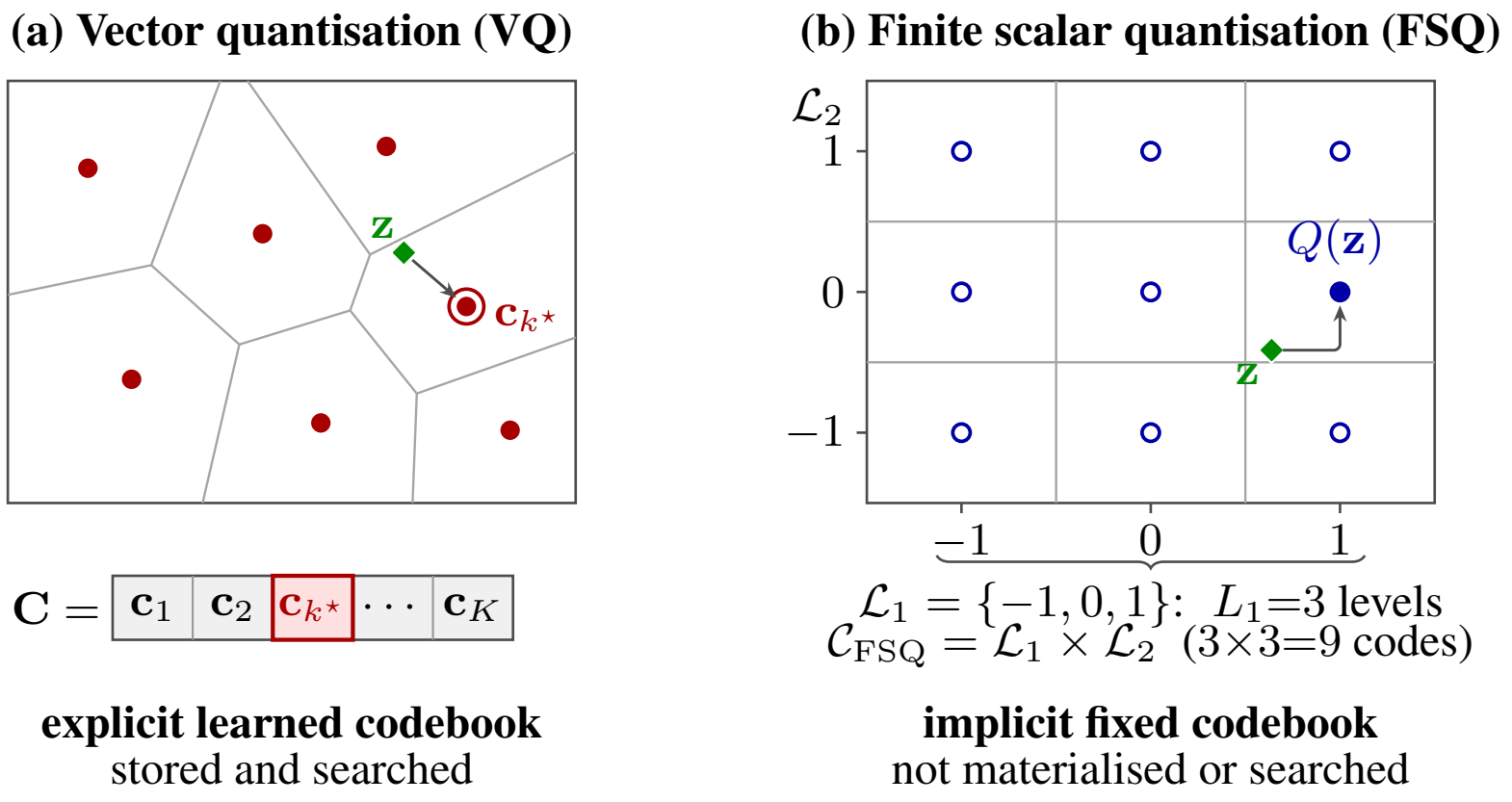

\centering
\tikzVqFsq
\caption{\textbf{Discrete latent geometry:} \textbf{(a)}~VQ searches an explicit codebook, inducing irregular Voronoi cells \citep{gray1998quantization}. \textbf{(b)}~FSQ quantises coordinates independently onto a fixed Cartesian grid $\mathcal L_1\times\mathcal L_2$, an implicit codebook that is never materialised or searched.}
\label{fig:vq-fsq}
\end{figure}

\textbf{Vector quantisation.} Discrete representation learning maps a continuous latent vector to a finite set of states. In vector quantisation (VQ), states are stored explicitly as codebook $\mathcal C=\{\mathbf c_1,\ldots,\mathbf c_K\}$, with $\mathbf c_k\in\mathbb R^d$, mapping input $\mathbf z\in\mathbb R^d$ to its nearest codeword:
\begin{equation}
Q_{\mathrm{VQ}}(\mathbf z)=\mathbf c_{k^\star},\quad
k^\star=\arg\min_k\|\mathbf z-\mathbf c_k\|_2^2.
\end{equation}
Nearest-neighbour assignment partitions latent space into Voronoi cells \citep{gray1998quantization}. However, learned codebooks require commitment losses, updates, and nearest-neighbour search (shown on Figure~\ref{fig:vq-fsq}a), risking centroid collapse where entries become inactive \citep{oord2017neural,mentzer2024finite}.

\textbf{Finite scalar quantisation.} FSQ replaces the learned vector codebook with fixed scalar levels in each latent coordinate \citep{mentzer2024finite}. Let $\mathcal L_i$ denote the finite set of allowed scalar values for coordinate $i$, and let $L_i=|\mathcal L_i|$ be the number of levels in that coordinate; for instance, in Figure~\ref{fig:vq-fsq}b every coordinate may take one of the $L_i=3$ values $\mathcal L_i=\{-1,0,1\}$. FSQ first bounds each encoder output and quantises coordinates independently: a coordinate $u$ is bounded and rounded as
\begin{equation}\label{eq:fsq-exact}
\tilde u \;=\; \tanh\!\big(u + \operatorname{atanh}(\delta/h)\big)\, h \;-\; \delta,
\qquad
\hat u \;=\; Q(u) \;=\; \tilde u + \operatorname{sg}\!\big(\lfloor \tilde u \rceil - \tilde u\big)\in\mathcal L_i,
\end{equation}
where $\lfloor \cdot \rceil$ denotes round-to-nearest with round-half-to-even tie breaking, the stop-gradient $\operatorname{sg}$ implements the straight-through estimator \citep{bengio2013estimating}, and $h$ and $\delta$ place the $L$ levels on the integers: bounding scale $\varepsilon = 10^{-3}$, half-width $h = (L-1)(1+\varepsilon)/2 = 7.5075$ for $L=16$, and centre shift $\delta = 0.5$ for even $L$ ($\delta = 0$ for odd $L$). Packing stores the unsigned integer $\lfloor L/2 \rfloor$ plus the rounded signed level ($0$ to $L-1$ for $L=16$, i.e.\ $-8,\ldots,7 \mapsto 0,\ldots,15$) in $b = \lceil \log_2 L \rceil$ bits; the decoder subtracts the same offset before applying $W_k$/$W_v$. The forward pass returns the rounded value; the stop-gradient makes the backward pass treat rounding as the identity, so gradients reach the encoder as if the rounding were absent. Rounding is applied in every training step, so the encoder learns representations that survive it.
The complete set of representable vectors is therefore the Cartesian product
\begin{equation}
\mathcal C_{\mathrm{FSQ}}
=\mathcal L_1\times\cdots\times\mathcal L_d,
\quad
|\mathcal C_{\mathrm{FSQ}}|=\prod_{i=1}^{d}L_i.
\end{equation}
When all coordinates use $L_i=L$ levels, the implicit code space contains $L^d$ vectors ($3^2=9$ grid points in Figure~\ref{fig:vq-fsq}b); in practice the $L_i$ may differ across coordinates to meet a target code-space size \citep{mentzer2024finite}. This space is combinatorially large but is never materialised: encoding requires only coordinate-wise rounding, with no centroid table or nearest-neighbour search. Because codewords are fixed to a regular Cartesian grid rather than learned vectors, FSQ avoids commitment losses, codebook updates, and learned codebook collapse \citep{mentzer2024finite}, while code utilisation across individual coordinates is empirically evaluated in Appendix~\ref{app:entropy}.

\paragraph{FSQ latents for KV-cache compression.} These properties make FSQ well suited to transformer KV-cache compression: (i)~\emph{Simplicity}: stateless coordinate-wise rounding requires no learned codebook or commitment objectives \citep{mentzer2024finite}; (ii)~\emph{Implicit codebook}: the code space ($L^d$) is never learned, so it cannot collapse; (iii)~\emph{Zero metadata}: fixed integer coordinates cost $\lceil\log_2 L_i\rceil$ bits each, with no per-token or per-channel scale overhead; and (iv)~\emph{Cheap decoding}: dequantisation is an integer offset with no table lookup, so a kernel can read the packed codes directly. Section~\ref{sec:codedkv} integrates this bottleneck into the attention pathway.

\section{Transformer-FSQ}
\label{sec:codedkv}\label{sec:method}

Now we retrofit a pretrained transformer with an extremely compact per-token state that can replace the conventional KV cache, without pretraining the model from scratch. This setting differs from both conventional discrete tokenisation and existing KV-cache quantisation. FSQ is typically used to map a continuous signal to a small number of quantised coordinates whose Cartesian product acts as a discrete token, while KV-cache quantisation usually operates after keys and values have already been formed. Transformer-FSQ instead makes two design choices tailored to the internal state of a pretrained language model.

\paragraph{Wide product codes.}
A cached language-model state must preserve enough information to support many future, query-dependent KV readouts. We therefore use FSQ in a different regime from conventional image and audio tokenisers~\citep{li2026discrete}: as a \emph{wide product representation} of many low-precision coordinates rather than a single categorical token. Under a fixed byte budget, lowering the precision of each coordinate buys proportionally more latent dimensions (RQ3 in \S\ref{sec:results} tests this trade-off), while FSQ can scale to this width without materialising or searching the exponentially large Cartesian code space. Thus $L^G$ should be understood as the number of representable product codes, not as a categorical vocabulary.

\begin{figure}[t]
\centering
\tikzset{
  ckvbox/.style={rectangle, draw=black!60, fill=black!5, rounded corners=2pt, font=\small, minimum height=0.6cm, inner sep=2pt, text centered},
  ckvfsq/.style={rectangle, draw=cFSQ!90, fill=cFSQ!18, rounded corners=2pt, font=\small, minimum height=0.6cm, inner sep=2pt, text centered},
  ckvcache/.style={rectangle, draw=cFSQ!95, fill=cFSQ!25, rounded corners=2pt, font=\small, minimum height=0.6cm, inner sep=2pt, text centered},
  ckvstdcache/.style={rectangle, draw=black!60, fill=black!10, rounded corners=2pt, font=\small, minimum height=0.6cm, inner sep=2pt, text centered},
  ckvarr/.style={->, >=latex, draw=black!75, line width=0.8pt},
  ckvop/.style={circle, draw=black!60, fill=white, inner sep=0pt, minimum size=4.5mm, font=\small},
  ckvlab/.style={font=\small, inner sep=1pt},
  ckvplate/.style={draw=black!25, fill=black!4, rounded corners=4pt, inner sep=4pt},
}
\begin{minipage}[b]{0.43\linewidth}
\centering
{\small\textbf{Standard attention}}\\[0.2cm]
\begin{tikzpicture}[y=0.86cm, box/.style={ckvbox}, cachebox/.style={ckvstdcache}, arr/.style={ckvarr},
                    op/.style={ckvop}, lab/.style={ckvlab}]
  \node[box, minimum width=1.2cm] (x) at (0,0) {$x_t$};
  \node[box, minimum width=1.2cm] (qp) at (-2.2,1.2) {Q proj};
  \node[box, minimum width=1.2cm] (kp) at (-0.7,1.2) {K proj};
  \node[box, minimum width=1.2cm] (vp) at (0.8,1.2) {V proj};
  \node[box, minimum width=1.2cm] (qr) at (-2.2,2.4) {RoPE};
  \node[box, minimum width=1.2cm] (kr) at (-0.7,2.4) {RoPE};
  \node[cachebox, minimum width=2.9cm] (cache) at (0,3.7) {K and V cache (16-bit)};
  \node[box, minimum width=2.9cm] (attn) at (0,5.2) {attention};
  \node[box, minimum width=1.6cm] (op) at (0,6.3) {Out proj};
  \node[op] (plus) at (0,7.2) {$+$};
  \node[box, minimum width=1.2cm] (out) at (0,8.1) {$x_{t+1}$};
  \coordinate (fork) at (0,0.65);
  \draw[arr] (x.north) -- (fork) -| (kp.south);
  \draw[arr] (fork) -| (vp.south);
  \draw[arr] (fork) -| (qp.south);
  \draw[arr] (qp) -- (qr);
  \draw[arr] (kp) -- (kr);
  \draw[arr] (kr.north) -- (kr.north |- cache.south);
  \draw[arr] (vp.north) -- (vp.north |- cache.south);
  \draw[arr] (cache) -- (attn);
  \draw[arr] (qr.north) |- (attn.west);
  \draw[arr] (attn) -- (op);
  \draw[arr] (op) -- (plus);
  \draw[arr] (plus) -- (out);
  \draw[arr] (x.east) -- (1.9,0) |- (plus.east);
  \begin{pgfonlayer}{background}
    \node[ckvplate, fit=(attn)] (pstd) {};
  \end{pgfonlayer}
  \node[font=\small\sffamily\bfseries, text=black!70, anchor=south east, inner sep=1pt] at (pstd.north east) {${\times}T$};
\end{tikzpicture}
\end{minipage}\hfill
\begin{minipage}[b]{0.55\linewidth}
\centering
{\small\textbf{Transformer-FSQ}}\\[0.2cm]
\begin{tikzpicture}[y=0.86cm, box/.style={ckvbox}, fsqbox/.style={ckvfsq}, cachebox/.style={ckvcache}, arr/.style={ckvarr},
                    op/.style={ckvop}, lab/.style={ckvlab}]
  \node[box, minimum width=1.2cm] (x) at (0,0) {$x_t$};
  \node[fsqbox, minimum width=2.2cm] (enc) at (0,1.1) {FSQ encoder};
  \node[cachebox, minimum width=2.2cm] (code) at (0,2.2) {code cache (packed)};
  \node[fsqbox, minimum width=2.2cm] (wk) at (-0.9,3.7) {$W_k z_t$ (tensor cores)};
  \node[box, minimum width=2.2cm] (nr) at (-0.9,4.95) {key norm, RoPE};
  \node[box, minimum width=3.1cm] (attn) at (-0.3,6.15) {scores, softmax, $\sum_t a_t z_t$};
  \node[fsqbox, minimum width=2.2cm] (mg) at (-0.3,7.6) {merge, $W_v$};
  \node[box, minimum width=1.6cm] (op) at (-0.3,8.65) {Out proj};
  \node[op] (plus) at (-0.3,9.45) {$+$};
  \node[box, minimum width=1.2cm] (out) at (-0.3,10.3) {$x_{t+1}$};
  \node[box, minimum width=1.2cm] (qp) at (-3.0,3.7) {Q proj};
  \node[box, minimum width=1.2cm] (qr) at (-3.0,4.95) {RoPE};
  \coordinate (fk) at (0,2.95);
  \draw[arr] (x) -- (enc);
  \draw[arr] (enc) -- (code);
  \draw[arr] (code.north) -- (fk);
  \draw[arr] (fk) -| (wk.south);
  \draw[arr] (wk) -- (nr);
  \draw[arr] (nr.north) -- (nr.north |- attn.south);
  \draw[arr] (fk) -| node[lab, pos=0.75, right]{$z_t$} ($(attn.south)+(0.9,0)$);
  \draw[arr] (x.west) -- (-3.0,0) -- (qp.south);
  \draw[arr] (qp) -- (qr);
  \draw[arr] (qr.north) |- (attn.west);
  \draw[arr] (attn) -- node[lab, left]{split partials} (mg);
  \draw[arr] (mg) -- (op);
  \draw[arr] (op) -- (plus);
  \draw[arr] (plus) -- (out);
  \draw[arr] (x.east) -- (1.7,0) |- (plus.east);
  \begin{pgfonlayer}{background}
    \node[ckvplate, fit=(wk)(nr)(attn)] (pfsq) {};
  \end{pgfonlayer}
  \node[font=\small\sffamily\bfseries, text=black!70, anchor=south east, inner sep=1pt] at (pfsq.north east) {${\times}T$};
\end{tikzpicture}
\end{minipage}
\caption{\textbf{One decoding step.} Left: standard attention caches keys and values. Right: Transformer-FSQ caches one bit-packed code $z_t$ per token, shared by all KV heads. Keys are reconstructed and rotated at every cached position (${\times}T$); values are decoded once, after attention pools the codes, $\bar z=\sum_t a_t z_t$. Orange: added by Transformer-FSQ; grey: frozen backbone. Details: Appendix~\ref{app:diagrams}.}
\label{fig:fsq-decode-main}
\end{figure}
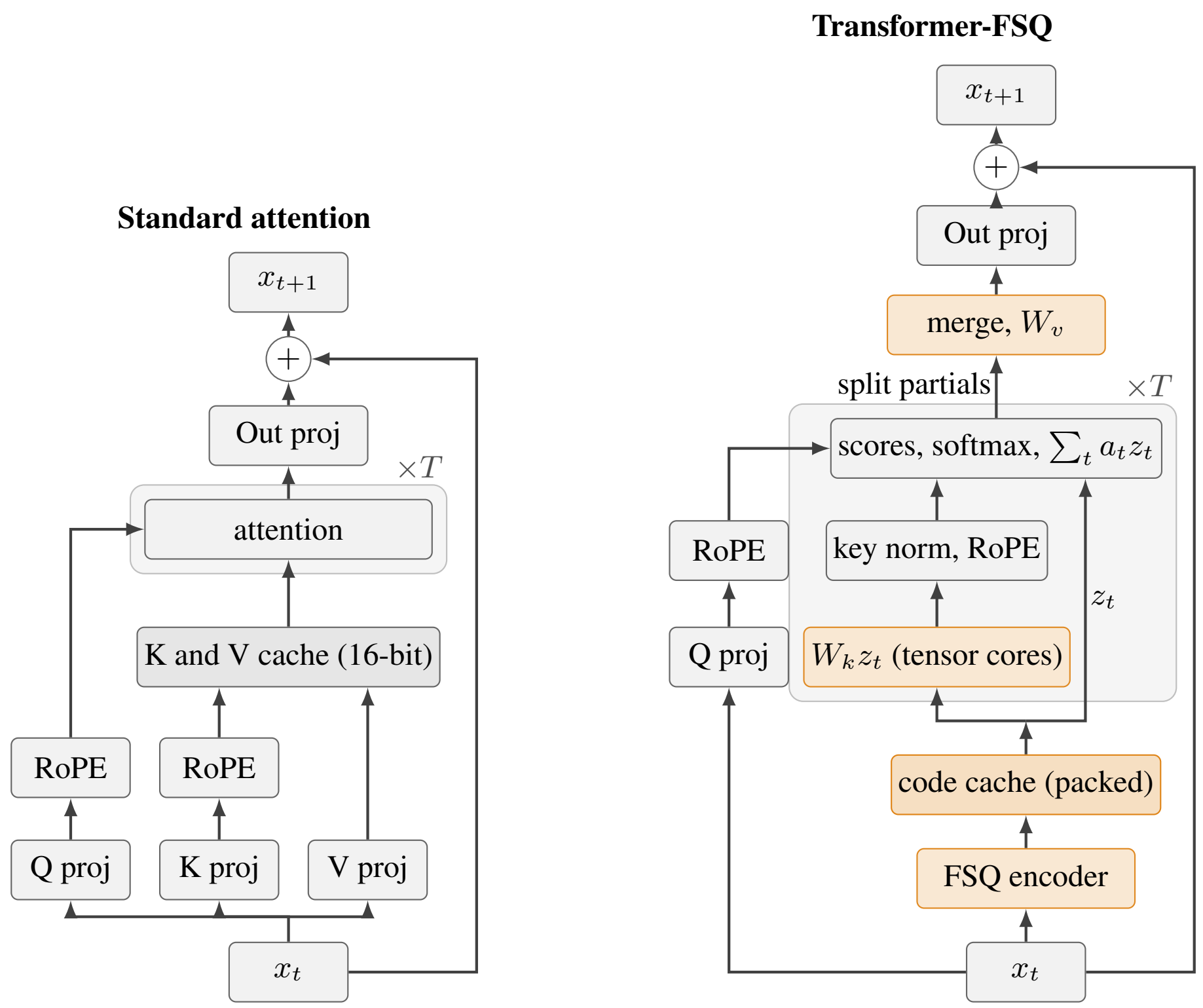

\paragraph{Compress once, before K/V expansion.}
Standard attention first maps the residual state $x_t$ into separate keys and values, possibly across several KV heads, and caches these derived representations. Post-hoc KV quantisation therefore compresses multiple projections of the same underlying state. Transformer-FSQ instead places the bottleneck on the shared residual representation,
\[
    x_t \longrightarrow z_t \longrightarrow
    \{\hat k_t^{(h)},\hat v_t^{(h)}\}_{h=1}^{n_{kv}},
\]
where $x_t$ is the residual state of token $t$, $z_t$ is its cached integer code, and $\hat k_t^{(h)}$ and $\hat v_t^{(h)}$ are the key and value reconstructed from $z_t$ for each of the $n_{kv}$ KV heads $h$. Each token is therefore compressed once, and the same cached code serves all keys, values and KV heads. This is the discrete analogue of caching a shared continuous latent as in MLA. Figure~\ref{fig:fsq-decode-main} illustrates the resulting architecture.

\subsection{The FSQ bottleneck}
\label{sec:bottleneck}

Let $x_t\in\mathbb{R}^{C}$ denote the input to the original key and value projections at token $t$: the residual stream after the layer's pre-attention normalisation.
A learned linear encoder
$E\in\mathbb{R}^{G\times C}$ maps it to a $G$-dimensional latent, which is
bounded and rounded coordinate-wise to $L$ fixed levels by the FSQ quantiser $Q_{\mathrm{FSQ}}$, which applies the map $Q$ of Eq.~\ref{eq:fsq-exact} to each coordinate:
\begin{equation}
    z_t = Q_{\mathrm{FSQ}}(E x_t)
    \in \mathcal{L}_1\times\cdots\times\mathcal{L}_G .
\end{equation}
We use the standard FSQ bounding parameterisation of \citet{mentzer2024finite}, with the exact form and integer packing convention of Eq.~\ref{eq:fsq-exact}.

With
    $b=\lceil\log_2 L\rceil$
bits per coordinate, the cache occupies
    $\left\lceil \frac{Gb}{8}\right\rceil$
bytes per token per layer and stores nothing beside the packed code. All main-text experiments use $L=16$, i.e.\ four bits per coordinate. At equal storage, this permits four times as many coordinates as an FP16 latent.

A single decoder $W_{kv}\in\mathbb{R}^{2n_{kv}d_h\times G}$, split into $W_k$ and $W_v$, reconstructs all KV heads:
\begin{equation}
    \hat k_t =
    \operatorname{RoPE}_t\!\left(
        N_k(W_k z_t)
    \right),
    \qquad
    \hat v_t = W_v z_t ,
\end{equation}
where $N_k$ denotes the model's native key normalisation when present.
The value projection need not be applied separately to every cached token.
For query head $h$ with attention weights $a_t^{(h)}$, writing $W_v$ for the rows of the value decoder that belong to the key-value head $g(h)$ shared by query head $h$,
\begin{equation}
    \sum_t a_t^{(h)} W_v z_t
    =
    W_v\!\left(
        \sum_t a_t^{(h)} z_t
    \right).
    \label{eq:value-fold}
\end{equation}
Values are therefore decoded only after attention has pooled the cached codes. Keys, in contrast, must be reconstructed before their dot products with the query because positional transformations act on each cached position separately. Transformer-FSQ thus trades cache traffic for additional key-side arithmetic; Section~\ref{sec:results} measures this runtime cost.

\subsection{Retrofitting a pretrained transformer}
\label{sec:operator}

Transformer-FSQ is applied as a post-training operator rather than requiring a new model to be pretrained from scratch. For each attention layer, the original write-time key and value projections are replaced by the encoder $E$ and shared decoder $W_{kv}$. These inserted tensors are initialised in closed form from a small calibration set using the pretrained model's own activations, and only the inserted parameters are subsequently finetuned under the language-model objective while the backbone remains frozen. Conversion alone already produces a functional model, analogous to the post-training conversion procedures of Palu \citep{chang2025palu} and TransMLA \citep{meng2025transmla}.

The closed-form initialisation uses principal directions of the calibration residual states for the encoder and a ridge-regularised least-squares fit for the decoder. Let $X \in \mathbb{R}^{m \times C}$ hold the residual states of a calibration sample of $m = 32{,}768$ ClimbMix tokens, and let $u_j$ and $\sigma_j$ ($j = 0,\dots,C-1$, in decreasing order of variance) be the principal directions of the centred $X$ and the standard deviation along each. The encoder's rows are
\begin{equation}
E_{i,:} \;=\; \frac{u_{i \bmod C}^\top}{3\,\sigma_{i \bmod C}}, \qquad i = 0,\dots,G-1,
\end{equation}
so that one standard deviation falls at a third of the $\tanh$ bounding range. The encoder has no bias, and every layer is calibrated at once on the original model's activations. When the latent coordinate width $G$ exceeds the hidden dimension $C$ ($G > C$, as occurs at moderate compression ratios), the encoder basis cycles through the $C$ orthonormal principal directions $i \bmod C$, so that every direction is repeated, and the leading directions once more when $G$ is not a multiple of $C$; repeated rows start identical and nothing breaks their symmetry, so at moderate ratios some coordinates carry duplicate information. The decoder is fitted by ridge regression to reconstruct the residual states from their quantised latents $Z = Q_{\mathrm{FSQ}}(X E^\top) \in \mathbb{R}^{m \times G}$, and is then composed with the original key and value projections:
\begin{equation}
D \;=\; (Z^\top Z + \lambda I_G)^{-1} Z^\top X \in \mathbb{R}^{G \times C},
\qquad
W_{kv} \;=\; [W^K; W^V]\, D^\top \in \mathbb{R}^{2 n_{kv} d_h \times G},
\end{equation}
with penalty $\lambda = 1$. The converted bottleneck replaces $2 C n_{kv} d_h$ parameters with $G(C + 2 n_{kv} d_h)$: at $97.5\times$ on Qwen3-1.7B-Base ($G=84$) the bottlenecks hold under 10M parameters in total and replace about 117M. The bottleneck therefore has fewer parameters than the projections it replaces whenever $G<2Cn_{kv}d_h/(C+2n_{kv}d_h)$: below $G=\fsqBreakevenSix$ on Qwen3-0.6B, $\fsqBreakevenSeventeen$ on Qwen3-1.7B and $\fsqBreakevenEight$ on Qwen3-8B, which at 4 bits per coordinate are the compression ratios $\fsqBreakevenSixRatio\times$, $\fsqBreakevenSeventeenRatio\times$ and $\fsqBreakevenEightRatio\times$. The initialised bottleneck already approximates the pretrained key and value computation.

\subsection{Interaction with positional encoding}
\label{sec:position}

The converted model retains its native positional encoding. In Qwen-style RoPE attention \citep{su2024roformer}, the cached code itself carries no explicit position; position is introduced only after reconstructing each cached key,
    $\hat k_t =
    \operatorname{RoPE}_t\!\left(N_k(W_k z_t)\right).$
Because the position-dependent rotation lies between $W_k$ and the query-key dot product, the key decoder cannot in general be folded into the query. The score of a query $q$ against cached token $t$ is $q^\top\operatorname{RoPE}_t(N_k(W_k z_t))$; without the rotation and the normalisation it would equal $(W_k^\top q)^\top z_t$, and $W_k$ could be applied to the query once per step, as Eq.~\ref{eq:value-fold} applies $W_v$ once. Because $\operatorname{RoPE}_t$ is a different rotation at every position $t$, no single folded query exists. Consequently, our implementation reconstructs keys at every decoding step, which is the main arithmetic overhead of Transformer-FSQ.

This overhead is specific to the interaction with RoPE rather than to the discrete bottleneck itself. Without position-dependent key rotation, the linear key decoder can be absorbed into the query when key normalisation is absent or handled separately. A pre-RoPE variant, FSQ-pre, is described in Appendices~\ref{app:ablations} and~\ref{app:diagrams}; the exact interaction with query-key normalisation is also detailed there.

\subsection{Sizing every cache from one byte budget}
\label{sec:budget}

We compare all methods using the same realised cache-byte budget. Every quantity stored per token is charged, including  quantisation metadata, full-precision windows, positional state, and auxiliary cached representations; Appendix~\ref{app:accounting} lists what each method is charged. Ratios are priced at a context of $8{,}192$ tokens. Reported compression ratios therefore divide the uncompressed FP16 KV-cache size,
\begin{equation}\label{eq:fullcost}
c_{\text{full}} \;=\; 2\, n_{kv}\, d_h \cdot \mathrm{sizeof}(\text{float16})
\qquad \text{bytes per token per layer,}
\end{equation}
by the actual bytes retained by each configuration; $c_{\text{full}}$ is $4096$\,B for the three Qwen models.

\subsection{The decode kernel}
\label{sec:kernel}

A fused Triton kernel \citep{tillet2019triton} computes decode-time attention directly from the packed codes: it reconstructs keys on chip, accumulates attention-weighted codes and applies $W_v$ once per step (Eq.~\ref{eq:value-fold}), so no reconstructed key or value is written to GPU memory. Appendix~\ref{app:kernel} gives its data flow and Appendix~\ref{app:decode} the timing protocol.

\vspace{-0.5em}
\section{Experiments}
\label{sec:experiments}

\begin{table}[t]
\centering
\caption{\textbf{Baselines and their limits on a 16\,GiB GPU.} \textbf{Deepest}: the deepest ratio at which the original authors report downstream accuracy, priced as in \S\ref{sec:budget} (PyramidKV: needle retrieval); for Transformer-FSQ, the deepest we evaluate, where RULER falls to $0.003$ on Qwen3-0.6B-Base. \textbf{Max tokens}: input tokens whose cache fits beside the converted model's weights in 16\,GiB, on Qwen3-0.6B-Base at RULER $\ge 0.30$; for PyramidKV, tokens represented after eviction.}
\label{tab:baselines}

\vspace{4pt}
\small
\setlength{\tabcolsep}{2.7pt}
\begin{tabular}{@{}l l >{\raggedright\arraybackslash}p{3.6cm} c r r@{}}
\toprule
\textbf{Method} & \textbf{Family} & \textbf{Cache contents per token} & \textbf{Trained} & \textbf{Deepest} & \begin{tabular}[t]{@{}r@{}}\textbf{Max tokens}\\\textbf{(16\,GiB, $\ge 0.30$)}\end{tabular} \\
\midrule
\swatch{cPyramidKV}~PyramidKV & token eviction & full-precision KV of kept tokens & no & $\deepPyramidKV\times$ & \reachTabPyramidKV \\
\swatch{cKIVI}~KIVI & scalar quantisation & 2-bit KV, full-precision window & no & $\deepKIVI\times$ & \reachTabKIVI \\
\swatch{cVecInfer}~VecInfer & vector quantisation & codebook indices, window, sinks & no & $\deepVecInfer\times$ & \reachTabVecInfer \\
\swatch{cPalu}~Palu & low-rank projection & quantised low-rank SVD latent & yes & $\deepPalu\times$ & \reachTabPalu \\
\swatch{cTransMLA}~TransMLA & latent attention & latent, fp16 rotary key & yes & $\deepTransMLA\times$ & \reachTabTransMLA \\
\midrule
\swatch{cFSQ}~\begin{tabular}[t]{@{}l@{}}Transformer-\\FSQ (ours)\end{tabular} & finite scalar quantisation & bit-packed integer code & yes & $\deepFSQ\times$ & \reachQwenPointsixFsqSixteenxSixteenGibMillionsValue{}M \\
\bottomrule
\end{tabular}
\end{table}

\vspace{-0.5em}
\subsection{Setup}
\label{sec:setup}
\looseness=-1 \textbf{Models.} We convert three pretrained models, Qwen3-0.6B-Base, Qwen3-1.7B-Base and Qwen3-8B-Base \citep{yang2025qwen3technicalreport} (Appendix~\ref{app:details}). Against Transformer-FSQ we compare representative methods from five distinct families (Table~\ref{tab:baselines}): token eviction (\swatch{cPyramidKV}~PyramidKV \citep{cai2024pyramidkv}), scalar quantisation (\swatch{cKIVI}~KIVI \citep{liu2024kivi}), vector quantisation (\swatch{cVecInfer}~VecInfer \citep{yao2026vecinfer}), low-rank projection (\swatch{cPalu}~Palu \citep{chang2025palu}), and latent attention (\swatch{cTransMLA}~TransMLA \citep{meng2025transmla}). The first three require no training, while Palu and TransMLA are finetuned with the same data, optimiser and token budget as \swatch{cFSQ}~Transformer-FSQ. Each follows its authors' specification except where noted (Appendix~\ref{app:details}), is sized by \S\ref{sec:budget}, and runs as far as that sizing allows. A converted model trains only inserted tensors, with backbone frozen, on ClimbMix \citep{diao2025climb} with AdamW \citep{loshchilov2019adamw}, at $0.25$ tokens per parameter (and $0.025$, $0.075$ on Qwen3-0.6B-Base to test finetuning budget). Conversion and finetuning took $\gpuHoursQwenTotalValue$ GPU-hours on NVIDIA GH200 GPUs (Appendix~\ref{app:compute}). Decode throughput was benchmarked on 4$\times$ NVIDIA L40S GPUs with the fused Triton decoding kernel (\S\ref{sec:kernel}) and, where available, each baseline's released kernels (Appendix~\ref{app:decode}).

\begin{figure}[t]
    \centering
    \includegraphics{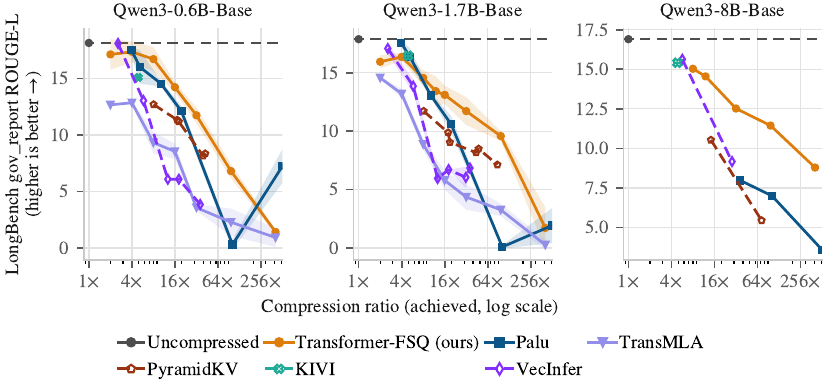}
    
    \caption{\textbf{LongBench \texttt{gov\_report} ROUGE-L against achieved compression} on Qwen3-0.6B, 1.7B and 8B. Solid: finetuned; dashed: training-free; grey horizontal line: uncompressed; bands: one standard deviation over seeds.}
    \label{fig:longbench}
\end{figure}

\looseness=-1 \textbf{Benchmarks.} RULER \citep{hsieh2024ruler} is our primary synthetic benchmark because its items provide exact ground truth across controlled sequence lengths without confounding from pretraining knowledge. We prefer RULER to common alternatives: single-needle retrieval \citep{kamradt2023needle} does not test associations across multiple tokens, sliding-window perplexity is dominated by local context and can miss long-range context loss, and synthetic reasoning suites like BABILong \citep{kuratov2024babilong} fail on uncompressed base models, confounding cache fidelity with reasoning limits. LongBench \citep{bai2024longbench} complements RULER with real-world multi-page tasks (summarisation, question answering and classification), testing whether results on synthetic retrieval carry over to natural documents.

\looseness=-1 \textbf{Metrics.} Long-context retrieval is evaluated at $8{,}192$ tokens across RULER's 13 standard tasks (50 items each). LongBench complements this by evaluating natural documents up to $32{,}768$ tokens (truncated to $32{,}736$) across three tasks. Generation is greedy, up to 32 (\texttt{narrativeqa}), 200 (\texttt{gov\_report}) and 16 (\texttt{trec}) new tokens; a prompt longer than the window minus that budget is truncated in the middle, keeping its start and end, and every method in a panel is scored on the same documents. All evaluations are scored with official task metrics. RULER runs in bfloat16, except KIVI, which runs in float16, the only precision its released kernel accepts; on LongBench, Transformer-FSQ and the uncompressed model run in bfloat16 and the other methods in float32; Appendix~\ref{app:evaluation} compares the two precisions on LongBench for the uncompressed model and Transformer-FSQ (\bfLbVerdict). Reported RULER points average up to three checkpoint seeds covering all items (one on Qwen3-8B-Base), with 95\% bootstrap confidence intervals that resample seeds and then items stratified by task; pairwise leads resample identical seeds and items. This protocol isolates compression degradation from prompt sampling noise under a unified evaluation harness.

\looseness=-1 \textbf{Decode throughput.} Decode throughput is the tokens generated per second by a full-model decode step (every layer, MLP and the output head; no sampling) divided by that of native SDPA decoding (PyTorch's \texttt{scaled\_dot\_product\_attention}) on the uncompressed model, timed in the same process with the same batch and context. We time every combination of batch size (\tpBatches) and $\tpContexts$ contexts from $\tpContextLo$ to $\tpContextHi$ tokens and report the median ratio over these configurations with a 95\% bootstrap interval. Each of the $\tpGpus$ GPUs of a host runs one process pinned to its own CPU core, and a calibration stage before every session admits concurrent timing only if native SDPA alone and beside three loaded GPUs agree within $[\tpCalibBandLo,\tpCalibBandHi]$ (95\% interval). KIVI, VecInfer and Palu use their authors' released kernels (PyramidKV and TransMLA have none; Appendix~\ref{app:decode}): KIVI's matrix--vector kernels for keys and values, VecInfer's fused kernel and Palu's reconstruction kernel, with Palu's key dequantisation, which has no released fused kernel, timed as a separate stage; KIVI and VecInfer run in float16 sessions against a float16 native-SDPA reference, so no ratio divides across precisions.

\subsection{Results}
\label{sec:results}
Our experiments address five central questions, explored in detail below:
\vspace{-2pt}
\begin{itemize}\setlength{\itemsep}{-1pt}\setlength{\parskip}{0pt}\setlength{\parsep}{0pt}\setlength{\topsep}{0pt}\setlength{\partopsep}{0pt}
    \item \textbf{RQ1:} How does Transformer-FSQ fare against baselines across compression regimes?
    \item \textbf{RQ2:} Do synthetic retrieval gains transfer to natural multi-page document summarisation?
    \item \textbf{RQ3:} Does low-bit latent geometry matter under extreme compression? 
    \item \textbf{RQ4:} What is the runtime cost in decode throughput?
    \item \textbf{RQ5:} Does Transformer-FSQ work on a weight-quantised backbone?
\end{itemize}
\vspace{-2pt}

\looseness=-1 \textbf{RQ1: Accuracy across compression regimes.} Across compression regimes, training-free and low-rank methods excel at moderate compression ($\le 5\times$), while Transformer-FSQ is the most accurate from $12\times$ to $97.5\times$, well into extreme compression ($>16\times$). At $\le 5\times$, VecInfer retains at least $93\%$ of uncompressed RULER scores across all models ($\rulerRelQwenPointsixVecinferBninehundredsixtyfourCtxeightkValue\%$, $\rulerRelQwenOnesevenVecinferBninehundredsixtyfourCtxeightkValue\%$, $\rulerRelQwenEightBVecinferBninehundredsixtyfourCtxeightkValue\%$; Figure~\ref{fig:ruler}). Beyond $12\times$, continuous latents and eviction collapse: Palu drops to $\rulerRelQwenPointsixPaluBtwohundredseventytwoCtxeightkValue\%$ at $15.1\times$, while Transformer-FSQ outclasses token eviction by $\rulerLeadQwenPointsixFsqTwelvexOverPyramidkvBthreehundredtwentysixCtxeightkBootMean$ at $12\times$ and $\rulerLeadQwenPointsixFsqThirtytwoxOverPyramidkvBonehundredfourCtxeightkBootMean$ at $32\times$ on Qwen3-0.6B-Base, reaching $\rulerQwenEightBFsqThirtytwoxCtxeightkBootMean$ on Qwen3-8B-Base (a lead of $\rulerLeadQwenEightBFsqThirtytwoxOverPyramidkvBonehundredfourCtxeightkBootMean$; Table~\ref{tab:qwen8b-ruler-breakdown}). Discrete FSQ latents thus establish the retrieval-accuracy--cache-size Pareto frontier from $12\times$ to $97.5\times$.

\looseness=-1 \textbf{RQ2: Natural document summarisation.} Synthetic retrieval resilience carries over to natural multi-page document summarisation. On LongBench \texttt{gov\_report} (Figure~\ref{fig:longbench}), baselines track uncompressed scores ($\govreportQwenEightBUncompressedValue$ ROUGE-L on Qwen3-8B-Base) at moderate compression but collapse at higher ratios. In contrast, Transformer-FSQ scores highest from $32\times$ to $97.5\times$ on all three models; on Qwen3-8B-Base it scores $\govreportQwenEightBFsqThirtytwoxValue$ at $32\times$ and $\govreportQwenEightBFsqNinetyeightxValue$ at $97.5\times$, where Palu at $102.4\times$ scores $\govreportQwenEightBPaluBfortyValue$. Palu's rise in ROUGE-L at extreme compression is likely the metric rewarding degraded text that repeats stopwords. Across all three scales (Tables~\ref{tab:qwen-longbench-breakdown}--\ref{tab:qwen8b-longbench-breakdown}), Transformer-FSQ, finetuned on $1{,}024$-token sequences, keeps the highest ROUGE-L under extreme compression on far longer documents.

\looseness=-1
\textbf{RQ3: Impact of latent geometry.}
Learning to fit a low-bit grid provides the strongest retrieval--memory
operating point under extreme compression
(Table~\ref{tab:discreteness}).
Table~\ref{tab:discreteness} compares bottlenecks at fixed budgets, including quantisation metadata; we discuss four of them at $80$\,B per token per
layer ($51\times$ compression).
All keep the pretrained Qwen backbone frozen and finetune only the
inserted bottleneck parameters.
Transformer-FSQ applies its grid throughout quantisation-aware finetuning
(QAFT); post-hoc FSQ instead trains a continuous latent and applies the
grid only at inference.

At the same width of $160$ and 4-bit precision (scores: mean $\pm$ 95\% bootstrap half-width), QAFT-FSQ achieves
$\qatFsqB\pm\qatFsqBsd$ RULER, compared with $\qatPosthocB\pm\qatPosthocBsd$ for post-hoc FSQ.
This comparison directly supports learning representations that accommodate
the inference-time grid.
Conventional min--max 4-bit QAFT achieves $\qatMinmaxB\pm\qatMinmaxBsd$ with width
$128$ (Table~\ref{tab:discreteness_appendix}): its scale metadata leaves fewer latent coordinates within the
same budget.
Pure continuous rank compression, stored at 16-bit precision, permits
only width $40$ and achieves $\qatRankB\pm\qatRankBsd$.

Together, these results favour a wider low-bit latent trained with its
grid over either inference-only quantisation or a narrow continuous
bottleneck.
The min--max comparison confounds geometry with width, so it only suggests
that FSQ's metadata-free grid is an effective budget-aware design.
Appendix~\ref{app:fsq-levels} further examines the trade-off between latent
width and per-coordinate precision at a fixed cache budget.

\begin{table}[t]
\centering
\caption{\textbf{Discreteness at matched byte budgets:} RULER at $8{,}192$ tokens on Qwen3-0.6B-Base (three seeds, $0.25$ tokens per parameter) at nominal $144$ and $80$\,B per token per layer; mean $\pm$ 95\% bootstrap half-width. The uncompressed model ($4{,}096$\,B) scores $0.718 \pm 0.023$. Continuous latents are $4\times$ narrower to fit the budget; post-hoc variants quantise a wide continuous latent at evaluation. KIVI's per-block scales add $36$ and $20$\,B; TransMLA spends $32$\,B on rotary keys. Training through the 4-bit grid (STE, $L{=}16$) scores highest at both budgets.}
\label{tab:discreteness}
\vspace{2pt}
\small
\setlength{\tabcolsep}{4pt}
\begin{tabular}{@{}l l l c r r@{}}
\toprule
\textbf{Method} & \textbf{Training} & \textbf{Cache} & \textbf{Width} & \textbf{B/tok} & \textbf{RULER} \\
\midrule
\multicolumn{6}{l}{\textit{Nominal budget: $144$\,B per token per layer ($28.4\times$ compression)}} \\[2pt]
Transformer-FSQ (ours) & Discrete (STE, $L{=}16$) & 4-bit implicit grid & 288 & 144 & $\mathbf{0.232 \pm 0.022}$ \\
Continuous latent & Continuous (bf16) & 16-bit float & 72 & 144 & $0.009 \pm 0.008$ \\
Latent + post-hoc FSQ & Continuous (bf16) & 4-bit post-hoc FSQ & 288 & 144 & $0.175 \pm 0.030$ \\
Latent + post-hoc KIVI & Continuous (bf16) & 4-bit blocks + scales & 288 & 180 & $0.206 \pm 0.021$ \\
Palu & Low-rank SVD & 4-bit SVD + scales & 144 & 144 & $0.000 \pm 0.000$ \\
TransMLA & Latent attention & 4-bit + 16-bit RoPE & 64 & 144 & $0.001 \pm 0.001$ \\
\midrule
\multicolumn{6}{l}{\textit{Nominal budget: $80$\,B per token per layer ($51.2\times$ compression)}} \\[2pt]
Transformer-FSQ (ours) & Discrete (STE, $L{=}16$) & 4-bit implicit grid & 160 & 80 & $\mathbf{0.177 \pm 0.017}$ \\
Continuous latent & Continuous (bf16) & 16-bit float & 40 & 80 & $0.001 \pm 0.001$ \\
Latent + post-hoc FSQ & Continuous (bf16) & 4-bit post-hoc FSQ & 160 & 80 & $0.136 \pm 0.014$ \\
Latent + post-hoc KIVI & Continuous (bf16) & 4-bit blocks + scales & 160 & 100 & $0.168 \pm 0.014$ \\
Palu & Low-rank SVD & 4-bit SVD + scales & 80 & 80 & $0.000 \pm 0.000$ \\
TransMLA & Latent attention & 4-bit + 16-bit RoPE & 32 & 76 & $0.000 \pm 0.000$ \\
\bottomrule
\end{tabular}
\end{table}

\begin{figure}[t]
    \centering
    \includegraphics{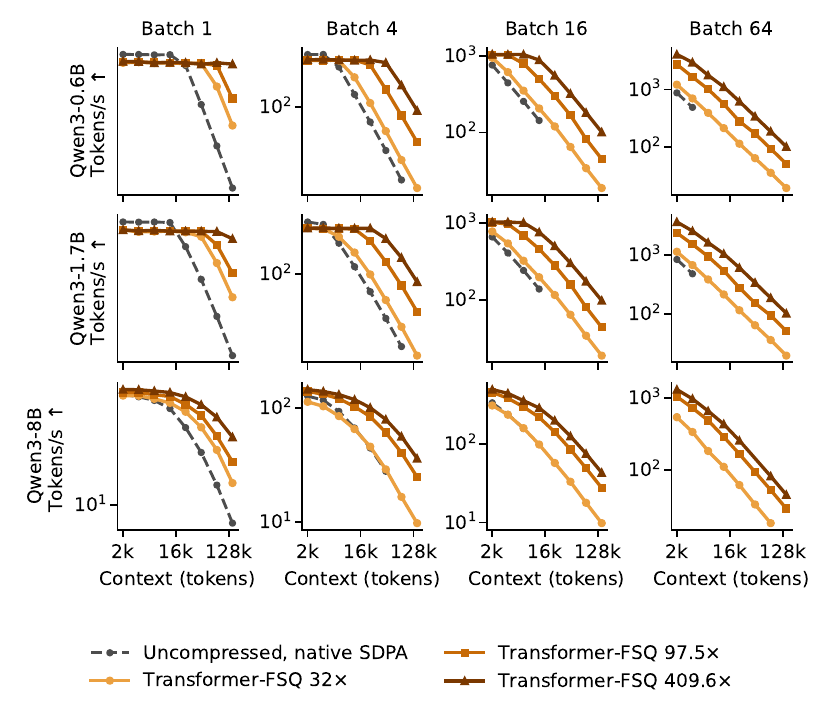}
    \vspace{-8pt}
    \caption{\textbf{Decode throughput against context length} on NVIDIA L40S in bfloat16; rows: model, columns: batch size, both axes logarithmic. Grey dashed: the uncompressed model with native SDPA, timed in the same process. Orange: Transformer-FSQ at $32\times$, $97.5\times$ and $409.6\times$, darker for deeper compression; a line that ends early ran out of memory. Protocol: Appendix~\ref{app:decode}.}
    \label{fig:decode-rate}
\end{figure}

\looseness=-1 \textbf{RQ4: Decode throughput across compression regimes.} Transformer-FSQ matches native SDPA decoding of the uncompressed model at $32\times$ and exceeds it at extreme compression. Benchmarking on NVIDIA L40S GPUs (Figure~\ref{fig:decode-rate}) shows median throughput relative to native SDPA of $\tpSixThirtytwo\times$, $\tpSeventeenThirtytwo\times$ and $\tpEightThirtytwo\times$ at $32\times$ on Qwen3-0.6B, 1.7B and 8B, where the saved cache reads and the per-step key reconstruction approximately offset each other. At extreme compression, memory bandwidth savings dominate arithmetic: $\tpSixFourhundred\times$, $\tpSeventeenFourhundred\times$ and $\tpEightFourhundred\times$ at $409.6\times$, and up to $\tpSixFourhundredKv\times$ on Qwen3-0.6B when the batch holds at least $\tpMinBatchTokens$ tokens. At matched cache bytes, Transformer-FSQ decodes faster than Palu in $\tpMatchPaluSeventeen$--$\tpMatchPaluEight\%$ of batch-size and context configurations, but slower than KIVI's released kernels in most: it is faster in only $\tpMatchKiviSix\%$ of configurations on Qwen3-0.6B and Qwen3-1.7B. At batch 16, native SDPA runs out of memory by $\tpNativeCtxSixHi$ tokens on Qwen3-0.6B and 1.7B, while Transformer-FSQ decodes to $\tpContextHi$, at least $\tpReachFactorLo\times$ further. Per-step key reconstruction is thus outweighed once bandwidth limits dominate.

\looseness=-1 \textbf{RQ5: Weight-quantised backbone.} Transformer-FSQ remains effective when paired with a weight-quantised backbone. To evaluate compatibility under compound compression, we quantise all $\nfLinears$ decoder linear layers of Qwen3-8B-Base to 4-bit NormalFloat (NF4) \citep{dettmers2023qlora} in groups of $\nfGroup$, keeping embeddings and the output head in 16-bit precision, and evaluate long-context retrieval on RULER at $8{,}192$ tokens. The uncompressed model with 4-bit NF4 weights achieves a RULER score of $\NFfourDense$ (within $\NFfourDeltaDense$ of its 16-bit baseline of $\NFfourDenseBf$). For $32\times$ Transformer-FSQ ($128$\,B per token per layer), applying NF4 after finetuning on 16-bit weights drops the RULER score from $\NFfourFSQbf$ to $\NFfourFSQ$ (a loss of $\NFfourDeltaFSQ$). However, when finetuned on the frozen NF4 backbone, Transformer-FSQ recovers to $\NFfourQATFSQ$ (95\% bootstrap CI $\NFfourQATFSQCI$), recovering $\NFfourQATRecovery$ of the loss from applying NF4 after finetuning and trailing the same model on the 16-bit backbone by only $\NFfourQATDelta$. Directly training the discrete KV bottleneck on top of frozen 4-bit weights confirms that Transformer-FSQ adapts to weight-quantised backbones, recovering long-context retrieval to within $\NFfourQATDelta$ of the same Transformer-FSQ model on the 16-bit backbone under joint 4-bit weight and $32\times$ KV-cache compression.

\subsection{Ablations and stacking with eviction}
\looseness=-1 Appendix~\ref{app:ablations} evaluates architectural variants and training dynamics: nonlinear encoders and per-head codes degrade validation loss, MLA-FSQ (Transformer-FSQ codes plus the separate half-precision rotary key of multi-head latent attention) imposes compression ceilings, recent-token windows do not shift Pareto frontiers, and lossless entropy coding saves at most about $13\%$ of cache bytes at a large decode-latency cost (\S\ref{app:entropy}). Because Transformer-FSQ shrinks per-token cache bytes while PyramidKV retains fewer tokens, the two compose at inference (Appendix~\ref{app:stack}). At a RULER threshold of \reachRulerEightkThresholdValue{}, stacking them on Qwen3-0.6B-Base yields an aggregate projected cache capacity of \reachQwenPointsixFsqPyramidkvThirtytwoxSixteenGibMillionsValue{}M tokens across concurrent sequences on a 16\,GiB GPU, \reachFactorValue{}$\times$ the furthest baseline, our head-shared PyramidKV (95\% interval \reachFactorLow--\reachFactorHigh; capped by peak prefill memory). These projections extrapolate $8{,}192$-token scores, extending memory capacity without establishing million-token retrieval accuracy.

\section{Limitations}

\textit{High-dimensional product grids operate beyond initial low-dimensional validations.} While \citet{mentzer2024finite} recommend $L_i \ge 5$ levels for dimensions $d \le 6$ (implicit codebooks $\le 2^{16}$), we deploy $L = 16$ across latent widths from $16$ to $4{,}096$. At wide settings, rounding thousands of coordinates independently acts as product scalar quantisation on a learned basis rather than joint vector quantisation, forgoing multidimensional cell shaping. Every configuration except FSQ-small (Transformer-FSQ with a capped code width and 8-bit levels, used only in the LongBench tables; Appendix~\ref{app:implementation}) uses these 16 levels, compared against 4 and 256 only at $28\times$ and $51\times$ (Appendix~\ref{app:fsq-levels}); at moderate compression, where existing methods remain stronger, fewer coordinates with finer levels are untested.

\textit{Cache memory is verified analytically at conversion.} Because HuggingFace cache objects maintain padded allocations to preserve attention masks, we verify physical footprint by counting stored bytes per token at conversion, rejecting deviations from target budgets. Measured GPU memory directly confirms this analytical model (Appendix~\ref{app:memory-check}): on Qwen3-0.6B-Base, after a $\memPromptTokens$-token prompt, the physical buffer equals the analytic size, tokens $\times$ 28 layers $\times$ $128$\,B ($\memFsqThirtytwoBytes$ bytes).

Our long-context evaluation measures retrieval, tracking and question answering on RULER at one context length, $8{,}192$ tokens, and three LongBench tasks on documents of up to $32{,}768$ tokens; it does not test arithmetic reasoning, code generation or many-shot in-context learning under compression, where prior work reports sharper degradation than on retrieval-style tasks. A method's score here does not show that it preserves multi-step reasoning under compression, which we leave to future measurement.

\section{Conclusion and future work}

Transformer-FSQ caches a quantised code of each token's normalised residual state in place of its keys and values. A learned finite scalar quantisation bottleneck replaces separate KV vectors with a shared, packed integer code, requiring no per-token quantisation metadata. Closed-form conversion and finetuning adapt pretrained models while leaving the backbone frozen. Across Qwen3 models from 0.6B to 8B parameters, Transformer-FSQ achieves the highest measured RULER accuracy from $12\times$ to $97.5\times$ compression and the highest \texttt{gov\_report} ROUGE-L from $32\times$ to $97.5\times$. In our experiments on Qwen3-0.6B-Base, many learned low-bit coordinates outperform a few high-precision ones at the same tight memory budget.

Three directions follow from our findings.

\textbf{Allocation.} Every main configuration in our work gives every layer the same code of 16-level coordinates. Varying code width and levels by layer, as PyramidKV varies kept tokens \citep{cai2024pyramidkv}, sharing projections across layers, and using fewer, finer coordinates at moderate ($2\times$--$4\times$) ratios, where existing methods remain stronger (Appendix~\ref{app:fsq-levels}), could each recover accuracy at the same cache size.

\textbf{Training and evaluation.} Distilling each block from the uncompressed model rather than training on the language-modelling loss alone, and low-rank adapters (LoRA) on the frozen backbone, may improve recovery. Evaluation beyond retrieval at $8{,}192$ tokens, on reasoning, code generation, many-shot learning and longer contexts, would test whether the gains extend to tasks where retaining isolated facts is not enough.

\textbf{Dynamic and hierarchical compression.} A cache could compress lightly while the sequence is short and more as it grows. A stored code can be coarsened in place by rounding it to fewer levels, whereas a coarse code cannot be refined, since rounding discards the detail; whether a richer stored representation or recomputation from the activation could support refinement is untested.

Attention does not need full-precision keys and values in its cache: a learned discrete state per token keeps working at compression ratios where the methods we tested fail.

\FloatBarrier
\clearpage
\section{Statements}

\subsection*{AI use statement}

In this work, we used generative AI tools, large-language-model coding and writing assistants, for two tasks that require disclosure: designing the experiment that stacks Transformer-FSQ with PyramidKV token eviction (Appendix~\ref{app:stack}), and implementing the fused decode kernels, part of the method itself. We have not used generative AI tools to generate synthetic datasets, develop theoretical models, formulate or prove mathematical claims, propose hypotheses, design the other experiments, implement the other parts of the methods, translate, clean or reformat datasets, analyse qualitative data or interpret results; the synthetic RULER items come from RULER's own generator. Additionally, we used generative AI tools for tasks with recommended disclosure: creating and editing the evaluation harness, benchmarking and plotting code (distinct from the compression methods' own implementation, disclosed above), creating the scientific figures and the scripts that draw them, editing the paper for readability, and formatting references.

We have reviewed all AI-assisted work. Code produced with AI assistance is covered by an automated test suite, and every reference was checked against its publisher's or the preprint server's record for title, authors, venue and year. We take responsibility for the final content of this work, including text, claims or artefacts produced with the aid of generative AI.

\subsection*{Reproducibility statement}

The reported tables and figures were generated by scripts from the result files rather than typed by hand. Appendix~\ref{app:models} lists the three released models this paper evaluates, how each is loaded and at what precision; Appendix~\ref{app:conversion} gives the closed-form conversion that inserts Transformer-FSQ's encoder and decoder, including the calibration sample and every constant in Eq.~\ref{eq:fsq-exact} (Appendix~\ref{app:fsq-bottleneck}); Appendix~\ref{app:finetuning} gives the optimiser, learning rate, schedule, data, sequence length, batch size and precision behind every finetuned configuration. Appendix~\ref{app:implementation} gives each baseline's deviation from its published specification, at the bit width, window, sink-token count, rank allocation and codebook grouping we run it at. Every cache size in every table and figure comes from the one byte-budget procedure of \S\ref{sec:budget} and Table~\ref{tab:baselines}, so methods are compared under a uniform byte-pricing model accounting for all cache tensors and metadata.

Appendix~\ref{app:evaluation} gives the RULER and LongBench protocols: task lists, item counts, the fixed item seed, per-task generation lengths, scoring rules and evaluation precision; Appendix~\ref{app:longctx} tabulates every per-task score behind the figures, and Appendix~\ref{app:results} gives the Qwen3-8B-Base campaign's narrower coverage. Appendix~\ref{app:compute} gives the GPU-hours by model and the software versions every run used.

Code for Transformer-FSQ, covering conversion, finetuning, the fused decode kernel, and RULER and LongBench evaluation, is available at \url{https://github.com/ASzecsenyi/transformer-fsq}.

\subsubsection*{Acknowledgments}
This work was supported by PhD studentships within the Centre for Doctoral Training (CDT) in Machine Learning Systems, School of Informatics, University of Edinburgh, funded by the UK Engineering and Physical Sciences Research Council (EPSRC) under grant EP/Y03516X/1. The project also received funding from the European Union's Horizon Europe research and innovation programme under grant agreement No. 101120726 and from UK Research and Innovation (UKRI) through the UK government's Horizon Europe funding Guarantee (10085198). The authors acknowledge the use of resources provided by the Isambard-AI National AI Research Resource (AIRR) \citep{mcintoshsmith2024isambard} under award 0261-4485-3326-1. Isambard-AI is operated by the University of Bristol and funded by the UK Government's Department for Science, Innovation and Technology (DSIT) via UK Research and Innovation and the Science and Technology Facilities Council [ST/AIRR/I-A-I/1023].

\bibliographystyle{iclr2027_conference}
\bibliography{iclr2027_conference}
\newpage
\appendix
\setcounter{topnumber}{3}
\setcounter{totalnumber}{4}

\etocdepthtag.toc{appendix}
\tableofcontents
\newpage

\section{Future directions}

\textbf{Interpretable and recurrent latents.} Discrete codes could be inspected as features \citep{bricken2023monosemanticity}, edited to steer generation \citep{turner2023activation}, or aggregated recurrently in place of attention \citep{katharopoulos2020transformers}, giving a state that does not grow with context.

\section{Extended background and related work}
\label{app:extended-bg}

\subsection{Attention memory and sequence-length scaling}

Under standard multi-head attention \citep{vaswani2017attention}, computing the output for each query token requires inner products against every preceding key, followed by a weighted aggregation over every preceding value. Consequently, decoding step $t$ must access $t-1$ cached key-value pairs per layer. For long sequences and modern LLM serving, this per-token cache storage quickly surpasses the static memory required for model weights.

\textbf{FlashAttention and compute optimisations.} FlashAttention \citep{dao2022flashattention} restructures attention computation using tiling and online softmax rescaling to minimise memory bandwidth between high-bandwidth memory (HBM) and on-chip SRAM. While FlashAttention speeds up prefill and generation by not materialising the intermediate $T \times T$ attention score matrix, it leaves the persistent key-value cache unchanged: each key and value vector must still reside in HBM for subsequent autoregressive decoding steps.

\textbf{Linear attention and recurrent state-space models.} A parallel line of research replaces the growing per-token KV cache with a fixed-size recurrent hidden state. Linear attention methods \citep{katharopoulos2020transformers,lingle2024transformervq} use kernel feature maps to enable efficient associative computation, while linear RNNs and state-space models (SSMs), including Mamba~\citep{gu2024mamba}, Gated DeltaNet~\citep{yang2025gated}, and Bayesian filtering-based models~\citep{shaj2026kalman}, compress sequence history into fixed-size states. Their effectiveness depends on the expressivity of their update rules and underlying online learning objectives~\citep{liu2025longhorn}. Recent architectures such as Echo~\citep{sridhar2026echo} combine recurrent mechanisms to improve associative recall, while hybrid models such as Nemotron-H~\citep{blakeman2025nemotron} and Kimi K3~\citep{team2026kimi} interleave recurrent mixers with full-attention layers.

However, fixed-size recurrent states do not preserve individually addressable representations of arbitrary past tokens, while hybrids with full attention still require a growing KV cache. These approaches also typically require training alternative architectures rather than directly compressing the cache of pretrained Transformers. We therefore focus exclusively on techniques that retain a compressed per-token cache in pretrained standard Transformers.

\subsection{Extended KV-cache taxonomy}

Methods that compress the per-token KV cache operate along different dimensions: along the sequence length (token eviction), along coordinate precision (scalar quantisation), along latent subspace projections (low-rank factorisation), and along coordinate clusters (vector quantisation).

\textbf{Token eviction.} Token eviction methods discard positions judged uninformative. During generation, StreamingLLM \citep{xiao2024streamingllm} observes that preserving initial ``attention sinks'' alongside a local sliding window prevents catastrophic perplexity spikes in autoregressive generation. H2O (Heavy Hitter Oracle) \citep{zhang2023h2o} dynamically tracks tokens with the highest accumulated attention weights, continually evicting positions falling outside the heavy-hitter set and local window. TOVA \citep{oren2024tova} maintains a fixed token quota by evicting the token with the lowest attention score at each decode step. For prompt compression, SnapKV \citep{li2024snapkv} uses an observation window at the end of the prompt to identify consistently attended prompt tokens. \swatch{cPyramidKV}~PyramidKV \citep{cai2024pyramidkv} extends this insight by allocating non-uniform token capacities across layers: early layers retain broader dispersed contexts while deeper layers enforce sharper pruning. However, because eviction reduces token count rather than per-token width, the resulting cache memory remains tied to prompt and generation dynamics.

\textbf{Low-rank projection.} Low-rank methods exploit cross-head redundancy by projecting key and value vectors into narrower continuous subspaces. Multi-head Latent Attention (MLA) \citep{deepseek2024v2} replaces separate per-head keys and values with a shared low-rank latent vector while using a dedicated rotary key for positional encoding. \swatch{cTransMLA}~TransMLA \citep{meng2025transmla}, MHA2MLA \citep{ji2025mha2mla}, and X-EcoMLA \citep{li2025xecomla} demonstrate that standard multi-head attention (MHA) and grouped-query attention (GQA) models can be converted into MLA post-training via rank-reduction and finetuning. \swatch{cPalu}~Palu \citep{chang2025palu} computes low-rank SVD approximations of existing projection weights, caching intermediate latent vectors that are projected back to key and value dimensions at attention time. While continuous low-rank methods reduce cache width, their representations remain unquantised floating-point vectors unless paired with subsequent quantisation stages.

\textbf{Scalar quantisation.} Scalar quantisation maps individual coordinates onto low-bit representations. \swatch{cKIVI}~KIVI \citep{liu2024kivi} quantises keys per channel and values per token to 2 bits without requiring finetuning, while preserving a small full-precision sliding window. KVQuant \citep{hooper2024kvquant} applies non-uniform scalar quantisation pre-RoPE with dense outlier handling. GEAR \citep{kang2024gear} compresses the cache to 2--4 bits and approximates quantisation residual error with low-rank and sparse matrices. QServe \citep{lin2025qserve} integrates 4-bit KV caching alongside 4-bit weights and 8-bit activations. These methods retain original activation dimensions and rely on per-channel or per-token scaling factors and auxiliary buffers.

\textbf{Vector quantisation of the KV cache.} Vector quantisation (VQ) maps activation blocks to discrete centroids in a codebook. VQ-VAE \citep{oord2017neural} and Product Quantisation \citep{jegou2011product} laid the groundwork for discrete representations in continuous spaces. VQ-VAE trains through a straight-through gradient estimator \citep{bengio2013estimating} because the codebook lookup, a nearest-neighbour argmin, is not differentiable; the Gumbel-softmax relaxation \citep{jang2017categorical,maddison2017concrete} is an alternative route around the same non-differentiability, and \citet{scannell2025discrete} use a discrete codebook as the latent state of a world model for continuous control. In LLM caching, Coupled Quantisation \citep{zhang2024coupled} groups 2 to 8 channels into learned codebooks without finetuning, though storing $67.1$M centroid parameters for LLaMA-7B beside the cache. CommVQ \citep{li2025commvq} trains additive multi-stage codebooks using $2 \times 2$ block structures that commute with RoPE. RVQ-KV \citep{kumar2024rvqkv} adapts residual vector quantisation with sequential codebook stages. \swatch{cVecInfer}~VecInfer \citep{yao2026vecinfer} combines per-channel smoothing and Walsh--Hadamard transforms with offline-fitted codebooks at 1.25 to 4 bits per coordinate. GSRQ \citep{kim2026gsrq} decouples magnitude from direction in residual updates, achieving sub-1-bit perplexity. PQCache \citep{zhang2025pqcache} utilises product quantisation to retrieve top-$k$ attention states while storing the complete cache on host memory. NOVA-KV \citep{fernandez2026spend} and VQKV \citep{wang2026vqkv} also explore codebook vector quantisation for keys and values. TurboQuant \citep{zandieh2026turboquant} applies random orthogonal rotations to concentrate coordinate distributions, followed by precomputed scalar quantisation grids.

\subsection{Weight compression}

Weight compression addresses the static portion of the GPU memory budget. For post-training weight quantisation, empirical scaling studies demonstrate that a larger model at lower precision consistently outperforms a smaller model at higher precision under identical memory budgets, with 4-bit weights achieving near-optimal performance across diverse model architectures \mbox{\citep{dettmers2023case}}. Quantisation-aware pretraining pushes this frontier further, achieving viable ternary and 1.58-bit models \citep{ma2024bitnet}. However, while weight footprint remains constant regardless of sequence length, the KV cache scales linearly with generation context, so it dominates memory at long contexts.

\subsection{Rate--distortion bounds}

\textbf{Theoretical bounds.} Rate--distortion theory characterises the fundamental trade-off between bit-rate and reconstruction error for arbitrary quantisers. \citet{zandieh2026turboquant} prove that for any quantiser operating at $b$ bits per coordinate, there exists a unit vector whose expected squared reconstruction error is lower-bounded by $4^{-b}$. Their construction approaches within a factor of $2.7$ of this lower bound. However, this theoretical bound addresses the fidelity of reconstructing a given, arbitrary input vector. Transformer-FSQ does not attempt to reproduce the continuous key and value activations of a frozen model. Instead, its linear encoder and decoder are trained end-to-end such that the representations surviving the discrete integer bottleneck optimise downstream attention task performance. This task-aware training changes what is minimised, a downstream loss rather than reconstruction error, but it does not escape the rate--distortion bound itself: at a fixed bit rate, some distortion is unavoidable under any objective, and training only chooses which directions absorb it.

\section{Experimental details}
\label{app:details}

\subsection{Models}
\label{app:models}

\looseness=-1 Table~\ref{tab:app-models} summarises architectural specifications and uncompressed cache sizes (Eq.~\ref{eq:fullcost}) for the three pretrained transformers evaluated on long-context benchmarks: Qwen3-0.6B-Base, Qwen3-1.7B-Base, and Qwen3-8B-Base \citep{yang2025qwen3technicalreport}, loaded through the Transformers library \citep{wolf2020transformers}. Models are loaded from their public Hugging Face repositories.

\begin{table}[t]
\centering
\caption{\textbf{Models and their memory on a 16\,GiB GPU.} $c_{\text{full}}$: the uncompressed half-precision cache per token per layer (Eq.~\ref{eq:fullcost}), the reference for every compression ratio. The uncompressed context limits are the baseline points of Figure~\ref{fig:frontier}. $^{\ddagger}$Qwen3-8B-Base runs one seed throughout (\S\ref{sec:setup}).}
\label{tab:app-models}
\vspace{4pt}
\small
\setlength{\tabcolsep}{1.8pt}
\begin{tabular}{lrrr}
\toprule
 & \textbf{Qwen3-0.6B-Base} & \textbf{Qwen3-1.7B-Base} & \textbf{Qwen3-8B-Base} \\
\midrule
parameters & $596{,}049{,}920$ & $1{,}720{,}574{,}976$ & $8{,}190{,}735{,}360$ \\
layers & $28$ & $28$ & $36$ \\
model width $C$ & $1024$ & $2048$ & $4096$ \\
query heads & $16$ & $16$ & $32$ \\
key-value heads $n_{kv}$ & $8$ & $8$ & $8$ \\
head width $d_h$ & $128$ & $128$ & $128$ \\
vocabulary & $151{,}936$ & $151{,}936$ & $151{,}936$ \\
sequence length (pretraining) & \multicolumn{3}{c}{$4096$, then $32{,}768$} \\
sequence length (finetuning) & \multicolumn{3}{c}{$1024$} \\
uncompressed cache $c_{\text{full}}$ & \multicolumn{3}{c}{$4096$\,B} \\
weights in half precision & $\weightsQwenPointsixGibValue$\,GiB & $\weightsQwenOnesevenGibValue$\,GiB & $\weightsQwenEightBGibValue$\,GiB \\
cache per token, all layers & $\cacheQwenPointsixPerTokenKibValue$\,KiB & $\cacheQwenOnesevenPerTokenKibValue$\,KiB & $\cacheQwenEightBPerTokenKibValue$\,KiB \\
cache for $200{,}000$ tokens & $\cacheQwenPointsixTwohundredthousandTokensGibValue$\,GiB & $\cacheQwenOnesevenTwohundredthousandTokensGibValue$\,GiB & $\cacheQwenEightBTwohundredthousandTokensGibValue$\,GiB \\
\quad as a multiple of the weights & $\cacheOverWeightsQwenPointsixTwohundredthousandTokensValue\times$ & $\cacheOverWeightsQwenOnesevenTwohundredthousandTokensValue\times$ & $\cacheOverWeightsQwenEightBTwohundredthousandTokensValue\times$ \\
tokens that fit on a 16\,GiB GPU & $\tokensQwenPointsixSixteenGibThousandsValue$\,k & $\tokensQwenOnesevenSixteenGibThousandsValue$\,k & $\tokensQwenEightBSixteenGibThousandsValue$\,k \\
seeds per configuration & $3$ & $3$ & $1^{\ddagger}$ \\
\bottomrule
\end{tabular}
\end{table}

\subsection{Finetuning}
\label{app:finetuning}

\looseness=-1 Table~\ref{tab:app-finetuning} gives the finetuning hyperparameters. All released models are finetuned using AdamW with learning rate $10^{-5}$ and constant weight decay $0.28$ on ClimbMix in bfloat16, the default of our finetuning script, without method-specific tuning. Checkpoints are evaluated in bfloat16 (Appendix~\ref{app:evaluation}). A shared training recipe controls the data and optimisation budget across methods, although it may not be optimal for every method.

\looseness=-1 Each run trains on $\tau N$ tokens, where $\tau \in \{0.025, 0.075, 0.25\}$ tokens per parameter ($N$ being the base model parameter count). Qwen3-0.6B-Base and Qwen3-1.7B-Base train on one GPU with micro-batches of $4$ sequences of $1{,}024$ tokens and gradient accumulation $4$ ($16{,}384$ tokens per step). Qwen3-8B-Base runs exclusively at $\tau = 0.25$ across 4 GH200 GPUs under data parallelism with micro-batch 1 sequence ($1{,}024$ tokens) and gradient accumulation 8, yielding an effective batch of $8{,}192$ tokens per GPU and a global effective batch of $32{,}768$ tokens ($32$ sequences), so the total training tokens are unchanged.

\begin{table}[t]
\centering
\caption{\textbf{Finetuning hyperparameters.} $\tau$: finetuning tokens per parameter; $\tau=0$ is conversion without finetuning. All methods at a model scale share data, optimiser and budget. The released models' results (\S\ref{sec:results}) use $\tau = 0.25$; Figure~\ref{fig:recovery} shows the effect of $\tau$ on Qwen3-0.6B-Base. $^{\ast}$Qwen3-0.6B-Base only.}
\label{tab:app-finetuning}
\vspace{4pt}
\small
\setlength{\tabcolsep}{4pt}
\begin{tabular}{lll}
\toprule
 & \textbf{Qwen3-0.6B / 1.7B-Base} & \textbf{Qwen3-8B-Base} \\
\midrule
finetuning budgets $\tau$ & $0.025^{\ast},\ 0.075^{\ast},\ 0.25$ & $0.25$ \\
optimiser & AdamW & AdamW \\
peak learning rate & $10^{-5}$ & $10^{-5}$ \\
schedule & \multicolumn{2}{l}{linear warm-up over 5\% of steps, cosine decay to 5\% of peak} \\
weight decay & $0.28$, constant & $0.28$, constant \\
data & ClimbMix & ClimbMix \\
sequence length & $1024$ & $1024$ \\
tokens per step & $16{,}384$ ($4\times1024$, accumulation 4) & $32{,}768$ ($4$ GPUs, $1\times1024$, accumulation 8) \\
trainable tensors & encoder, decoder & encoder, decoder \\
backbone & frozen & frozen \\
\bottomrule
\end{tabular}
\end{table}

\begin{figure}[t]
    \centering
    \includegraphics{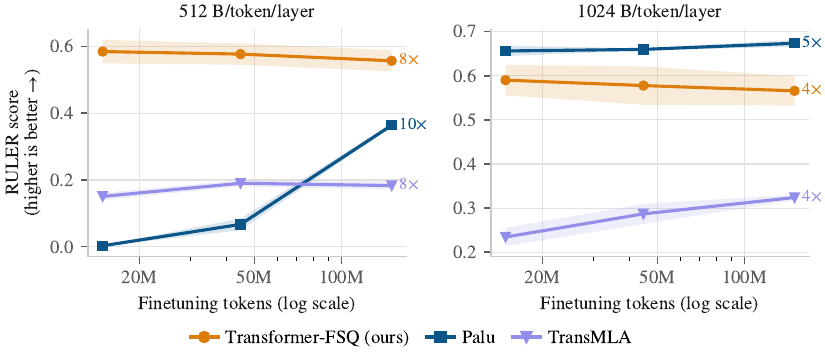}
    \caption{\textbf{RULER against finetuning tokens} on Qwen3-0.6B-Base ($0.025$, $0.075$ and $0.25$ per parameter) at nominal $512$ and $1{,}024$\,B per token per layer. Labels give achieved ratios (Palu stores $400$ and $784$\,B). Transformer-FSQ plateaus with fewer finetuning tokens than the baselines.}
    \label{fig:recovery}
\end{figure}

\subsection{Conversion}
\label{app:conversion}

\looseness=-1 Conversion fits the inserted tensors in closed form (\S\ref{sec:operator}) from activations captured on a calibration sample of ClimbMix: $32{,}768$ tokens for the released models, Qwen3-8B-Base included. The sample is read from the training shards at an offset set by the seed, so without finetuning a seed changes only the calibration text for methods fitted in closed form; the RULER items are drawn once from a fixed item seed shared by every configuration, and generation is greedy. We report conversion without finetuning for Palu at its mildest ratio on Qwen3-0.6B-Base; every TransMLA configuration is finetuned.

\subsection{Cache-byte accounting}
\label{app:accounting}

KIVI is charged its 2-bit keys and values, a scale and zero point per group, and its full-precision window of the $\byResidualWindow$ most recent tokens; VecInfer its codebook indices plus the same window and $\vecinferSinks$ sink tokens (its codebook is fitted once and shared, so it is not charged per token); PyramidKV the full-precision keys and values of the kept prompt tokens and of every generated token, since it evicts only at prefill, plus a $\byPosBytes$-byte position per kept token; Palu its quantised low-rank latent with its scales; TransMLA its latent and its separate rotary key; and Transformer-FSQ its packed code, plus the window and sink tokens in the variant that keeps them. The rotary table is not charged: it is computed from the model's own frequencies and does not grow with the number of cached tokens; methods that cache their own rotary key, such as TransMLA, are charged for that key.

\subsection{Evaluation}
\label{app:evaluation}

\looseness=-1 RULER \citep{hsieh2024ruler}, the suite behind every RULER number this paper reports (\S\ref{sec:setup}), evaluates all 13 official tasks at $8{,}192$ tokens with 50 items per task across all three Qwen sizes, drawn once from a fixed item seed and shared by every configuration. Scoring follows RULER's published protocol, by substring containment of the reference answers: the fraction of references the output contains on multi-answer tasks and whether it contains any reference on QA tasks, with outputs capped at official generation lengths. As detailed in \S\ref{sec:setup}, evaluation forward passes run in bfloat16, and KIVI's in float16.

LongBench \citep{bai2024longbench} runs \texttt{narrativeqa}, \texttt{gov\_report} and \texttt{trec} at $20$ examples per task, scored by F1, ROUGE-L and classification accuracy ($0$ to $100$). On Qwen3-8B-Base the LongBench draw is $200$ documents per task, and each task is scored on the documents that every configuration completed: some methods run out of memory in prefill on the longest documents, leaving $180$ \texttt{gov\_report}, $70$ \texttt{narrativeqa} and all $200$ \texttt{trec} documents (Appendix~\ref{app:results}). Prompts exceeding $32{,}768$ tokens are middle-truncated to $32{,}736$ tokens. Long-context benchmarks run three seeds at $0.25$ tokens per parameter, except Qwen3-8B-Base, which runs one seed throughout (\S\ref{sec:setup}).

\textbf{Precision tests.} For RULER, a configuration is one bfloat16 file paired item by item with the float32 file of the same configuration. A group is one (model, method) pair, pooling its configurations weighted by item count. The 95\% bootstrap resamples configurations, then items within each. The precisions are equivalent if the interval lies inside $\pm\bfMargin$, different if it lies wholly outside that band, and inconclusive otherwise; separately, no configuration's mean difference may exceed $\bfOutlierBound$. LongBench applies this rule on a 0--100 scale, with margin $\bfLbMargin$ and per-configuration bound $\bfLbOutlierBound$, to the uncompressed model and Transformer-FSQ per model. KIVI runs in float16, the only precision its kernel accepts, and is reported separately. NF4 rounds every decoder linear layer of Qwen3-8B-Base, inserted ones included but not the embeddings or output head, in groups of $\nfGroup$ at evaluation after finetuning; the hypothesis holds if the upper bound of the interval of $\Delta_{\mathrm{FSQ}}-\Delta_{\mathrm{dense}}$, with $\Delta$ the bfloat16 minus NF4 score, lies below $\nfMargin$. Finetuning directly on the NF4-quantised backbone recovers the RULER score to $\NFfourQATFSQ$ (recovering $\NFfourQATRecovery$ of the post-hoc gap), consistent with Transformer-FSQ adapting to 4-bit backbone weights during recovery training.

\textbf{Benchmark selection.} We select RULER and LongBench to probe both structured associative retrieval and realistic document comprehension. Vanilla single-needle retrieval is insufficiently discriminative, as coarse scalar quantisation or naive token retention can retrieve isolated keys against uniform text without preserving relational context. In contrast, RULER tests 13 synthetic patterns---including multi-key, multi-value, and variable tracking---directly exposing when aggressive compression destroys token relationships. LongBench complements synthetic retrieval by evaluating multi-page real documents on summarisation, narrative QA, and classification.

Unless stated otherwise, a spread is a sample standard deviation across seeds, and an interval is a 95\% bootstrap interval that resamples seeds and then items stratified by task (items only for the one-seed Qwen3-8B-Base).

\subsection{Compute}
\label{app:compute}

The reported configurations' conversion and finetuning took $\gpuHoursQwenTotalValue$ GPU-hours on NVIDIA GH200 GPUs (1 GPU per run for $\le 1.7$B, 4 GPUs for 8B); Table~\ref{tab:hf-compute-breakdown} gives the average time of one run by model and finetuning budget. Over the whole project, Slurm accounting of every job on Isambard-AI, including jobs that failed or reached the time limit, records about $\gpuHoursProjectTotalValue$ GH200 GPU-hours: $\gpuHoursProjectTrainValue$ for all conversion and finetuning, including unreported baseline configurations, ablations and Qwen3-8B-Base runs, $\gpuHoursProjectEvalValue$ for RULER and LongBench evaluation, and $\gpuHoursProjectThroughputValue$ for decode-throughput benchmarking. The final evaluation campaign used a further $\gpuHoursFinalEvalValue$ GPU-hours on NVIDIA RTX 6000 Ada GPUs, $\gpuHoursFinalEvalFailedValue$ of them in failed jobs. Reported decode throughput was measured on NVIDIA L40S GPUs.

\begin{table}[t]
\centering
\small
\begin{tabular}{lrrr}
\toprule
Model & $\tau=0.025$ & $\tau=0.075$ & $\tau=0.25$ \\
\midrule
Qwen3-0.6B-Base & 0.19 \scriptsize{$n{=}30$; 0.15--0.25} & 0.56 \scriptsize{$n{=}27$; 0.44--0.78} & 2.59 \scriptsize{$n{=}54$; 1.47--5.14} \\
Qwen3-1.7B-Base & -- & -- & 7.32 \scriptsize{$n{=}63$; 5.57--9.19} \\
Qwen3-8B-Base & -- & -- & 138.34 (34.6\,h) \scriptsize{$n{=}9$; 108.88--158.72} \\
\bottomrule
\end{tabular}
\caption{\textbf{Conversion and finetuning GPU-hours per run} on NVIDIA GH200 (one GPU per run; four for Qwen3-8B). Each cell: the mean over $n$ finished runs of the finetuned methods (Transformer-FSQ, Palu, TransMLA; all budgets and seeds), then the min--max range; --: not run; for Qwen3-8B, mean wall-clock hours in parentheses. $\tau$: finetuning tokens per parameter.}
\label{tab:hf-compute-breakdown}
\end{table}

\subsection{Implementation}
\label{app:implementation}

All models are implemented in PyTorch $2.9.1$ \citep{paszke2019pytorch} against CUDA $12.8$ wheels, and the fused decode kernels are written in Triton $3.5.1$ \citep{tillet2019triton}. The Qwen conversion and finetuning path loads models through Transformers \citep{wolf2020transformers} ($4.57.6$). Baselines follow published specifications: KIVI preserves its 128-token full-precision window, while VecInfer keeps 4 sink tokens following StreamingLLM \citep{xiao2024streamingllm}. PyramidKV selects tokens per head, as published, following SnapKV.

VecInfer uses shared $k$-means codebooks across heads and symmetric bit allocations; its released fused kernel is used for decode timing. Transformer-FSQ keeps no full-precision window or sink tokens except in the ablation of Appendix~\ref{app:ablations}.

\textbf{Choice of baselines.} We take one published method per family of KV-cache compression (Table~\ref{tab:baselines}). For token eviction, PyramidKV \citep{cai2024pyramidkv} rather than SnapKV, StreamingLLM or H2O, since it also varies the budget across layers. For scalar quantisation, KIVI \citep{liu2024kivi} rather than round-to-nearest, since its per-channel keys, per-token values and full-precision window handle activation outliers. For vector quantisation, VecInfer \citep{yao2026vecinfer} rather than Coupled Quantisation, which stores tens of millions of codebook parameters; VecInfer spreads outliers with a Walsh--Hadamard transform instead. For low-rank projection, Palu \citep{chang2025palu}, and for latent attention, TransMLA \citep{meng2025transmla}, both of which convert a pretrained model by finetuning rather than pretraining an MLA architecture \citep{deepseek2024v2}. TurboQuant \citep{zandieh2026turboquant} rotates every cached key and value by a normalised Walsh--Hadamard matrix and quantises it per vector on a min--max grid, with an optional one-bit QJL residual correction; we run its PyTorch reference path training-free at nominal budgets of $1{,}024$ and $2{,}048$\,B per token per layer (one seed), and it appears only in Figure~\ref{fig:ruler}. FSQ-small, which appears only in the LongBench tables, is Transformer-FSQ with its code width capped where the bottleneck stops adding parameters over the key and value projections it replaces, using 8-bit levels where they fit the budget. State-space and linear-attention models keep no per-token cache and are not compared (Appendix~\ref{app:extended-bg}).

\section{Additional results}
\label{app:results}

\subsection{Qwen3-8B-Base}

\textit{Qwen3-8B-Base retrieval and question classification.} Figure~\ref{fig:ruler} reports RULER and Figure~\ref{fig:longbench} reports \texttt{gov\_report} ($50$ RULER examples per task, $180$ \texttt{gov\_report} documents, seed 42 throughout). At extreme compression, Transformer-FSQ at $409.6\times$ and Palu at $512\times$ reach floor retrieval scores, as on the smaller models. On \texttt{trec} (Figure~\ref{fig:longbench-trec}; 200 documents, uncompressed score $\trecQwenEightBUncompressedValue$), KIVI and VecInfer stay close to uncompressed at moderate compression ($\trecQwenEightBKiviBonethousandtwentyfourValue$ at $\compressionQwenEightBKiviBonethousandtwentyfourTrecValue\times$, $\trecQwenEightBVecinferBninehundredfourValue$ at $\compressionQwenEightBVecinferBninehundredfourTrecValue\times$) before dropping. For methods whose cache depends on the tokens held (KIVI, VecInfer, PyramidKV), ratios in this subsection are those of the named task's documents; Table~\ref{tab:qwen8b-longbench-breakdown} averages them over a row's tasks. Transformer-FSQ maintains $\trecQwenEightBFsqThirtytwoxValue$ at $32\times$ and $\trecQwenEightBFsqNinetyeightxValue$ at $97.5\times$, outperforming Palu ($\trecQwenEightBPaluBonehundredtwelveValue$ and $\trecQwenEightBPaluBfortyValue$). TransMLA ran out of memory during long-document prefill and is omitted.

\textit{Document-level narrative comprehension on 8B parameters.} On LongBench \texttt{narrativeqa} (Figure~\ref{fig:longbench-narrativeqa}; the 70 documents every configuration completed, the others exceeding prefill memory for at least one method), the uncompressed model scores $\narrativeqaQwenEightBUncompressedValue$ F1. Token eviction performs strongly on this narrative task: PyramidKV scores $\narrativeqaQwenEightBPyramidkvBonethousandtwentythreeValue$ at $\compressionQwenEightBPyramidkvBonethousandtwentythreeNarrativeqaValue\times$ and $\narrativeqaQwenEightBPyramidkvBthreehundredfortyValue$ at $\compressionQwenEightBPyramidkvBthreehundredfortyNarrativeqaValue\times$. Transformer-FSQ scores $\narrativeqaQwenEightBFsqThirtytwoxValue$ ($32\times$) and $\narrativeqaQwenEightBFsqNinetyeightxValue$ ($97.5\times$), above Palu's $\narrativeqaQwenEightBPaluBonehundredtwelveValue$ and $\narrativeqaQwenEightBPaluBfortyValue$.

\subsection{Long context}
\label{app:longctx-main}
\begin{figure}[!ht]
    \centering
    \includegraphics{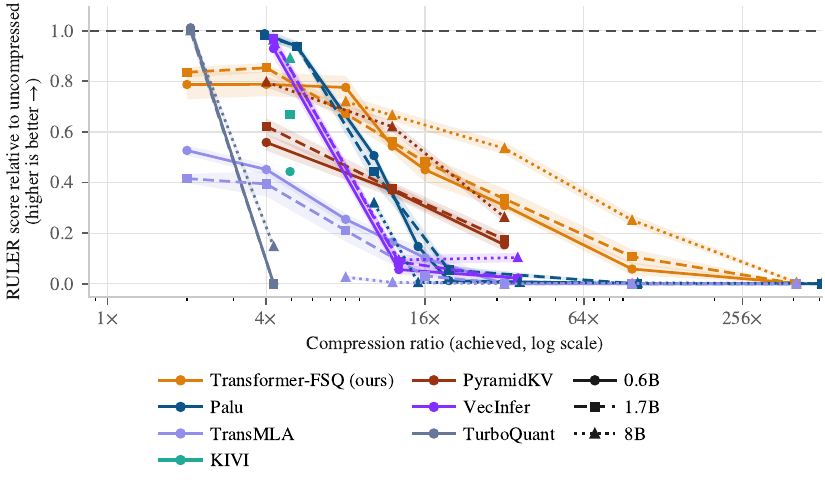}
    \caption{\textbf{RULER relative to the uncompressed model:} score at $8{,}192$ tokens divided by the same model's uncompressed score; bands as in Figure~\ref{fig:ruler}. Below about $5\times$, VecInfer and Palu keep the most; from $32\times$ to $97.5\times$, Transformer-FSQ does.}
\label{fig:ruler-relative}
\end{figure}

\begin{figure}[!ht]
    \centering
    \includegraphics{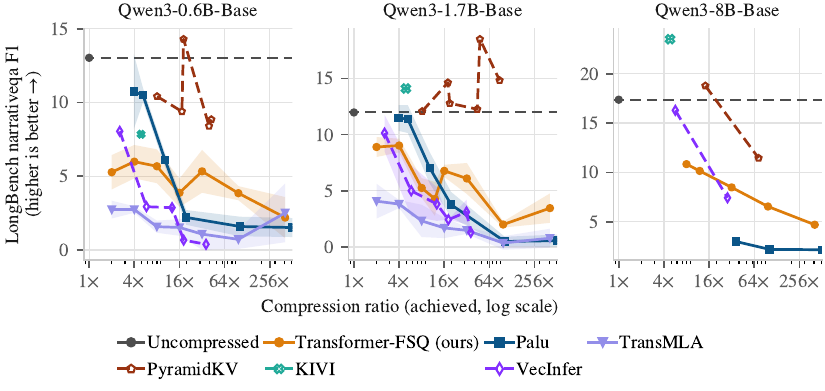}
    \caption{\textbf{LongBench \texttt{narrativeqa} F1 (0--100) against achieved compression.} Solid: finetuned; dashed: training-free; grey horizontal line: uncompressed. PyramidKV scores highest on 0.6B and 1.7B; on 8B, Transformer-FSQ scores above Palu.}
    \label{fig:longbench-narrativeqa}
\end{figure}

\begin{figure}[!ht]
    \centering
    \includegraphics{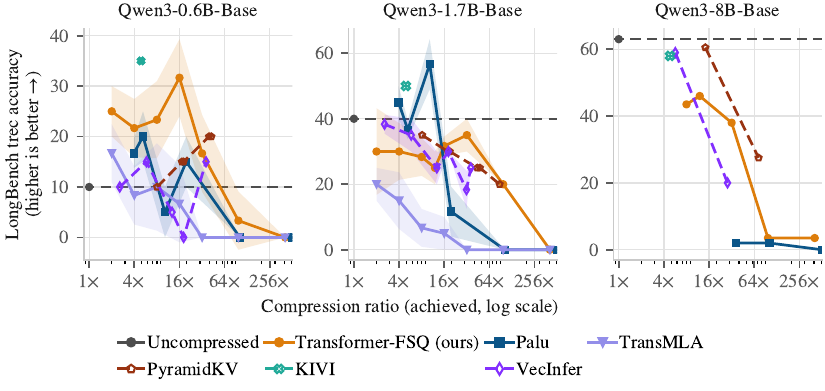}
    \caption{\textbf{LongBench \texttt{trec} accuracy (0--100) against achieved compression.} Line styles as in Figure~\ref{fig:longbench-narrativeqa}.}
    \label{fig:longbench-trec}
\end{figure}

This subsection evaluates long-context retrieval and comprehension across all three Qwen models under compression. Figure~\ref{fig:ruler-relative} presents RULER scores relative to uncompressed models, while Figures~\ref{fig:longbench-narrativeqa} and~\ref{fig:longbench-trec} report LongBench \texttt{narrativeqa} and \texttt{trec} performance across models. Complete per-task scores appear in Tables~\ref{tab:qwen-ruler-breakdown}--\ref{tab:qwen8b-ruler-breakdown} and~\ref{tab:qwen-longbench-breakdown}--\ref{tab:qwen8b-longbench-breakdown}.

\looseness=-1 On Qwen3-0.6B-Base (Figure~\ref{fig:ruler-relative}), Transformer-FSQ keeps $\rulerRelQwenPointsixFsqFourxCtxeightkValue\%$ of the uncompressed RULER score at $4\times$ and $\rulerRelQwenPointsixFsqEightxCtxeightkValue\%$ at $8\times$ (Table~\ref{tab:qwen-ruler-breakdown}), ahead of TransMLA's $\rulerRelQwenPointsixTransmlaEightxCtxeightkValue\%$ at $8\times$ and Palu's $\rulerRelQwenPointsixPaluBtwohundredseventytwoCtxeightkValue\%$ at $15.1\times$. Single-key retrieval (\texttt{niah\_single\_1}) survives longest at extreme compression, while multi-key retrieval with distractors and variable tracking collapse past $16\times$. Training-free baselines drop earlier: VecInfer retains $\rulerRelQwenPointsixVecinferBninehundredsixtyfourCtxeightkValue\%$ at $4.3\times$ but falls to $\rulerRelQwenPointsixVecinferBthreehundredtwentyfourCtxeightkValue\%$ at $12.7\times$, and PyramidKV holds $\rulerRelQwenPointsixPyramidkvBonethousandthirtyfourCtxeightkValue\%$ at $4.0\times$.

\looseness=-1 On Qwen3-1.7B-Base (Figure~\ref{fig:ruler-relative}), the same hierarchy holds at higher retention: Transformer-FSQ keeps $\rulerRelQwenOnesevenFsqEightxCtxeightkValue\%$ of uncompressed RULER at $8\times$ and $\rulerRelQwenOnesevenFsqThirtytwoxCtxeightkValue\%$ at $32\times$ (Table~\ref{tab:qwen17b-ruler-breakdown}), exceeding TransMLA ($\rulerRelQwenOnesevenTransmlaEightxCtxeightkValue\%$ at $8\times$) and Palu ($\rulerRelQwenOnesevenPaluBsevenhundredeightyfourCtxeightkValue\%$ at $5.2\times$, falling to $\rulerRelQwenOnesevenPaluBtwohundredeightCtxeightkValue\%$ at $19.7\times$). VecInfer drops to $\rulerRelQwenOnesevenVecinferBthreehundredtwentyfourCtxeightkValue\%$ at $12.7\times$, while PyramidKV holds $\rulerRelQwenOnesevenPyramidkvBonethousandthirtytwoCtxeightkValue\%$ at $4.0\times$. As on 0.6B, single-key retrieval survives longest; the ordering of methods is unchanged and every score is higher.

\looseness=-1 Document comprehension and question classification reveal complementary strengths between uniform compression and token eviction (Figures~\ref{fig:longbench-narrativeqa} and~\ref{fig:longbench-trec}). On LongBench \texttt{narrativeqa}, token eviction (PyramidKV) stays within a few points of the uncompressed model across budgets ($8.4$--$14.3$ F1 on 0.6B, $12.1$--$18.5$ on 1.7B), outperforming uniform compression on long narrative tracking where broad context can be pruned. Conversely, on \texttt{trec} classification and \texttt{gov\_report} summarisation (Figures~\ref{fig:longbench-trec} and~\ref{fig:longbench}), Transformer-FSQ scores above Palu and TransMLA from $32\times$ to $97.5\times$, where both collapse.

\looseness=-1 On Qwen3-8B-Base (Table~\ref{tab:qwen8b-ruler-breakdown}), Transformer-FSQ keeps $\rulerRelQwenEightBFsqThirtytwoxCtxeightkValue\%$ of RULER at $32\times$ and $\rulerRelQwenEightBFsqNinetyeightxCtxeightkValue\%$ at $97.5\times$, with single-key retrieval again surviving longest.

\subsection{Combining Transformer-FSQ with eviction}
\label{app:stack}

\begin{figure}[t]
\centering
\includegraphics{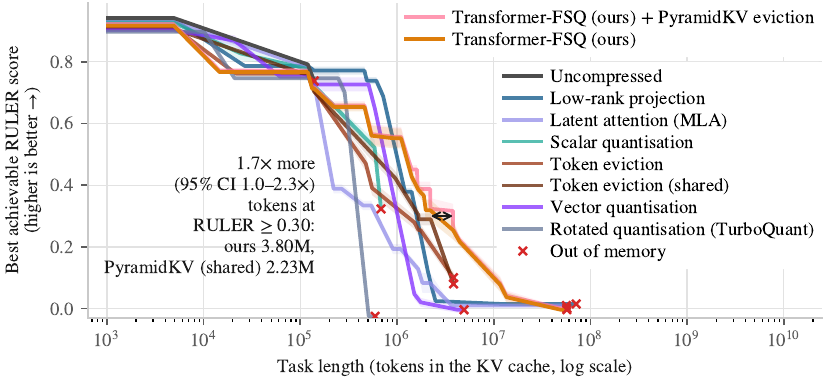}
\caption{\textbf{Tokens held on a 16\,GiB GPU against RULER at $8{,}192$ tokens.} Crosses: out of memory. The stacked line includes only stacks in which Transformer-FSQ supplies at least half the compression in log terms. At matched RULER thresholds, Transformer-FSQ and its eviction stacks hold up to \reachFactorValue{}$\times$ more tokens than the baselines.}
\label{fig:frontier}
\end{figure}

\textit{Stacking FSQ with token eviction on Qwen3-0.6B-Base.} We applied PyramidKV at inference to finetuned Transformer-FSQ checkpoints without retraining, using our own variant of PyramidKV rather than the published method: it keeps one set of prompt tokens per layer, shared by all its heads, where published PyramidKV selects tokens per head, so that the kept tokens line up with Transformer-FSQ's codes, which every head shares. This variant scores $0.13$--$0.24$ higher on RULER than the published per-head method; it is the PyramidKV of this subsection, of Figure~\ref{fig:frontier} and of Table~\ref{tab:stack-grid-8k}, and every other PyramidKV result in this paper is the published method. We evaluated RULER ($8{,}192$ tokens) across combinations where FSQ carries at least half the log-compression (Table~\ref{tab:stack-grid-8k}). Stacked ratios charge PyramidKV's 4-byte position per kept prompt token and the generated tokens at the full code width, so they are not the product of the two standalone ratios. Stacking FSQ $8\times$ with PyramidKV $2\times$ achieves $16\times$ total compression ($256$\,B/token/layer) and scores $\stackEightTwoSix$, outperforming PyramidKV alone at the matched budget ($\pkvMatchedSixteenSix$ at $255.75$\,B). Stacking FSQ $16\times$ with PyramidKV $2\times$ achieves $32\times$ total compression ($128$\,B/token/layer) and scores $\stackSixteenTwoSix$, outperforming matched PyramidKV alone ($\pkvMatchedThirtytwoSix$ at $127.75$\,B). At $64$\,B/token/layer ($64\times$, FSQ $16\times$ with PyramidKV $4\times$), the stack maintains $\stackSixteenFourSix$, whereas PyramidKV alone cannot reach $64$\,B at $8{,}192$ tokens, below its minimum window.

\textit{Diminishing returns at extreme stacking ratios.} Returns diminish as Transformer-FSQ's own compression grows (FSQ $32\times$ stacked with PyramidKV $32\times$ drops to $\stackThirtytwoThirtytwoSix$). On Qwen3-1.7B-Base (three seeds), the $8\times + 2\times$ stack scores $\stackEightTwoSeventeen \pm \stackEightTwoSeventeenHalf$ against $\pkvMatchedSixteenSeventeen$ for matched PyramidKV alone, and single-seed stacks on Qwen3-8B-Base follow the same trend. Figure~\ref{fig:frontier} summarises the resulting capability frontier on a 16\,GiB GPU.

\begin{figure}[t]
  \centering
  \includegraphics{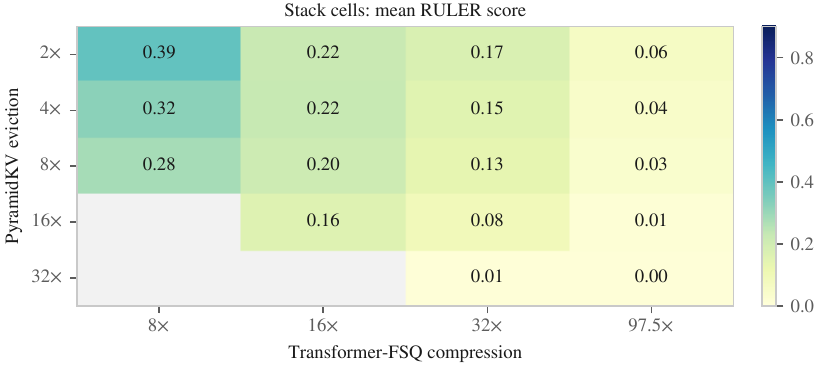}
  \caption{\textbf{Mean RULER at $8{,}192$ tokens for each Transformer-FSQ$\,\times\,$PyramidKV stack} on Qwen3-0.6B-Base. Grey cells fail the rule, fixed before the runs, that Transformer-FSQ supplies at least half the compression in log terms; values in Table~\ref{tab:stack-grid-8k}. Retrieval degrades gradually up to $32\times$ total and falls off at extreme stacked ratios.}
\end{figure}

\begin{table}[t]
\centering
\caption{\textbf{Stack-grid results at $8{,}192$ tokens on Qwen3-0.6B-Base} (RULER, 13 tasks of 50 items; Appendix~\ref{app:stack}). Every stack that passes the rule fixed before the runs is listed, against Transformer-FSQ alone, the uncompressed model, and PyramidKV alone at the same achieved total budget (own block below). Ratios and B/token/layer are achieved, not nominal. Score: mean over complete seeds $\pm$ item-bootstrap half-width (Appendix~\ref{app:evaluation}); \textbf{seeds}: complete finetuning seeds.}
\label{tab:stack-grid-8k}
\vspace{4pt}
\small
\begin{tabular}{@{}cc rr c c@{}}
\toprule
\textbf{FSQ} & \textbf{PyramidKV} & \textbf{Total ratio} & \textbf{B/tok/layer} & \textbf{Score $\pm$ half-CI} & \textbf{Seeds} \\
\midrule
-- & -- & $1.0\times$ & 4096 & $0.72 \pm 0.02$ & 1 \\
$8\times$ & -- & $8.0\times$ & 512 & $0.56 \pm 0.03$ & 3 \\
$16\times$ & -- & $16.0\times$ & 256 & $0.32 \pm 0.03$ & 3 \\
$32\times$ & -- & $32.0\times$ & 128 & $0.22 \pm 0.01$ & 3 \\
$97.5\times$ & -- & $97.5\times$ & 42 & $0.04 \pm 0.02$ & 3 \\
\midrule
$8\times$ & $2\times$ & $16.0\times$ & 256 & $0.39 \pm 0.02$ & 1 \\
$8\times$ & $4\times$ & $32.0\times$ & 128 & $0.32 \pm 0.02$ & 1 \\
$8\times$ & $8\times$ & $64.0\times$ & 64 & $0.28 \pm 0.02$ & 1 \\
\midrule
$16\times$ & $2\times$ & $32.0\times$ & 128 & $0.22 \pm 0.02$ & 1 \\
$16\times$ & $4\times$ & $64.0\times$ & 64 & $0.22 \pm 0.02$ & 1 \\
$16\times$ & $8\times$ & $128.1\times$ & 32 & $0.20 \pm 0.02$ & 1 \\
$16\times$ & $16\times$ & $256.3\times$ & 16 & $0.16 \pm 0.02$ & 1 \\
\midrule
$32\times$ & $2\times$ & $64.0\times$ & 64 & $0.17 \pm 0.02$ & 1 \\
$32\times$ & $4\times$ & $128.0\times$ & 32 & $0.15 \pm 0.02$ & 1 \\
$32\times$ & $8\times$ & $256.1\times$ & 16 & $0.13 \pm 0.02$ & 1 \\
$32\times$ & $16\times$ & $512.5\times$ & 8.0 & $0.08 \pm 0.02$ & 1 \\
$32\times$ & $32\times$ & $1026.0\times$ & 4.0 & $0.01 \pm 0.01$ & 1 \\
\midrule
$97.5\times$ & $2\times$ & $195.1\times$ & 21 & $0.06 \pm 0.02$ & 1 \\
$97.5\times$ & $4\times$ & $390.2\times$ & 10 & $0.04 \pm 0.01$ & 1 \\
$97.5\times$ & $8\times$ & $780.6\times$ & 5.2 & $0.03 \pm 0.01$ & 1 \\
$97.5\times$ & $16\times$ & $1562.1\times$ & 2.6 & $0.01 \pm 0.01$ & 1 \\
$97.5\times$ & $32\times$ & $3129.1\times$ & 1.3 & $0.00 \pm 0.01$ & 1 \\
\midrule
\multicolumn{6}{@{}l}{\textbf{PyramidKV (shared, our variant; Appendix~\ref{app:implementation}) alone, at a stack cell's own achieved total budget}} \\
-- & matched & $32.1\times$ & 127.75 & $0.10$ & 1 \\
-- & matched & $16.0\times$ & 255.75 & $0.30$ & 1 \\
\bottomrule
\end{tabular}
\end{table}

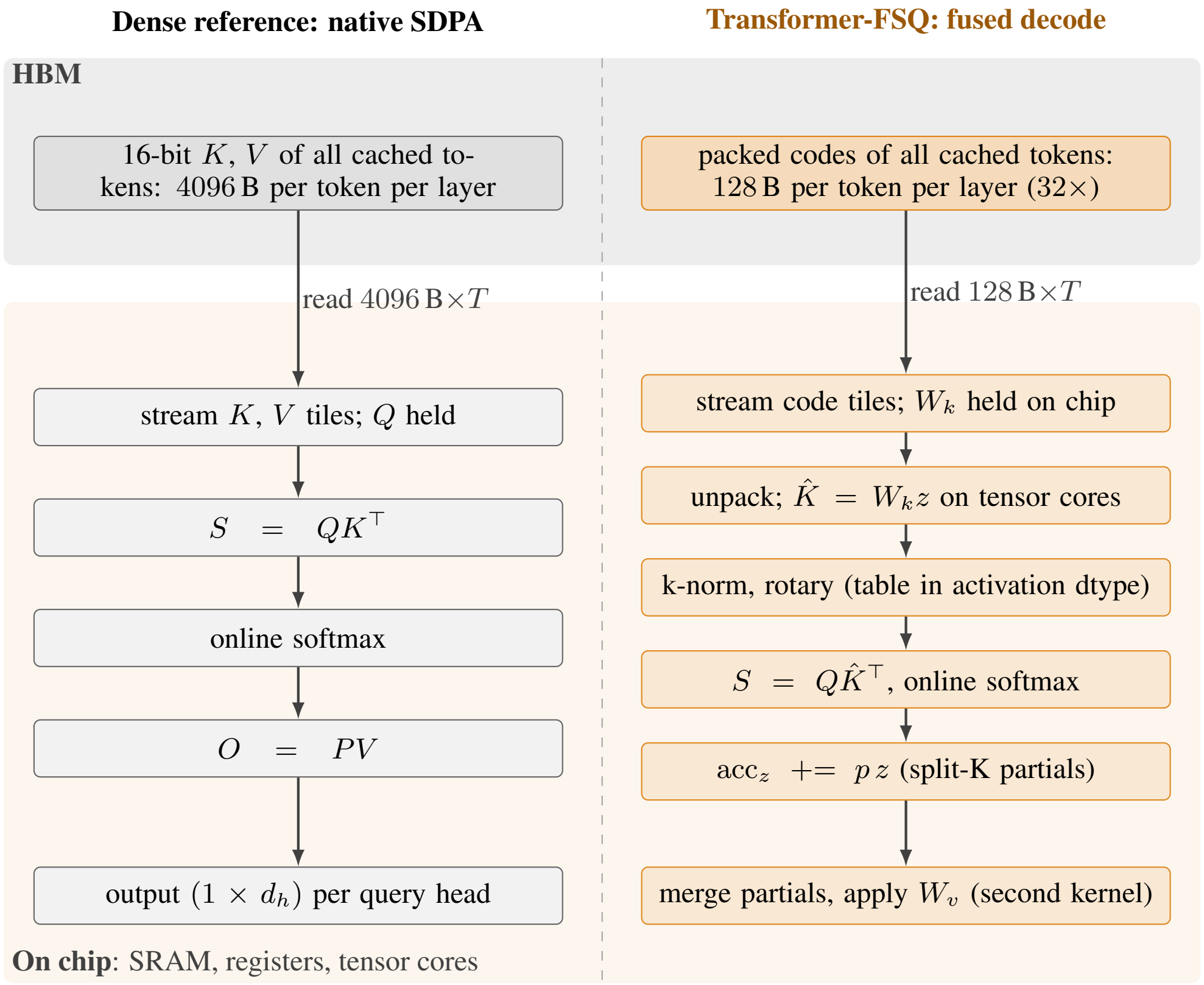
\begin{figure}[t]
\centering
\begin{tikzpicture}[
    kbox/.style={rectangle, draw=black!60, fill=black!5, rounded corners=2pt, font=\small, minimum height=0.62cm, inner sep=2pt, text centered, text width=5.6cm},
    kfsq/.style={rectangle, draw=cFSQ!90, fill=cFSQ!18, rounded corners=2pt, font=\small, minimum height=0.62cm, inner sep=2pt, text centered, text width=5.6cm},
    kmem/.style={rectangle, draw=black!60, fill=black!12, rounded corners=2pt, font=\small, minimum height=0.8cm, inner sep=2pt, text centered, text width=5.6cm},
    kmemf/.style={rectangle, draw=cFSQ!95, fill=cFSQ!28, rounded corners=2pt, font=\small, minimum height=0.8cm, inner sep=2pt, text centered, text width=5.6cm},
    karr/.style={->, >=latex, draw=black!75, line width=0.8pt},
    klab/.style={font=\small, inner sep=1pt, text=black!75},
  ]
  \path[fill=black!7, draw=none, rounded corners=3pt] (-6.5,0.9) rectangle (6.5,-1.35);
  \path[fill=cFSQ!7, draw=none, rounded corners=3pt] (-6.5,-1.75) rectangle (6.5,-9.15);
  \node[klab, anchor=north west] at (-6.45,0.88) {\textbf{HBM}};
  \node[klab, anchor=south west] at (-6.45,-9.12) {\textbf{On chip}: SRAM, registers, tensor cores};
  \draw[black!35, dashed] (0,0.9) -- (0,-9.15);
  \node[font=\small\bfseries] at (-3.3,1.3) {Dense reference: native SDPA};
  \node[font=\small\bfseries, text=cFSQ!70!black] at (3.3,1.3) {Transformer-FSQ: fused decode};
  \node[kmem] (dm) at (-3.3,-0.35) {16-bit $K$, $V$ of all cached tokens: $\byDense$\,B per token per layer};
  \node[kbox] (d1) at (-3.3,-3.0) {stream $K$, $V$ tiles; $Q$ held};
  \node[kbox] (d2) at (-3.3,-4.2) {$S = Q K^{\top}$};
  \node[kbox] (d3) at (-3.3,-5.4) {online softmax};
  \node[kbox] (d4) at (-3.3,-6.6) {$O = P V$};
  \node[kbox] (d5) at (-3.3,-8.2) {output $(1\times d_h)$ per query head};
  \draw[karr] (dm) -- node[klab, right]{read $\byDense$\,B${\times}T$} (d1);
  \draw[karr] (d1) -- (d2); \draw[karr] (d2) -- (d3); \draw[karr] (d3) -- (d4); \draw[karr] (d4) -- (d5);
  \node[kmemf] (fm) at (3.3,-0.35) {packed codes of all cached tokens: $\byFsqThirtytwo$\,B per token per layer ($32\times$)};
  \node[kfsq] (f1) at (3.3,-2.85) {stream code tiles; $W_k$ held on chip};
  \node[kfsq] (f2) at (3.3,-3.85) {unpack; $\hat K = W_k z$ on tensor cores};
  \node[kfsq] (f3) at (3.3,-4.85) {k-norm, rotary (table in activation dtype)};
  \node[kfsq] (f4) at (3.3,-5.85) {$S = Q \hat K^{\top}$, online softmax};
  \node[kfsq] (f5) at (3.3,-6.85) {$\mathrm{acc}_z \mathrel{+}= p\,z$ (split-K partials)};
  \node[kfsq] (f6) at (3.3,-8.2) {merge partials, apply $W_v$ (second kernel)};
  \draw[karr] (fm) -- node[klab, right]{read $\byFsqThirtytwo$\,B${\times}T$} (f1);
  \draw[karr] (f1) -- (f2); \draw[karr] (f2) -- (f3); \draw[karr] (f3) -- (f4); \draw[karr] (f4) -- (f5); \draw[karr] (f5) -- (f6);
\end{tikzpicture}
\caption{\textbf{One decoding step of one layer: native SDPA (left) and the Transformer-FSQ kernel (right).} Top band: GPU memory (HBM); everything below runs on chip. Native SDPA streams $\byDense$\,B of 16-bit keys and values per cached token. At $\rungThirtytwo\times$, the kernel streams $\byFsqThirtytwo$\,B of packed codes per token, reconstructs keys on the tensor cores and accumulates attention-weighted codes; $W_v$ is applied once, after the blocks are merged.}
\label{fig:kernel-dataflow}
\end{figure}

\subsection{The decode kernel}
\label{app:kernel}

A compressed cache saves time only if attention reads it without first expanding it in memory. We implement the decoding step of Figure~\ref{fig:fsq-decode-main} as a fused Triton kernel \citep{tillet2019triton}, \texttt{fsq\_rope\_decode\_mma}. Figure~\ref{fig:kernel-dataflow} compares its data flow with that of native SDPA decoding on the uncompressed cache. For each key-value head, the kernel splits the cached positions into blocks and, for each block,
\begin{enumerate}
\item reads the packed codes from GPU memory and unpacks them into registers;
\item reconstructs the keys $W_k z_t$ with a matrix multiplication on the tensor cores;
\item applies the key normalisation and the rotary embedding in registers, using a rotary table in the activation precision;
\item computes the scores of the query heads that share the key-value head, and updates a running softmax (elementwise arithmetic, not tensor cores);
\item adds the attention-weighted codes $a_t z_t$ to an accumulator, as Equation~\ref{eq:value-fold} allows.
\end{enumerate}
Blocks of positions run in parallel, each producing a partial accumulator and softmax statistics. A second, small kernel merges the partials and applies $W_v$ once; it rounds to the activation precision only at its output. Neither kernel writes a reconstructed key or value to GPU memory. Per cached token, the cache therefore stores $c_{\text{FSQ}}$ bytes where the uncompressed cache stores $c_{\text{full}}$, in exchange for the key reconstruction in step 2. Every key-value head's program streams the same shared codes, so the bytes issued per token exceed $c_{\text{FSQ}}$ unless repeated reads hit in the L2 cache; the dataflow figure shows the stored representation, not measured traffic.

\paragraph{Kernel checks.} Every timed or scored run declares the attention kernel its method should use, and an observation inside each attention call records the kernel that executed. Throughput runs abort on any mismatch, for example when a kernel rejects its inputs and a slower reference path runs instead; such rows are discarded. KIVI keeps its most recent tokens in full precision; on steps where its quantised part is still empty, no kernel is expected, and the log records this. Quality runs record the kernel without aborting; Palu's quality scores come from its reference implementation, which computes the same arithmetic as its kernel. Baselines use their authors' kernels where available; PyramidKV, which has none, runs in PyTorch, and TransMLA runs on our latent-attention kernel.

\subsection{Decode throughput protocol}
\label{app:decode}

The throughput campaign's grid, statistics and exclusion rules were fixed before it ran. Each host has $\tpGpus$ NVIDIA L40S GPUs; one process runs on each, pinned to its own CPU core on the GPU's socket, and the host's one-minute load average is recorded before every timed row. A session times one group of (model, precision) pairs, so every ratio divides by a native-SDPA row of the same model, precision and session. Each cell is one (method, cache size, batch, context) combination, with batch sizes \tpBatches{} and $\tpContexts$ log-spaced contexts from $\tpContextLo$ to $\tpContextHi$ tokens; every cache is priced at $\tpPriceLen$ tokens. A cell discards $\tpWarmups$ warm-up repetitions and times $\tpReps$ repetitions of $\tpSteps$ decode steps, each bounded by device synchronisation, and reports their median. The native-SDPA row is timed before and after the method in the same process. A row starts only when the host load is below $\tpLoadGate$.

\paragraph{Calibration.} Running four processes at once could slow each through shared host resources. Before each session, a calibration stage times native SDPA on one GPU alone and beside $\tpSiblings$ GPUs running the heaviest cells of the grid, in at least $\tpCalibPairs$ interleaved pairs. The session proceeds only if the 95\% bootstrap interval of the ratio of the two lies inside $[\tpCalibBandLo,\tpCalibBandHi]$; otherwise it runs one GPU at a time.

\paragraph{Statistics.} A method's ratio at a cache size is the median over its cells of its throughput divided by the native-SDPA throughput of the same cell, with a 95\% percentile bootstrap over cells ($\tpBootstrapN$ resamples). We report the median over every cell both sides completed, and over the subset of cells with at least $\tpMinBatchTokens$ tokens across the batch, where reading the cache dominates the step. A cell in which native SDPA runs out of memory has no ratio; such cells give the context reach of RQ4 instead.

\paragraph{Per-cell results.} Kernels are checked as described in Appendix~\ref{app:kernel}. Figures~\ref{fig:decode-vs-comp-qwen}--\ref{fig:decode-vs-comp-qwen8b} give the measured throughput of every baseline in each (batch, context) cell, against the compression ratio it achieves.

\begin{figure*}[t]
    \centering
    \includegraphics[width=\textwidth]{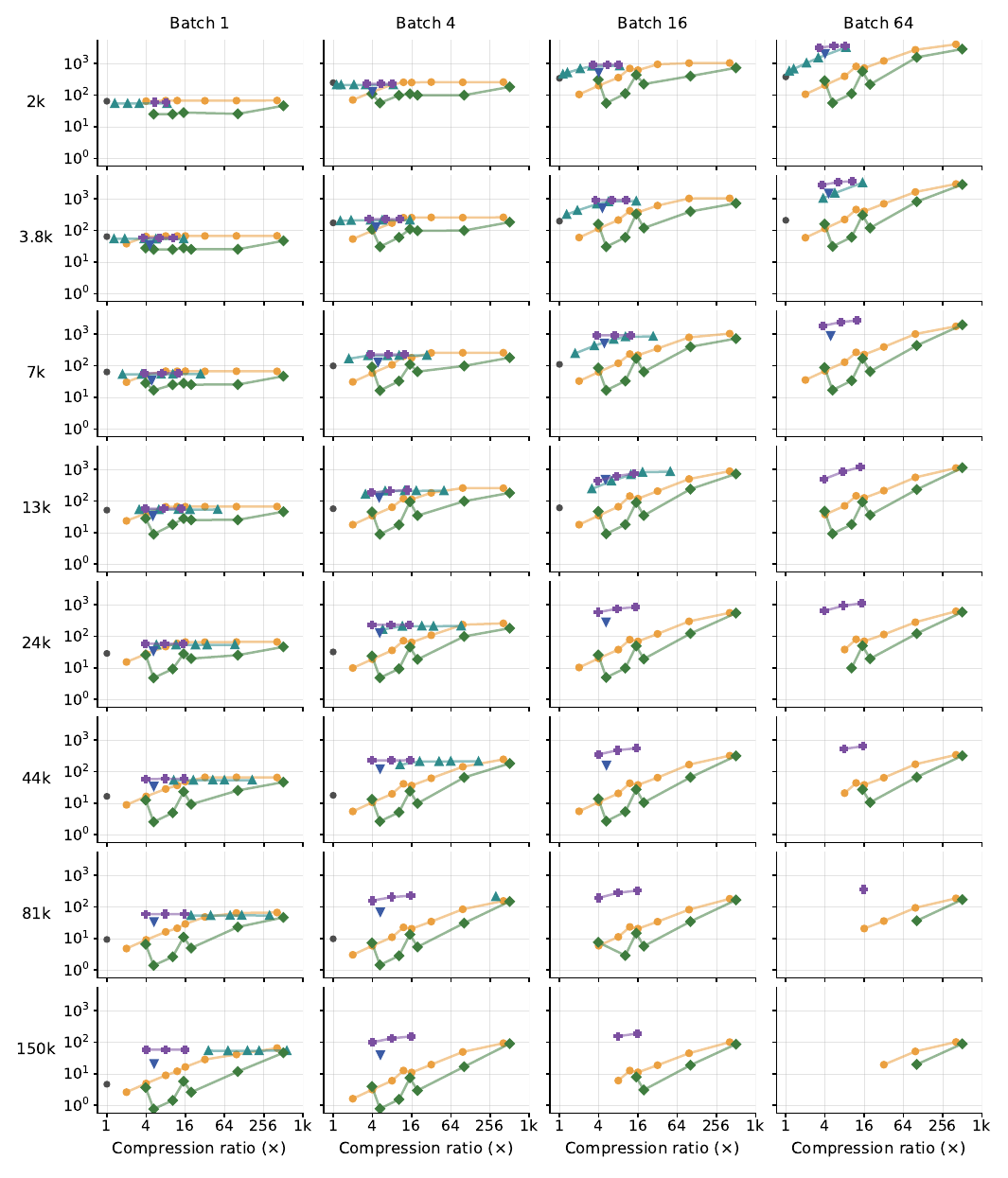}
    \vspace{-8pt}
    \caption{\textbf{Decode throughput against compression ratio, Qwen3-0.6B-Base:} median tokens/s on NVIDIA L40S; rows: sequence length, columns: batch size, both axes logarithmic. Grey: the uncompressed model with native SDPA, timed in the same process; orange circles: Transformer-FSQ at every ratio timed; teal upward triangles: PyramidKV; green diamonds: Palu; blue downward triangles: KIVI; purple plus signs: VecInfer. KIVI and VecInfer run in their native float16, the rest in bfloat16. Missing markers: out of memory or outside the method's grid.}
    \label{fig:decode-vs-comp-qwen}
\end{figure*}

\begin{figure*}[t]
    \centering
    \includegraphics[width=\textwidth]{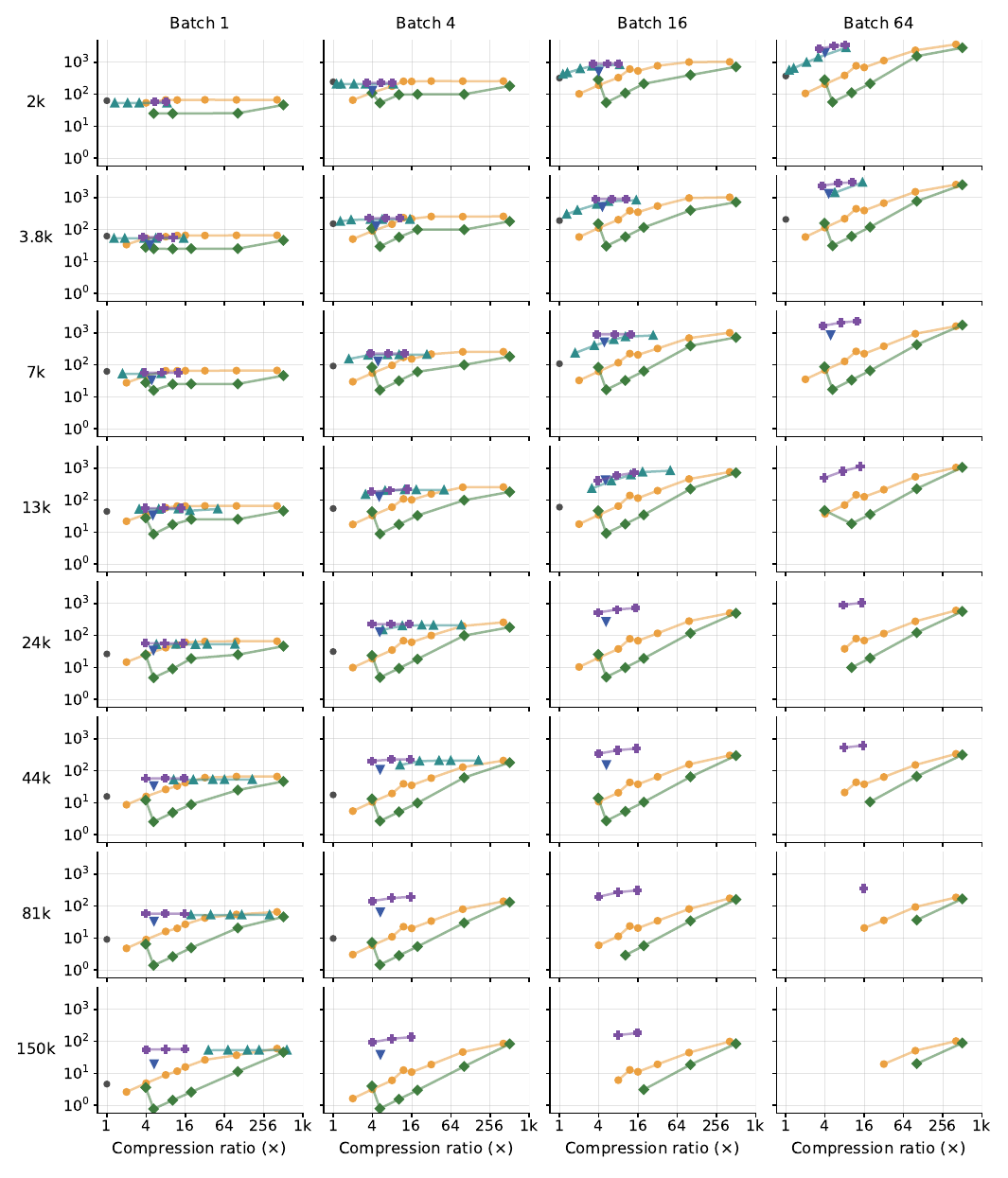}
    \vspace{-8pt}
    \caption{\textbf{Decode throughput against compression ratio, Qwen3-1.7B-Base;} layout as in Figure~\ref{fig:decode-vs-comp-qwen}.}
\end{figure*}

\begin{figure*}[t]
    \centering
    \includegraphics[width=\textwidth]{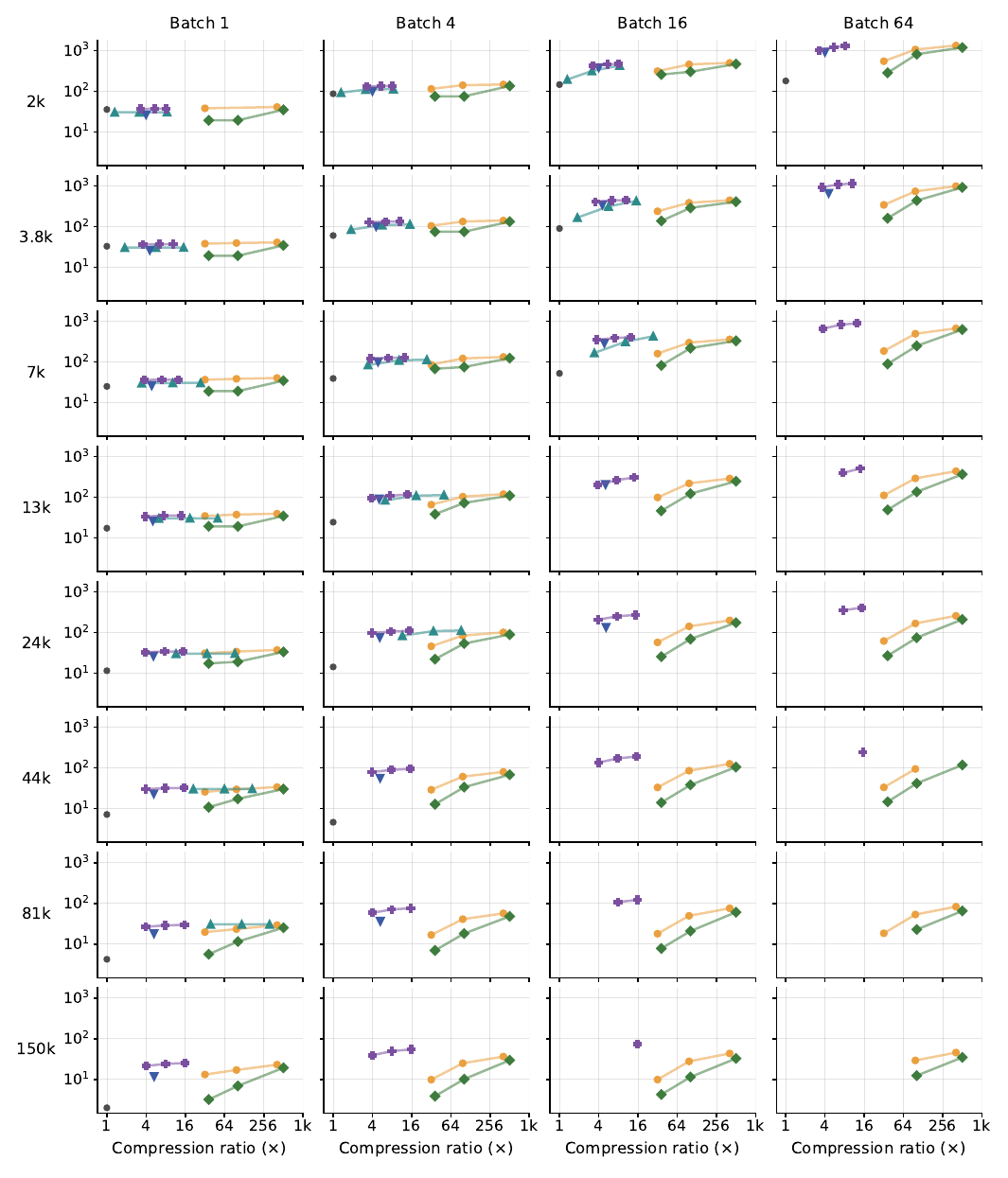}
    \vspace{-8pt}
    \caption{\textbf{Decode throughput against compression ratio, Qwen3-8B-Base;} layout as in Figure~\ref{fig:decode-vs-comp-qwen}.}
    \label{fig:decode-vs-comp-qwen8b}
\end{figure*}

\subsection{Empirical memory footprint}
\label{app:memory-check}

\begin{figure}[!ht]
    \centering
    \includegraphics{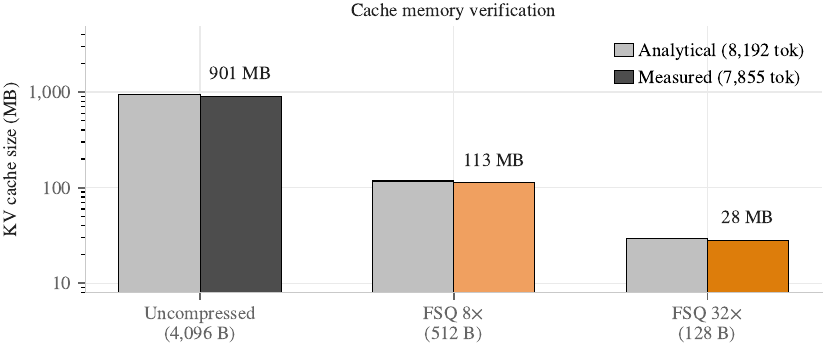}
    \caption{\textbf{Measured against analytical KV-cache memory} on Qwen3-0.6B-Base in bfloat16 at $8{,}192$ context ($\memPromptTokens$ prompt tokens). Grey: the analytical byte count (\S\ref{sec:budget}) at $8{,}192$ tokens; coloured: the buffer allocated after the prompt, which equals the analytical bytes per token times the prompt's tokens exactly.}
    \label{fig:measured-memory-breakdown}
\end{figure}

To check the byte accounting against what is allocated, we profile resident GPU memory on Qwen3-0.6B-Base in bfloat16 at an $8{,}192$-token context window ($\memPromptTokens$ prompt tokens in practice; Figure~\ref{fig:measured-memory-breakdown}). Across configurations, the physical buffer allocated for the KV cache equals the analytic formula, tokens $\times$ layers $\times$ bytes per token per layer, exactly. At $128$\,B per token per layer ($32\times$ compression), the measured cache buffer requires exactly $\memFsqThirtytwoBytes$\,bytes ($\memFsqThirtytwoMB$\,MB, or $\memFsqThirtytwoLayerMB$\,MB per layer for $28$ layers; $\memFsqThirtytwoFullMB$\,MB at full $8{,}192$ capacity). For FSQ at $512$\,B per token per layer ($8\times$ compression), measured cache occupancy is $\memFsqEightBytes$\,bytes ($\memFsqEightMB$\,MB; $\memFsqEightFullMB$\,MB at full window). The uncompressed cache holds $\memDenseBytes$\,bytes ($\memDenseMB$\,MB; $\memDenseFullMB$\,MB at full window).

\section{Ablations}
\label{app:ablations}

\looseness=-1 The ablations below test our architectural choices by validation bits per byte (BPB), long-context retrieval and decode throughput. Spreads report sample standard deviations over evaluated seeds.

\subsection{MLA-FSQ}

MLA-FSQ combines discrete FSQ codes for key-value content with Multi-Head Latent Attention's separate rotary key $k^r$, cached in half precision and shared across heads (Figure~\ref{fig:coded_kv_variants}). On Qwen3-0.6B-Base, shared query-key normalisation causes validation BPB to spike from $0.75$ to $5.5$, as large rotary norms ($121.8$ RMS vs $19.6$ content) suppress discrete representations. The half-precision rotary key and the per-head key norms take a fixed share of every token's bytes, and the deepest configuration we trained with query-key normalisation is $32\times$. We therefore keep the rotary embedding on reconstructed keys.

\subsection{FSQ-pre and positional encodings}

FSQ-pre evaluates applying RoPE to the residual stream before the encoder (Figure~\ref{fig:coded_kv_variants}), rotating activations as $C/d_h$ blocks of width $d_h$ so attention reads cached codes without reconstructing keys (\S\ref{sec:position}). In this topology, query-key normalisation breaks relative-position preservation, inducing severe logit errors in a 12-layer, 768-wide model trained from scratch ($0.60 \pm 0.08$ relative error vs $3 \times 10^{-7}$ without normalisation). We therefore keep the bottleneck after the projection, as in \S\ref{sec:operator}.

\subsection{Discreteness and coordinate allocations} \label{app:fsq-levels}

Table~\ref{tab:discreteness_appendix} compares, at two equal cache budgets on Qwen3-0.6B-Base, Transformer-FSQ (16 levels per coordinate) against a trained half-precision latent at the same byte budget, a wider continuously trained latent quantised to 4 bits post hoc with FSQ's grid or with KIVI's block scheme, a latent trained through a min--max 4-bit grid, and Palu and TransMLA with 4-bit latents, on RULER at $8{,}192$ tokens. Transformer-FSQ scores above every alternative at both sizes. Table~\ref{tab:fsq-levels} isolates levels per coordinate: at fixed cache memory, 16 levels per coordinate score higher than 4 or 256.

\begin{table}[t]
\centering
\caption{\textbf{Cache width, precision and when quantisation is applied.} RULER at $8{,}192$ tokens on Qwen3-0.6B-Base (three seeds, $0.25$ tokens per parameter); the uncompressed cache ($4{,}096$\,B per token per layer) scores $0.718 \pm 0.023$. Scores: mean $\pm$ 95\% bootstrap half-width; bold: best. Bytes: actual storage per token per layer, metadata included; Width: bottleneck dimension. \textit{Trained FSQ} rounds during training (straight-through estimator); \textit{post-hoc} FSQ and KIVI quantise a continuously trained latent only at evaluation. KIVI adds fp16 scales and offsets ($25\%$ over budget); Palu stores 4-bit SVD codes with scales; TransMLA stores 4-bit content codes plus 16-bit rotary keys.}
\label{tab:discreteness_appendix}
\vspace{2pt}
\small
\setlength{\tabcolsep}{8pt}
\begin{tabular}{@{}l r r r@{}}
\toprule
\textbf{Compression / quantisation} &
\textbf{Width} &
\textbf{Bytes} &
\textbf{RULER} \\
\midrule
\multicolumn{4}{l}{\textit{Nominal budget: 144 B/token/layer (28.4$\times$)}} \\[2pt]
Continuous latent & 72 & 144 & $ 0.009 \pm 0.008 $ \\
Latent + post-hoc FSQ & 288 & 144 & $ 0.175 \pm 0.030 $ \\
Transformer-FSQ (ours) & 288 & 144 & $ \mathbf{0.232 \pm 0.022} $ \\
Latent + trained min--max 4-bit & 224 & 140 & $ 0.187 \pm 0.020 $ \\
Latent + post-hoc KIVI & 288 & 180 & $ 0.206 \pm 0.021 $ \\
Palu, 4-bit & 144 & 144 & $ 0.000 \pm 0.000 $ \\
TransMLA, 4-bit & 64 & 144 & $ 0.001 \pm 0.001 $ \\
\midrule
\multicolumn{4}{l}{\textit{Nominal budget: 80 B/token/layer (51.2$\times$)}} \\[2pt]
Continuous latent & 40 & 80 & $ 0.001 \pm 0.001 $ \\
Latent + post-hoc FSQ & 160 & 80 & $ 0.136 \pm 0.014 $ \\
Transformer-FSQ (ours) & 160 & 80 & $ \mathbf{0.177 \pm 0.017} $ \\
Latent + trained min--max 4-bit & 128 & 80 & $ 0.151 \pm 0.021 $ \\
Latent + post-hoc KIVI & 160 & 100 & $ 0.168 \pm 0.014 $ \\
Palu, 4-bit & 80 & 80 & $ 0.000 \pm 0.000 $ \\
TransMLA, 4-bit & 32 & 76 & $ 0.000 \pm 0.000 $ \\
\bottomrule
\end{tabular}
\end{table}

\begin{table}[t]
\centering
\caption{\textbf{Levels per coordinate:} RULER at $8{,}192$ tokens on Qwen3-0.6B-Base ($0.25$ finetuning tokens per parameter) for FSQ with 4, 16 and 256 levels at $144$ and $80$\,B per token per layer; mean $\pm$ 95\% bootstrap half-width over three seeds and items. 16 levels score highest at both sizes.}
\label{tab:fsq-levels}
\vspace{4pt}
\small
\begin{tabular}{@{}l rrl@{}}
\toprule
\textbf{Cache} & \textbf{B/token/layer} & \textbf{Ratio} & \textbf{RULER ($8{,}192$)} \\
\midrule
FSQ, 4 levels & 144 & 28$\times$ & $0.178 \pm 0.013$ \\
FSQ, 16 levels & 144 & 28$\times$ & $0.232 \pm 0.022$ \\
FSQ, 256 levels & 144 & 28$\times$ & $0.161 \pm 0.011$ \\
\midrule
FSQ, 4 levels & 80 & 51$\times$ & $0.104 \pm 0.013$ \\
FSQ, 16 levels & 80 & 51$\times$ & $0.177 \pm 0.017$ \\
FSQ, 256 levels & 80 & 51$\times$ & $0.035 \pm 0.014$ \\
\bottomrule
\end{tabular}
\end{table}

\subsection{Straight-through estimation vs continuous relaxations}

When optimising discrete bottlenecks, alternatives to the straight-through estimator (STE) include continuous relaxations such as Gumbel-softmax \citep{jang2017categorical,maddison2017concrete} or codebook exponential moving average (EMA) updates \citep{oord2017neural,scannell2025discrete}. In our architecture, FSQ projects activations onto a fixed, implicit Cartesian grid by rounding, with the straight-through estimator taking $\partial \operatorname{round}(z)/\partial z = 1$ on the backward pass. Because the grid is fixed, there is no codebook to collapse and no commitment loss. Relaxations would add a temperature schedule across hundreds of coordinates, while STE trained stably with no extra hyperparameters; we did not evaluate the relaxations.

\subsection{Joint end-to-end training vs layerwise conversion}

An alternative to joint end-to-end training of all discrete bottlenecks is sequential layerwise conversion, progressively freezing earlier layers and training each bottleneck in isolation. While layerwise training avoids backpropagating through multiple STE rounding steps simultaneously, joint end-to-end training allows subsequent layers to adapt directly to the discrete quantisation errors of earlier representations. We use joint training throughout; layerwise conversion was not evaluated.

\subsection{Code utilisation and entropy coding}\label{app:entropy}

We measure the empirical bit-rate of the codes across two million held-out tokens per configuration: coordinates carry $79$--$83\%$ of available 4-bit capacity on Qwen3-0.6B-Base, with fewer than $0.3\%$ of channels inactive even at $4{,}096$ dimensions. Entropy coding is lossless, so it would change the cache size, not accuracy.

An ideal order-0 entropy coder would shrink the cache by $17$--$21\%$ on Qwen3-0.6B-Base. We implemented a per-channel static Huffman coder over blocks of 128--512 tokens; net of per-block offset tables and byte alignment ($\approx 3.8\%$ overhead), savings reached $12.9\%$ at extreme compression on Qwen3-0.6B-Base. However, variable-length decoding ran $16.6\times$--$21.2\times$ slower than our fused kernel (measured on a 12-layer, 768-wide model trained from scratch) and makes the cache size input-dependent, so we store codes packed.

\subsection{Nonlinear encoders}

We evaluated replacing the linear encoder with a rank-128 residual MLP ($z = W_{\text{down}} x + W_2\,\mathrm{SiLU}(W_1 x)$, with $W_2$ initialised to zero) on Qwen3-0.6B-Base (three seeds). The nonlinear encoder degraded validation loss across all cache budgets, increasing validation bits per byte from $1.000 \pm 0.001$ to $1.015 \pm 0.002$ at $128$\,B per token per layer ($32\times$). In a one-seed probe on a small model, a residual squared-ReLU encoder likewise raised validation BPB after a short finetune from conversion ($3.466$ vs $3.337$). We therefore retain the linear encoder, which also eliminates non-linear projection overhead during prefill.

\subsection{Per-head codes}

\looseness=-1 Grouping codes by key-value head allows each head to decode its own slice independently. We evaluated block-diagonal key and value decoders where each of Qwen3-0.6B-Base's 8 heads reads $1/8$ of the code (64, 32 and 10 coordinates per head, so $256$, $128$ and $40$\,B, against the shared code's $256$, $128$ and $42$\,B; the smallest per-head code has 80 coordinates in total, the shared one 84). Per-head codes are worse at every cache size, increasing validation bits per byte from $0.797 \pm 0.001$ to $0.911 \pm 0.001$ at $16\times$, $0.879 \pm 0.002$ to $1.109 \pm 0.003$ at $32\times$, and $1.173 \pm 0.003$ to $1.718 \pm 0.008$ at $97.5\times$. A shared code pools what the heads have in common, so we keep it.

\subsection{Recent-token window on Transformer-FSQ}

\looseness=-1 Keeping the newest 128 tokens and the first four (sink) tokens as unrounded latents beside Transformer-FSQ's codes shifts achieved memory: the nominal $8\times$ target ($512$\,B per token per layer) expands to $578$\,B ($7.1\times$ achieved) on Qwen3-0.6B-Base, and $32\times$ expands to $28.4\times$ on 8B. At nearly matched achieved ratios on RULER (Tables~\ref{tab:qwen-ruler-breakdown} and~\ref{tab:qwen8b-ruler-breakdown}), windowed FSQ scores $\rulerQwenPointsixFsqWindowBfivehundredseventyeightCtxeightkBootMean$ at $7.1\times$ vs plain FSQ's $\rulerQwenPointsixFsqEightxCtxeightkBootMean$ at $8\times$ on 0.6B, and $\rulerQwenEightBFsqWindowBonehundredfortyfourCtxeightkBootMean$ vs $\rulerQwenEightBFsqThirtytwoxCtxeightkBootMean$ on 8B. Across ratios from $1.8\times$ to $372.4\times$, windowed variants lie on the same score-against-compression curve as plain Transformer-FSQ once the window's bytes are counted (Tables~\ref{tab:qwen-ruler-breakdown}--\ref{tab:qwen8b-ruler-breakdown}), so at $8{,}192$ tokens the window neither helps nor hurts.

\section{Architecture diagrams}
\label{app:diagrams}

\subsection{The FSQ bottleneck}
\label{app:fsq-bottleneck}
The quantiser and its packing are defined in Eq.~\ref{eq:fsq-exact} (\S\ref{sec:background_quantization}).

\looseness=-1 Figure~\ref{fig:fsq-decode-main} contrasts one decoding step of an unmodified transformer with one step of Transformer-FSQ, and Figure~\ref{fig:coded_kv_variants} shows the two variants of Appendix~\ref{app:ablations} that change how the cache is read: FSQ-pre and MLA-FSQ. The shaded region of each diagram marks the work that grows with context length.

\tikzset{
  ckvbox/.style={rectangle, draw=black!60, fill=black!5, rounded corners=2pt, font=\small, minimum height=0.6cm, inner sep=2pt, text centered},
  ckvfsq/.style={rectangle, draw=cFSQ!90, fill=cFSQ!18, rounded corners=2pt, font=\small, minimum height=0.6cm, inner sep=2pt, text centered},
  ckvcache/.style={rectangle, draw=cFSQ!95, fill=cFSQ!25, rounded corners=2pt, font=\small, minimum height=0.6cm, inner sep=2pt, text centered},
  ckvstdcache/.style={rectangle, draw=black!60, fill=black!10, rounded corners=2pt, font=\small, minimum height=0.6cm, inner sep=2pt, text centered},
  ckvarr/.style={->, >=latex, draw=black!75, line width=0.8pt},
  ckvop/.style={circle, draw=black!60, fill=white, inner sep=0pt, minimum size=4.5mm, font=\small},
  ckvlab/.style={font=\small, inner sep=1pt},
  ckvplate/.style={draw=black!25, fill=black!4, rounded corners=4pt, inner sep=4pt},
}

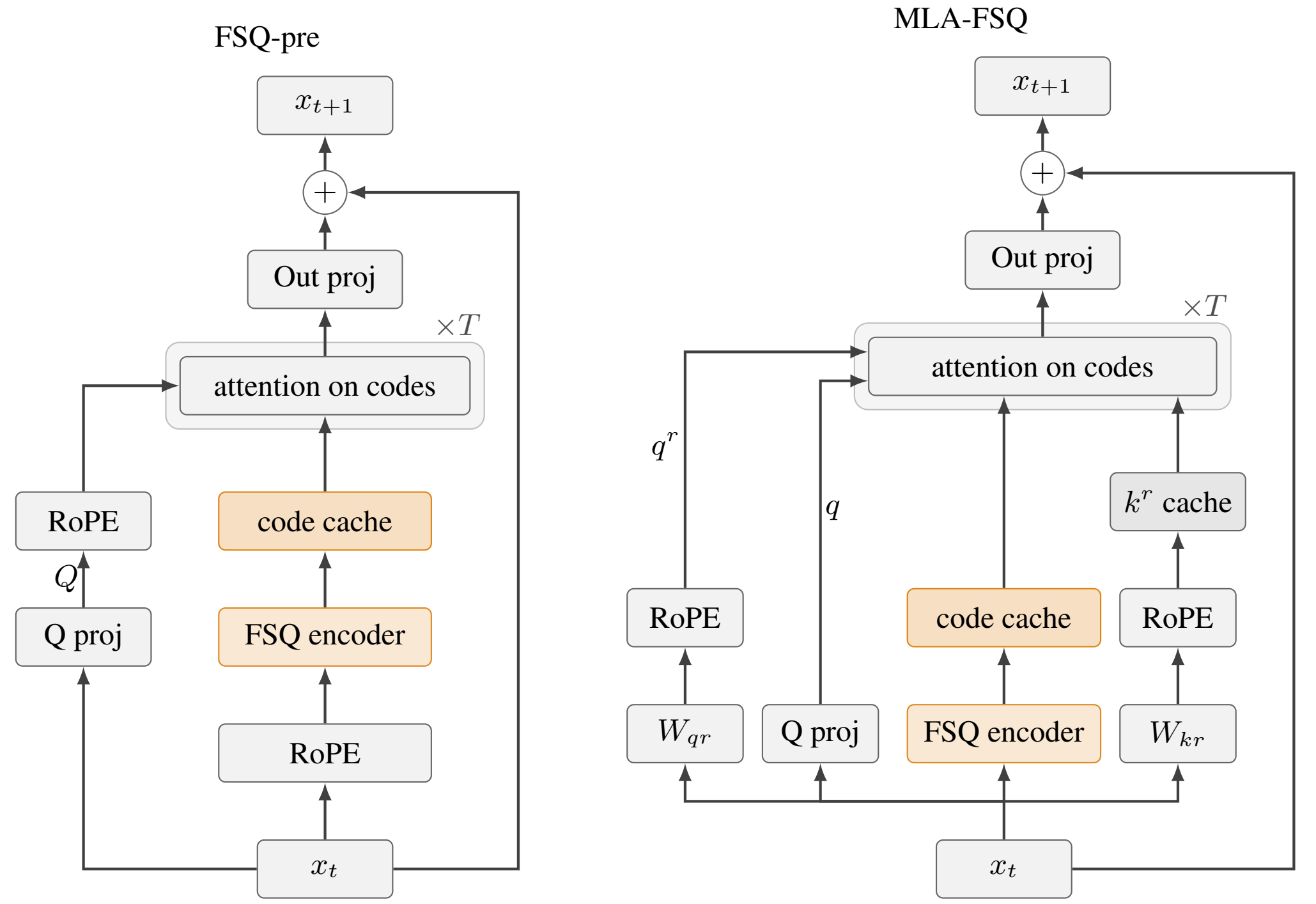
\begin{figure*}[t]
\centering
\begin{minipage}[b]{0.40\linewidth}
\centering
{\small FSQ-pre}\\[0.2cm]
\begin{tikzpicture}[box/.style={ckvbox}, fsqbox/.style={ckvfsq}, cachebox/.style={ckvcache}, arr/.style={ckvarr},
                    op/.style={ckvop}, lab/.style={ckvlab}]
  \node[box, minimum width=1.4cm] (x) at (0,0) {$x_t$};
  \node[box, minimum width=2.2cm] (rope) at (0,1.2) {RoPE};
  \node[fsqbox, minimum width=2.2cm] (enc) at (0,2.4) {FSQ encoder};
  \node[cachebox, minimum width=2.2cm] (code) at (0,3.6) {code cache};
  \node[box, minimum width=3.0cm] (attn) at (0,5.0) {attention on codes};
  \node[box, minimum width=1.6cm] (op) at (0,6.1) {Out proj};
  \node[op] (plus) at (0,7.0) {$+$};
  \node[box, minimum width=1.4cm] (out) at (0,7.9) {$x_{t+1}$};
  \node[box, minimum width=1.4cm] (qp) at (-2.5,2.4) {Q proj};
  \node[box, minimum width=1.4cm] (qr) at (-2.5,3.6) {RoPE};
  \draw[arr] (x) -- (rope);
  \draw[arr] (rope) -- (enc);
  \draw[arr] (enc) -- (code);
  \draw[arr] (code) -- (attn);
  \draw[arr] (x.west) -- (-2.5,0) -- (qp.south);
  \draw[arr] (qp) -- node[lab, left]{$Q$} (qr);
  \draw[arr] (qr.north) |- (attn.west);
  \draw[arr] (attn) -- (op);
  \draw[arr] (op) -- (plus);
  \draw[arr] (plus) -- (out);
  \draw[arr] (x.east) -- (2.0,0) |- (plus.east);
  \begin{pgfonlayer}{background}
    \node[ckvplate, fit=(attn)] (ppre) {};
  \end{pgfonlayer}
  \node[font=\small\sffamily\bfseries, text=black!70, anchor=south east, inner sep=1pt] at (ppre.north east) {${\times}T$};
\end{tikzpicture}
\end{minipage}\hfill
\begin{minipage}[b]{0.56\linewidth}
\centering
{\small MLA-FSQ}\\[0.2cm]
\begin{tikzpicture}[box/.style={ckvbox}, fsqbox/.style={ckvfsq}, cachebox/.style={ckvcache}, stdcache/.style={ckvstdcache}, arr/.style={ckvarr},
                    op/.style={ckvop}, lab/.style={ckvlab}]
  \node[box, minimum width=1.4cm] (x) at (0,0) {$x_t$};
  \node[box, minimum width=1.2cm] (wqr) at (-3.3,1.4) {$W_{qr}$};
  \node[box, minimum width=1.2cm] (cq) at (-1.9,1.4) {Q proj};
  \node[fsqbox, minimum width=2.0cm] (enc) at (0,1.4) {FSQ encoder};
  \node[box, minimum width=1.2cm] (wkr) at (1.8,1.4) {$W_{kr}$};
  \node[box, minimum width=1.2cm] (rqr) at (-3.3,2.6) {RoPE};
  \node[cachebox, minimum width=2.0cm] (code) at (0,2.6) {code cache};
  \node[box, minimum width=1.2cm] (rkr) at (1.8,2.6) {RoPE};
  \node[stdcache, minimum width=1.4cm] (krc) at (1.8,3.8) {$k^r$ cache};
  \node[box, minimum width=3.6cm] (attn) at (0.4,5.2) {attention on codes};
  \node[box, minimum width=1.6cm] (op) at (0.4,6.3) {Out proj};
  \node[op] (plus) at (0.4,7.2) {$+$};
  \node[box, minimum width=1.4cm] (out) at (0.4,8.1) {$x_{t+1}$};
  \coordinate (bus) at (0,0.7);
  \draw[arr] (x.north) -- (bus) -| (wqr.south);
  \draw[arr] (bus) -| (cq.south);
  \draw[arr] (bus) -- (enc.south);
  \draw[arr] (bus) -| (wkr.south);
  \draw[arr] (wqr) -- (rqr);
  \draw[arr] (enc) -- (code);
  \draw[arr] (wkr) -- (rkr);
  \draw[arr] (rkr) -- (krc);
  \draw[arr] (code.north) -- (code.north |- attn.south);
  \draw[arr] (krc.north) -- (krc.north |- attn.south);
  \draw[arr] (cq.north) |- node[lab, pos=0.3, right]{$q$} ([yshift=-1.5mm]attn.west);
  \draw[arr] (rqr.north) |- node[lab, pos=0.3, left]{$q^r$} ([yshift=1.5mm]attn.west);
  \draw[arr] (attn) -- (op);
  \draw[arr] (op) -- (plus);
  \draw[arr] (plus) -- (out);
  \draw[arr] (x.east) -- (3.0,0) |- (plus.east);
  \begin{pgfonlayer}{background}
    \node[ckvplate, fit=(attn)] (pmla) {};
  \end{pgfonlayer}
  \node[font=\small\sffamily\bfseries, text=black!70, anchor=south east, inner sep=1pt] at (pmla.north east) {${\times}T$};
\end{tikzpicture}
\end{minipage}
\caption{\looseness=-1 \textbf{Alternative decoding topologies: FSQ-pre (left) and MLA-FSQ (right).} FSQ-pre applies RoPE before the encoder, so codes carry position and keys need no per-step reconstruction. MLA-FSQ caches a half-precision rotary key $k^r$ beside the content codes and adds its query dot product to the logits. Shaded plates (${\times}T$) scale with context length $T$. Appendix~\ref{app:ablations} gives why we keep post-projection Transformer-FSQ.}
\label{fig:coded_kv_variants}
\end{figure*}

\section{Per-task results in full}
\label{app:longctx}

The tables below give the complete per-task RULER and LongBench results for every model.

\subsection{Evaluation benchmarks and notation}

\paragraph{Benchmarks.} We evaluate long-context retrieval and comprehension on two suites:
\begin{itemize}
    \item \textbf{RULER \citep{hsieh2024ruler}:} 13 synthetic retrieval tasks at $8{,}192$ tokens (50 items each): single-key retrieval (\texttt{niah\_single\_1}, \texttt{2}, \texttt{3}), multi-key retrieval (\texttt{niah\_multikey\_1}, \texttt{2}, \texttt{3}), multi-value retrieval (\texttt{niah\_multivalue}), multi-query retrieval (\texttt{niah\_multiquery}), variable tracking (\texttt{vt}), common/frequent words extraction (\texttt{cwe}, \texttt{fwe}), and QA (\texttt{qa\_1}, \texttt{qa\_2}).
    \item \textbf{LongBench \citep{bai2024longbench}:} Real-world multi-page document comprehension: \texttt{narrativeqa} (narrative question answering, F1), \texttt{gov\_report} (government report summarisation, ROUGE-L), and \texttt{trec} (few-shot in-context question classification, accuracy).
\end{itemize}

\paragraph{Table notation.}
In RULER tables (Tables~\ref{tab:qwen-ruler-breakdown}--\ref{tab:qwen8b-ruler-breakdown}), rows report the achieved compression ratio computed from the analytical byte budget of \S\ref{sec:budget} and the mean over seeds; standard deviations are omitted for compactness. Finetuned methods (Transformer-FSQ, Palu, TransMLA) use $0.25$ finetuning tokens per parameter; KIVI, PyramidKV, and VecInfer are training-free.
In LongBench tables (Tables~\ref{tab:qwen-longbench-breakdown}--\ref{tab:qwen8b-longbench-breakdown}), columns report achieved bytes per token per layer (\textbf{B/tok}), achieved compression ratio (\textbf{Ratio}), evaluated seeds, scored examples ($n$), and task scores. $^{\S}$ marks dynamic-cache methods (KIVI, PyramidKV, VecInfer) whose cache footprint depends on context length.

\clearpage
\subsection{Per-task long-context retrieval: RULER}

\looseness=-1 Tables~\ref{tab:qwen-ruler-breakdown}--\ref{tab:qwen8b-ruler-breakdown} give the per-task RULER breakdown behind Figure~\ref{fig:ruler}, Figure~\ref{fig:ruler-relative}, and Appendix~\ref{app:longctx-main}; every RULER number in the main text uses this protocol (Appendix~\ref{app:evaluation}).

\begin{table}[H]
\centering
\caption{Per-task RULER on Qwen3-0.6B-Base (13 tasks of 50 items, $8{,}192$ tokens, bfloat16). Finetuned methods use $0.25$ tokens per parameter; KIVI, PyramidKV and VecInfer are training-free. Rows give achieved compression ratios; scores average complete seeds. Shading encodes the task score.}
\label{tab:qwen-ruler-breakdown}
\small
\setlength{\tabcolsep}{1.2pt}
\renewcommand{\arraystretch}{0.98}
\begin{tabular}{lccccccccccccccc}
\toprule
\textbf{Method} & \rotatebox{90}{\textbf{seeds}} & \rotatebox{90}{\shortstack[l]{\textbf{niah\_}\\\textbf{single\_1}}} & \rotatebox{90}{\shortstack[l]{\textbf{niah\_}\\\textbf{single\_2}}} & \rotatebox{90}{\shortstack[l]{\textbf{niah\_}\\\textbf{single\_3}}} & \rotatebox{90}{\shortstack[l]{\textbf{niah\_}\\\textbf{multikey\_1}}} & \rotatebox{90}{\shortstack[l]{\textbf{niah\_}\\\textbf{multikey\_2}}} & \rotatebox{90}{\shortstack[l]{\textbf{niah\_}\\\textbf{multikey\_3}}} & \rotatebox{90}{\shortstack[l]{\textbf{niah\_}\\\textbf{multivalue}}} & \rotatebox{90}{\shortstack[l]{\textbf{niah\_}\\\textbf{multiquery}}} & \rotatebox{90}{\textbf{vt}} & \rotatebox{90}{\textbf{cwe}} & \rotatebox{90}{\textbf{fwe}} & \rotatebox{90}{\shortstack[l]{\textbf{qa\_}\\\textbf{1}}} & \rotatebox{90}{\shortstack[l]{\textbf{qa\_}\\\textbf{2}}} & \rotatebox{90}{\textbf{Mean}} \\
\midrule
Uncompressed & 1 & \cellcolor[HTML]{B2C1D1}1.00 & \cellcolor[HTML]{B5C3D2}0.96 & \cellcolor[HTML]{B4C2D1}0.98 & \cellcolor[HTML]{B9C6D5}0.90 & \cellcolor[HTML]{BECAD8}0.82 & \cellcolor[HTML]{DAE1E9}0.42 & \cellcolor[HTML]{BDC9D7}0.83 & \cellcolor[HTML]{BAC7D5}0.88 & \cellcolor[HTML]{BFCBD9}0.80 & \cellcolor[HTML]{E3E8EE}0.30 & \cellcolor[HTML]{D3DCE5}0.51 & \cellcolor[HTML]{CCD5E0}0.62 & \cellcolor[HTML]{E3E9EE}0.30 & \cellcolor[HTML]{C5D0DC}0.72 \\
\midrule
Transformer-FSQ (2.0$\times$) & 3 & \cellcolor[HTML]{B2C1D1}1.00 & \cellcolor[HTML]{B7C5D4}0.92 & \cellcolor[HTML]{B4C2D2}0.97 & \cellcolor[HTML]{C9D3DE}0.66 & \cellcolor[HTML]{DBE2E9}0.41 & \cellcolor[HTML]{FBFCFC}0.03 & \cellcolor[HTML]{C8D3DE}0.67 & \cellcolor[HTML]{D3DBE4}0.53 & \cellcolor[HTML]{C6D1DD}0.71 & \cellcolor[HTML]{E0E6EC}0.34 & \cellcolor[HTML]{DAE1E9}0.42 & \cellcolor[HTML]{D7DEE7}0.47 & \cellcolor[HTML]{E9EDF2}0.23 & \cellcolor[HTML]{D0D9E2}0.57 \\
Transformer-FSQ (4.0$\times$) & 3 & \cellcolor[HTML]{B2C1D1}1.00 & \cellcolor[HTML]{B9C6D5}0.89 & \cellcolor[HTML]{B4C2D2}0.97 & \cellcolor[HTML]{C7D2DD}0.69 & \cellcolor[HTML]{D9E0E8}0.43 & \cellcolor[HTML]{FBFCFC}0.03 & \cellcolor[HTML]{C7D2DD}0.68 & \cellcolor[HTML]{D2DAE4}0.54 & \cellcolor[HTML]{CCD6E0}0.61 & \cellcolor[HTML]{E2E8EE}0.31 & \cellcolor[HTML]{DAE1E9}0.42 & \cellcolor[HTML]{D2DBE4}0.53 & \cellcolor[HTML]{E8ECF1}0.25 & \cellcolor[HTML]{D0D9E2}0.57 \\
Transformer-FSQ (8.0$\times$) & 3 & \cellcolor[HTML]{B2C1D1}1.00 & \cellcolor[HTML]{BBC8D6}0.87 & \cellcolor[HTML]{B4C2D2}0.97 & \cellcolor[HTML]{C8D2DE}0.68 & \cellcolor[HTML]{E1E7ED}0.33 & \cellcolor[HTML]{FAFBFC}0.04 & \cellcolor[HTML]{C7D1DD}0.69 & \cellcolor[HTML]{D3DBE4}0.52 & \cellcolor[HTML]{CED7E2}0.59 & \cellcolor[HTML]{E3E8EE}0.31 & \cellcolor[HTML]{D8DFE7}0.45 & \cellcolor[HTML]{D2DAE4}0.53 & \cellcolor[HTML]{E6EBF0}0.27 & \cellcolor[HTML]{D0D9E3}0.56 \\
Transformer-FSQ (12.0$\times$) & 3 & \cellcolor[HTML]{B3C1D1}0.99 & \cellcolor[HTML]{BCC8D6}0.85 & \cellcolor[HTML]{C9D3DE}0.66 & \cellcolor[HTML]{DCE2EA}0.40 & \cellcolor[HTML]{F3F5F8}0.11 & \cellcolor[HTML]{FEFEFE}0.01 & \cellcolor[HTML]{E1E7ED}0.33 & \cellcolor[HTML]{DBE2E9}0.41 & \cellcolor[HTML]{EDF0F4}0.18 & \cellcolor[HTML]{E3E8EE}0.31 & \cellcolor[HTML]{E9EDF1}0.24 & \cellcolor[HTML]{DAE1E9}0.43 & \cellcolor[HTML]{EEF1F5}0.17 & \cellcolor[HTML]{DCE3EA}0.39 \\
Transformer-FSQ (16.0$\times$) & 3 & \cellcolor[HTML]{B3C1D1}0.99 & \cellcolor[HTML]{C7D2DD}0.69 & \cellcolor[HTML]{CDD6E1}0.61 & \cellcolor[HTML]{E7EBF0}0.26 & \cellcolor[HTML]{FBFCFD}0.03 & \cellcolor[HTML]{FFFFFF}0.00 & \cellcolor[HTML]{E4E9EF}0.29 & \cellcolor[HTML]{E5EAF0}0.27 & \cellcolor[HTML]{EDF1F4}0.18 & \cellcolor[HTML]{EAEEF2}0.22 & \cellcolor[HTML]{EAEEF3}0.21 & \cellcolor[HTML]{E2E8EE}0.31 & \cellcolor[HTML]{EFF2F6}0.15 & \cellcolor[HTML]{E2E7ED}0.32 \\
Transformer-FSQ (32.0$\times$) & 3 & \cellcolor[HTML]{B6C4D3}0.94 & \cellcolor[HTML]{D6DEE6}0.48 & \cellcolor[HTML]{E1E6ED}0.33 & \cellcolor[HTML]{E9EDF2}0.23 & \cellcolor[HTML]{FFFFFF}0.00 & \cellcolor[HTML]{FFFFFF}0.00 & \cellcolor[HTML]{EAEEF2}0.22 & \cellcolor[HTML]{E9EDF2}0.23 & \cellcolor[HTML]{FFFFFF}0.00 & \cellcolor[HTML]{F2F4F7}0.13 & \cellcolor[HTML]{FDFDFE}0.01 & \cellcolor[HTML]{ECF0F4}0.19 & \cellcolor[HTML]{F2F5F7}0.12 & \cellcolor[HTML]{EAEEF2}0.22 \\
Transformer-FSQ (97.5$\times$) & 3 & \cellcolor[HTML]{E7ECF1}0.25 & \cellcolor[HTML]{F5F7F9}0.09 & \cellcolor[HTML]{FFFFFF}0.00 & \cellcolor[HTML]{F7F9FA}0.07 & \cellcolor[HTML]{FFFFFF}0.00 & \cellcolor[HTML]{FFFFFF}0.00 & \cellcolor[HTML]{FBFCFD}0.03 & \cellcolor[HTML]{FBFCFD}0.03 & \cellcolor[HTML]{FFFFFF}0.00 & \cellcolor[HTML]{FFFFFF}0.00 & \cellcolor[HTML]{FFFFFF}0.00 & \cellcolor[HTML]{F9FAFB}0.05 & \cellcolor[HTML]{FCFDFD}0.02 & \cellcolor[HTML]{FAFBFC}0.04 \\
Transformer-FSQ (409.6$\times$) & 3 & \cellcolor[HTML]{FFFFFF}0.00 & \cellcolor[HTML]{FFFFFF}0.00 & \cellcolor[HTML]{FFFFFF}0.00 & \cellcolor[HTML]{FFFFFF}0.00 & \cellcolor[HTML]{FFFFFF}0.00 & \cellcolor[HTML]{FFFFFF}0.00 & \cellcolor[HTML]{FFFFFF}0.00 & \cellcolor[HTML]{FFFFFF}0.00 & \cellcolor[HTML]{FFFFFF}0.00 & \cellcolor[HTML]{FFFFFF}0.00 & \cellcolor[HTML]{FFFFFF}0.00 & \cellcolor[HTML]{FCFDFD}0.02 & \cellcolor[HTML]{FFFFFF}0.00 & \cellcolor[HTML]{FFFFFF}0.00 \\
FSQ + window (1.8$\times$) & 1 & \cellcolor[HTML]{B2C1D1}1.00 & \cellcolor[HTML]{BAC7D5}0.88 & \cellcolor[HTML]{B2C1D1}1.00 & \cellcolor[HTML]{C8D2DE}0.68 & \cellcolor[HTML]{DAE1E9}0.42 & \cellcolor[HTML]{FFFFFF}0.00 & \cellcolor[HTML]{CAD4DF}0.65 & \cellcolor[HTML]{D4DCE5}0.51 & \cellcolor[HTML]{D5DDE6}0.49 & \cellcolor[HTML]{E0E6EC}0.34 & \cellcolor[HTML]{D8DFE7}0.45 & \cellcolor[HTML]{D4DCE5}0.50 & \cellcolor[HTML]{E5EAEF}0.28 & \cellcolor[HTML]{D1D9E3}0.55 \\
FSQ + window (3.5$\times$) & 1 & \cellcolor[HTML]{B2C1D1}1.00 & \cellcolor[HTML]{BAC7D5}0.88 & \cellcolor[HTML]{B2C1D1}1.00 & \cellcolor[HTML]{C6D1DD}0.70 & \cellcolor[HTML]{E0E6EC}0.34 & \cellcolor[HTML]{FFFFFF}0.00 & \cellcolor[HTML]{CDD6E1}0.61 & \cellcolor[HTML]{C9D3DF}0.66 & \cellcolor[HTML]{D5DDE6}0.49 & \cellcolor[HTML]{E3E8EE}0.31 & \cellcolor[HTML]{D8DFE7}0.45 & \cellcolor[HTML]{D4DCE5}0.50 & \cellcolor[HTML]{E7EBF0}0.26 & \cellcolor[HTML]{D1D9E3}0.55 \\
FSQ + window (7.1$\times$) & 1 & \cellcolor[HTML]{B2C1D1}1.00 & \cellcolor[HTML]{B5C3D2}0.96 & \cellcolor[HTML]{B2C1D1}1.00 & \cellcolor[HTML]{C5D0DC}0.72 & \cellcolor[HTML]{E2E7ED}0.32 & \cellcolor[HTML]{FAFBFC}0.04 & \cellcolor[HTML]{C7D1DD}0.69 & \cellcolor[HTML]{CCD6E0}0.61 & \cellcolor[HTML]{D8DFE7}0.45 & \cellcolor[HTML]{E4E9EF}0.30 & \cellcolor[HTML]{DCE2EA}0.40 & \cellcolor[HTML]{CFD8E2}0.58 & \cellcolor[HTML]{E7EBF0}0.26 & \cellcolor[HTML]{D0D9E2}0.56 \\
FSQ + window (10.6$\times$) & 1 & \cellcolor[HTML]{B4C2D1}0.98 & \cellcolor[HTML]{BECAD8}0.82 & \cellcolor[HTML]{C2CEDA}0.76 & \cellcolor[HTML]{DDE4EB}0.38 & \cellcolor[HTML]{F2F5F7}0.12 & \cellcolor[HTML]{FFFFFF}0.00 & \cellcolor[HTML]{E7EBF0}0.26 & \cellcolor[HTML]{DFE5EC}0.35 & \cellcolor[HTML]{F8F9FB}0.06 & \cellcolor[HTML]{E2E7ED}0.32 & \cellcolor[HTML]{EAEEF2}0.22 & \cellcolor[HTML]{E0E6EC}0.34 & \cellcolor[HTML]{EFF2F5}0.16 & \cellcolor[HTML]{DEE4EB}0.37 \\
FSQ + window (14.2$\times$) & 1 & \cellcolor[HTML]{B2C1D1}1.00 & \cellcolor[HTML]{D3DBE4}0.52 & \cellcolor[HTML]{D0D9E3}0.56 & \cellcolor[HTML]{E5EAEF}0.28 & \cellcolor[HTML]{FCFDFD}0.02 & \cellcolor[HTML]{FFFFFF}0.00 & \cellcolor[HTML]{E9EDF2}0.23 & \cellcolor[HTML]{E9EDF2}0.23 & \cellcolor[HTML]{FEFEFE}0.01 & \cellcolor[HTML]{E7ECF1}0.25 & \cellcolor[HTML]{F4F6F8}0.11 & \cellcolor[HTML]{E3E9EE}0.30 & \cellcolor[HTML]{F1F3F6}0.14 & \cellcolor[HTML]{E5EAEF}0.28 \\
FSQ + window (28.4$\times$) & 1 & \cellcolor[HTML]{B6C4D3}0.94 & \cellcolor[HTML]{DCE2EA}0.40 & \cellcolor[HTML]{DDE4EB}0.38 & \cellcolor[HTML]{EAEEF2}0.22 & \cellcolor[HTML]{FFFFFF}0.00 & \cellcolor[HTML]{FFFFFF}0.00 & \cellcolor[HTML]{EAEEF2}0.22 & \cellcolor[HTML]{ECF0F4}0.19 & \cellcolor[HTML]{FDFEFE}0.01 & \cellcolor[HTML]{F0F3F6}0.15 & \cellcolor[HTML]{FCFDFD}0.02 & \cellcolor[HTML]{EAEEF2}0.22 & \cellcolor[HTML]{F2F5F7}0.12 & \cellcolor[HTML]{EAEEF2}0.22 \\
FSQ + window (87.1$\times$) & 1 & \cellcolor[HTML]{D6DEE6}0.48 & \cellcolor[HTML]{EDF0F4}0.18 & \cellcolor[HTML]{FFFFFF}0.00 & \cellcolor[HTML]{F2F5F7}0.12 & \cellcolor[HTML]{FFFFFF}0.00 & \cellcolor[HTML]{FFFFFF}0.00 & \cellcolor[HTML]{F9FAFB}0.05 & \cellcolor[HTML]{F9FAFB}0.05 & \cellcolor[HTML]{FFFFFF}0.00 & \cellcolor[HTML]{FFFFFF}0.00 & \cellcolor[HTML]{FFFFFF}0.00 & \cellcolor[HTML]{FCFDFD}0.02 & \cellcolor[HTML]{FFFFFF}0.00 & \cellcolor[HTML]{F7F9FA}0.07 \\
FSQ + window (372.4$\times$) & 1 & \cellcolor[HTML]{FFFFFF}0.00 & \cellcolor[HTML]{FFFFFF}0.00 & \cellcolor[HTML]{FFFFFF}0.00 & \cellcolor[HTML]{FFFFFF}0.00 & \cellcolor[HTML]{FFFFFF}0.00 & \cellcolor[HTML]{FFFFFF}0.00 & \cellcolor[HTML]{FFFFFF}0.00 & \cellcolor[HTML]{FFFFFF}0.00 & \cellcolor[HTML]{FFFFFF}0.00 & \cellcolor[HTML]{FFFFFF}0.00 & \cellcolor[HTML]{FFFFFF}0.00 & \cellcolor[HTML]{FCFDFD}0.02 & \cellcolor[HTML]{FFFFFF}0.00 & \cellcolor[HTML]{FFFFFF}0.00 \\
Palu (3.9$\times$) & 3 & \cellcolor[HTML]{B2C1D1}1.00 & \cellcolor[HTML]{B2C1D1}1.00 & \cellcolor[HTML]{B3C1D1}0.99 & \cellcolor[HTML]{B7C4D3}0.93 & \cellcolor[HTML]{BECAD8}0.82 & \cellcolor[HTML]{E8ECF1}0.25 & \cellcolor[HTML]{C0CCD9}0.80 & \cellcolor[HTML]{C1CCD9}0.78 & \cellcolor[HTML]{C0CCD9}0.79 & \cellcolor[HTML]{E0E6EC}0.34 & \cellcolor[HTML]{D6DDE6}0.48 & \cellcolor[HTML]{C7D1DD}0.69 & \cellcolor[HTML]{E0E6EC}0.34 & \cellcolor[HTML]{C6D0DC}0.71 \\
Palu (5.2$\times$) & 3 & \cellcolor[HTML]{B2C1D1}1.00 & \cellcolor[HTML]{B2C1D1}1.00 & \cellcolor[HTML]{B3C1D1}0.99 & \cellcolor[HTML]{B8C6D4}0.91 & \cellcolor[HTML]{C3CEDB}0.75 & \cellcolor[HTML]{F4F6F8}0.11 & \cellcolor[HTML]{BFCBD8}0.81 & \cellcolor[HTML]{C1CDDA}0.78 & \cellcolor[HTML]{C5D0DC}0.71 & \cellcolor[HTML]{DEE5EB}0.36 & \cellcolor[HTML]{D7DFE7}0.46 & \cellcolor[HTML]{D0D8E2}0.57 & \cellcolor[HTML]{E2E8EE}0.31 & \cellcolor[HTML]{C8D2DE}0.67 \\
Palu (10.2$\times$) & 3 & \cellcolor[HTML]{B3C1D1}0.99 & \cellcolor[HTML]{BBC7D6}0.87 & \cellcolor[HTML]{BCC8D6}0.85 & \cellcolor[HTML]{CFD8E2}0.57 & \cellcolor[HTML]{F9FAFC}0.05 & \cellcolor[HTML]{FFFFFF}0.00 & \cellcolor[HTML]{DDE3EA}0.39 & \cellcolor[HTML]{E1E6ED}0.34 & \cellcolor[HTML]{FFFFFF}0.00 & \cellcolor[HTML]{F9FAFC}0.05 & \cellcolor[HTML]{EEF1F5}0.17 & \cellcolor[HTML]{E2E7ED}0.32 & \cellcolor[HTML]{F2F4F7}0.13 & \cellcolor[HTML]{DEE5EB}0.36 \\
Palu (15.1$\times$) & 3 & \cellcolor[HTML]{D5DDE6}0.49 & \cellcolor[HTML]{EDF0F4}0.19 & \cellcolor[HTML]{FDFDFE}0.01 & \cellcolor[HTML]{EEF1F5}0.17 & \cellcolor[HTML]{FFFFFF}0.00 & \cellcolor[HTML]{FFFFFF}0.00 & \cellcolor[HTML]{F3F6F8}0.11 & \cellcolor[HTML]{F5F7F9}0.09 & \cellcolor[HTML]{FFFFFF}0.00 & \cellcolor[HTML]{FEFEFE}0.01 & \cellcolor[HTML]{FAFBFC}0.04 & \cellcolor[HTML]{F0F3F6}0.15 & \cellcolor[HTML]{F2F4F7}0.13 & \cellcolor[HTML]{F4F6F8}0.11 \\
Palu (19.7$\times$) & 3 & \cellcolor[HTML]{FFFFFF}0.00 & \cellcolor[HTML]{FFFFFF}0.00 & \cellcolor[HTML]{FFFFFF}0.00 & \cellcolor[HTML]{FEFEFE}0.01 & \cellcolor[HTML]{FFFFFF}0.00 & \cellcolor[HTML]{FFFFFF}0.00 & \cellcolor[HTML]{FFFFFF}0.00 & \cellcolor[HTML]{FFFFFF}0.00 & \cellcolor[HTML]{FFFFFF}0.00 & \cellcolor[HTML]{FEFEFE}0.01 & \cellcolor[HTML]{FEFEFF}0.00 & \cellcolor[HTML]{F8F9FB}0.06 & \cellcolor[HTML]{FAFBFC}0.04 & \cellcolor[HTML]{FEFEFE}0.01 \\
Palu (102.4$\times$) & 3 & \cellcolor[HTML]{FFFFFF}0.00 & \cellcolor[HTML]{FFFFFF}0.00 & \cellcolor[HTML]{FFFFFF}0.00 & \cellcolor[HTML]{FFFFFF}0.00 & \cellcolor[HTML]{FFFFFF}0.00 & \cellcolor[HTML]{FFFFFF}0.00 & \cellcolor[HTML]{FFFFFF}0.00 & \cellcolor[HTML]{FFFFFF}0.00 & \cellcolor[HTML]{FFFFFF}0.00 & \cellcolor[HTML]{FFFFFF}0.00 & \cellcolor[HTML]{FFFFFF}0.00 & \cellcolor[HTML]{FFFFFF}0.00 & \cellcolor[HTML]{FFFFFF}0.00 & \cellcolor[HTML]{FFFFFF}0.00 \\
Palu (512.0$\times$) & 3 & \cellcolor[HTML]{FFFFFF}0.00 & \cellcolor[HTML]{FFFFFF}0.00 & \cellcolor[HTML]{FFFFFF}0.00 & \cellcolor[HTML]{FFFFFF}0.00 & \cellcolor[HTML]{FFFFFF}0.00 & \cellcolor[HTML]{FFFFFF}0.00 & \cellcolor[HTML]{FFFFFF}0.00 & \cellcolor[HTML]{FFFFFF}0.00 & \cellcolor[HTML]{FFFFFF}0.00 & \cellcolor[HTML]{FFFFFF}0.00 & \cellcolor[HTML]{FFFFFF}0.00 & \cellcolor[HTML]{FFFFFF}0.00 & \cellcolor[HTML]{FFFFFF}0.00 & \cellcolor[HTML]{FFFFFF}0.00 \\
TransMLA (2.0$\times$) & 3 & \cellcolor[HTML]{B5C3D2}0.96 & \cellcolor[HTML]{BBC8D6}0.86 & \cellcolor[HTML]{C0CCD9}0.79 & \cellcolor[HTML]{E0E6EC}0.34 & \cellcolor[HTML]{FFFFFF}0.00 & \cellcolor[HTML]{FFFFFF}0.00 & \cellcolor[HTML]{DDE3EB}0.38 & \cellcolor[HTML]{E2E7ED}0.32 & \cellcolor[HTML]{E0E6ED}0.34 & \cellcolor[HTML]{F0F2F6}0.15 & \cellcolor[HTML]{E5EAEF}0.28 & \cellcolor[HTML]{E1E7ED}0.33 & \cellcolor[HTML]{EEF1F5}0.17 & \cellcolor[HTML]{DDE4EB}0.38 \\
TransMLA (4.0$\times$) & 3 & \cellcolor[HTML]{B2C1D1}1.00 & \cellcolor[HTML]{C6D1DC}0.71 & \cellcolor[HTML]{D2DAE4}0.53 & \cellcolor[HTML]{E3E8EE}0.31 & \cellcolor[HTML]{FFFFFF}0.00 & \cellcolor[HTML]{FFFFFF}0.00 & \cellcolor[HTML]{DFE5EC}0.35 & \cellcolor[HTML]{E7EBF0}0.26 & \cellcolor[HTML]{E5EAEF}0.29 & \cellcolor[HTML]{F0F3F6}0.14 & \cellcolor[HTML]{EBEFF3}0.20 & \cellcolor[HTML]{E5EAEF}0.28 & \cellcolor[HTML]{F0F3F6}0.15 & \cellcolor[HTML]{E2E7ED}0.32 \\
TransMLA (8.0$\times$) & 3 & \cellcolor[HTML]{B7C4D3}0.93 & \cellcolor[HTML]{DEE4EB}0.37 & \cellcolor[HTML]{E7ECF1}0.25 & \cellcolor[HTML]{F0F3F6}0.15 & \cellcolor[HTML]{FFFFFF}0.00 & \cellcolor[HTML]{FFFFFF}0.00 & \cellcolor[HTML]{EDF0F4}0.18 & \cellcolor[HTML]{F2F5F7}0.12 & \cellcolor[HTML]{FFFFFF}0.00 & \cellcolor[HTML]{F6F8FA}0.08 & \cellcolor[HTML]{FDFEFE}0.01 & \cellcolor[HTML]{F0F3F6}0.15 & \cellcolor[HTML]{F1F3F6}0.14 & \cellcolor[HTML]{EDF0F4}0.18 \\
TransMLA (12.0$\times$) & 3 & \cellcolor[HTML]{B8C5D4}0.91 & \cellcolor[HTML]{F1F3F6}0.14 & \cellcolor[HTML]{FAFBFC}0.04 & \cellcolor[HTML]{F5F7F9}0.09 & \cellcolor[HTML]{FFFFFF}0.00 & \cellcolor[HTML]{FFFFFF}0.00 & \cellcolor[HTML]{F6F7F9}0.09 & \cellcolor[HTML]{FCFCFD}0.02 & \cellcolor[HTML]{FFFFFF}0.00 & \cellcolor[HTML]{FDFDFE}0.02 & \cellcolor[HTML]{F9FAFB}0.05 & \cellcolor[HTML]{F4F6F8}0.11 & \cellcolor[HTML]{F6F8FA}0.08 & \cellcolor[HTML]{F2F5F7}0.12 \\
TransMLA (16.0$\times$) & 3 & \cellcolor[HTML]{CDD6E1}0.61 & \cellcolor[HTML]{F7F8FA}0.07 & \cellcolor[HTML]{FFFFFF}0.00 & \cellcolor[HTML]{F7F9FA}0.07 & \cellcolor[HTML]{FFFFFF}0.00 & \cellcolor[HTML]{FFFFFF}0.00 & \cellcolor[HTML]{F8F9FB}0.06 & \cellcolor[HTML]{FEFEFE}0.01 & \cellcolor[HTML]{FFFFFF}0.00 & \cellcolor[HTML]{FEFEFF}0.00 & \cellcolor[HTML]{FDFDFE}0.02 & \cellcolor[HTML]{F5F7F9}0.09 & \cellcolor[HTML]{FBFCFC}0.03 & \cellcolor[HTML]{F7F8FA}0.07 \\
TransMLA (32.0$\times$) & 3 & \cellcolor[HTML]{FFFFFF}0.00 & \cellcolor[HTML]{FFFFFF}0.00 & \cellcolor[HTML]{FFFFFF}0.00 & \cellcolor[HTML]{FFFFFF}0.00 & \cellcolor[HTML]{FFFFFF}0.00 & \cellcolor[HTML]{FFFFFF}0.00 & \cellcolor[HTML]{FFFFFF}0.00 & \cellcolor[HTML]{FFFFFF}0.00 & \cellcolor[HTML]{FFFFFF}0.00 & \cellcolor[HTML]{FFFFFF}0.00 & \cellcolor[HTML]{FFFFFF}0.00 & \cellcolor[HTML]{FFFFFF}0.00 & \cellcolor[HTML]{FFFFFF}0.00 & \cellcolor[HTML]{FFFFFF}0.00 \\
TransMLA (97.5$\times$) & 3 & \cellcolor[HTML]{FFFFFF}0.00 & \cellcolor[HTML]{FFFFFF}0.00 & \cellcolor[HTML]{FFFFFF}0.00 & \cellcolor[HTML]{FFFFFF}0.00 & \cellcolor[HTML]{FFFFFF}0.00 & \cellcolor[HTML]{FFFFFF}0.00 & \cellcolor[HTML]{FFFFFF}0.00 & \cellcolor[HTML]{FFFFFF}0.00 & \cellcolor[HTML]{FFFFFF}0.00 & \cellcolor[HTML]{FFFFFF}0.00 & \cellcolor[HTML]{FFFFFF}0.00 & \cellcolor[HTML]{FFFFFF}0.00 & \cellcolor[HTML]{FFFFFF}0.00 & \cellcolor[HTML]{FFFFFF}0.00 \\
TransMLA (409.6$\times$) & 3 & \cellcolor[HTML]{FFFFFF}0.00 & \cellcolor[HTML]{FFFFFF}0.00 & \cellcolor[HTML]{FFFFFF}0.00 & \cellcolor[HTML]{FFFFFF}0.00 & \cellcolor[HTML]{FFFFFF}0.00 & \cellcolor[HTML]{FFFFFF}0.00 & \cellcolor[HTML]{FFFFFF}0.00 & \cellcolor[HTML]{FFFFFF}0.00 & \cellcolor[HTML]{FFFFFF}0.00 & \cellcolor[HTML]{FFFFFF}0.00 & \cellcolor[HTML]{FFFFFF}0.00 & \cellcolor[HTML]{FFFFFF}0.00 & \cellcolor[HTML]{FFFFFF}0.00 & \cellcolor[HTML]{FFFFFF}0.00 \\
KIVI (4.9$\times$) & 1 & \cellcolor[HTML]{BFCBD9}0.80 & \cellcolor[HTML]{C4CFDB}0.74 & \cellcolor[HTML]{F2F5F7}0.12 & \cellcolor[HTML]{DFE5EC}0.36 & \cellcolor[HTML]{FAFBFC}0.04 & \cellcolor[HTML]{FFFFFF}0.00 & \cellcolor[HTML]{E8ECF1}0.24 & \cellcolor[HTML]{E0E6EC}0.34 & \cellcolor[HTML]{DDE4EB}0.38 & \cellcolor[HTML]{EAEEF2}0.22 & \cellcolor[HTML]{E2E7ED}0.32 & \cellcolor[HTML]{DCE2EA}0.40 & \cellcolor[HTML]{EDF0F4}0.18 & \cellcolor[HTML]{E2E7ED}0.32 \\
PyramidKV (4.0$\times$) & 1 & \cellcolor[HTML]{B2C1D1}1.00 & \cellcolor[HTML]{C4CFDB}0.74 & \cellcolor[HTML]{FCFDFD}0.02 & \cellcolor[HTML]{CDD7E1}0.60 & \cellcolor[HTML]{F6F8FA}0.08 & \cellcolor[HTML]{FFFFFF}0.00 & \cellcolor[HTML]{E1E7ED}0.33 & \cellcolor[HTML]{E4E9EF}0.29 & \cellcolor[HTML]{C1CCD9}0.78 & \cellcolor[HTML]{E5EAEF}0.28 & \cellcolor[HTML]{F0F3F6}0.15 & \cellcolor[HTML]{C8D2DE}0.68 & \cellcolor[HTML]{E7EBF0}0.26 & \cellcolor[HTML]{DCE2EA}0.40 \\
PyramidKV (12.0$\times$) & 1 & \cellcolor[HTML]{B2C1D1}1.00 & \cellcolor[HTML]{E3E9EE}0.30 & \cellcolor[HTML]{FFFFFF}0.00 & \cellcolor[HTML]{F1F3F6}0.14 & \cellcolor[HTML]{FFFFFF}0.00 & \cellcolor[HTML]{FFFFFF}0.00 & \cellcolor[HTML]{F9FBFC}0.04 & \cellcolor[HTML]{FCFDFD}0.02 & \cellcolor[HTML]{C2CEDA}0.76 & \cellcolor[HTML]{EFF2F6}0.15 & \cellcolor[HTML]{E9EDF2}0.23 & \cellcolor[HTML]{CAD4DF}0.64 & \cellcolor[HTML]{EFF2F5}0.16 & \cellcolor[HTML]{E6EBF0}0.26 \\
PyramidKV (32.0$\times$) & 1 & \cellcolor[HTML]{D9E0E8}0.44 & \cellcolor[HTML]{FAFBFC}0.04 & \cellcolor[HTML]{FFFFFF}0.00 & \cellcolor[HTML]{FAFBFC}0.04 & \cellcolor[HTML]{FFFFFF}0.00 & \cellcolor[HTML]{FFFFFF}0.00 & \cellcolor[HTML]{FFFFFF}0.00 & \cellcolor[HTML]{FFFFFF}0.00 & \cellcolor[HTML]{E8ECF1}0.24 & \cellcolor[HTML]{FCFDFD}0.02 & \cellcolor[HTML]{FDFDFE}0.01 & \cellcolor[HTML]{D4DCE5}0.50 & \cellcolor[HTML]{F1F3F6}0.14 & \cellcolor[HTML]{F3F5F8}0.11 \\
VecInfer (4.3$\times$) & 1 & \cellcolor[HTML]{B2C1D1}1.00 & \cellcolor[HTML]{B9C6D5}0.90 & \cellcolor[HTML]{B5C3D2}0.96 & \cellcolor[HTML]{BFCBD9}0.80 & \cellcolor[HTML]{C4CFDB}0.74 & \cellcolor[HTML]{EBEFF3}0.20 & \cellcolor[HTML]{C3CEDB}0.75 & \cellcolor[HTML]{BECBD8}0.81 & \cellcolor[HTML]{C2CDDA}0.77 & \cellcolor[HTML]{E0E6EC}0.35 & \cellcolor[HTML]{D2DAE4}0.53 & \cellcolor[HTML]{CDD7E1}0.60 & \cellcolor[HTML]{E7EBF0}0.26 & \cellcolor[HTML]{C9D3DE}0.67 \\
VecInfer (12.7$\times$) & 1 & \cellcolor[HTML]{FCFDFD}0.02 & \cellcolor[HTML]{FAFBFC}0.04 & \cellcolor[HTML]{FAFBFC}0.04 & \cellcolor[HTML]{F2F5F7}0.12 & \cellcolor[HTML]{FCFDFD}0.02 & \cellcolor[HTML]{FFFFFF}0.00 & \cellcolor[HTML]{FEFEFE}0.01 & \cellcolor[HTML]{F9FBFC}0.04 & \cellcolor[HTML]{FFFFFF}0.00 & \cellcolor[HTML]{FFFFFF}0.00 & \cellcolor[HTML]{F9FAFB}0.05 & \cellcolor[HTML]{F6F8FA}0.08 & \cellcolor[HTML]{F4F6F8}0.10 & \cellcolor[HTML]{FAFBFC}0.04 \\
VecInfer (35.9$\times$) & 1 & \cellcolor[HTML]{FFFFFF}0.00 & \cellcolor[HTML]{FAFBFC}0.04 & \cellcolor[HTML]{FFFFFF}0.00 & \cellcolor[HTML]{FAFBFC}0.04 & \cellcolor[HTML]{FFFFFF}0.00 & \cellcolor[HTML]{FFFFFF}0.00 & \cellcolor[HTML]{FFFFFF}0.00 & \cellcolor[HTML]{FDFEFE}0.01 & \cellcolor[HTML]{FFFFFF}0.00 & \cellcolor[HTML]{FFFFFF}0.00 & \cellcolor[HTML]{FFFFFF}0.00 & \cellcolor[HTML]{FAFBFC}0.04 & \cellcolor[HTML]{F6F8FA}0.08 & \cellcolor[HTML]{FDFDFE}0.02 \\
\bottomrule
\end{tabular}
\par\vspace{3pt}\parbox{\textwidth}{\small\textbf{Notes:} \textbf{seeds} states how many complete finetuning seeds the score above averages; where \textbf{seeds} ${}>1$, the score is the mean of that many complete seeds' own per-task means. $\pm$ is omitted for compactness. Ratio: uncompressed bytes per token per layer over the cache actually held (\S\ref{sec:budget}), not the nominal budget. FSQ + window: Transformer-FSQ that also keeps the newest 128 tokens and the first 4 tokens as unrounded latents (Appendix~\ref{app:ablations}), whose bytes the ratio includes. Omitted: PyramidKV and VecInfer at 42 and 10\,B, where no configuration of the method fits the budget.}
\end{table}

\clearpage
\begin{table}[H]
\centering
\caption{Per-task RULER on Qwen3-1.7B-Base (13 tasks of 50 items, $8{,}192$ tokens, bfloat16). Finetuned methods use $0.25$ tokens per parameter; KIVI, PyramidKV and VecInfer are training-free. Rows give achieved compression ratios; scores average complete seeds. Shading encodes the task score.}
\label{tab:qwen17b-ruler-breakdown}
\small
\setlength{\tabcolsep}{1.2pt}
\renewcommand{\arraystretch}{0.98}
\begin{tabular}{lccccccccccccccc}
\toprule
\textbf{Method} & \rotatebox{90}{\textbf{seeds}} & \rotatebox{90}{\shortstack[l]{\textbf{niah\_}\\\textbf{single\_1}}} & \rotatebox{90}{\shortstack[l]{\textbf{niah\_}\\\textbf{single\_2}}} & \rotatebox{90}{\shortstack[l]{\textbf{niah\_}\\\textbf{single\_3}}} & \rotatebox{90}{\shortstack[l]{\textbf{niah\_}\\\textbf{multikey\_1}}} & \rotatebox{90}{\shortstack[l]{\textbf{niah\_}\\\textbf{multikey\_2}}} & \rotatebox{90}{\shortstack[l]{\textbf{niah\_}\\\textbf{multikey\_3}}} & \rotatebox{90}{\shortstack[l]{\textbf{niah\_}\\\textbf{multivalue}}} & \rotatebox{90}{\shortstack[l]{\textbf{niah\_}\\\textbf{multiquery}}} & \rotatebox{90}{\textbf{vt}} & \rotatebox{90}{\textbf{cwe}} & \rotatebox{90}{\textbf{fwe}} & \rotatebox{90}{\shortstack[l]{\textbf{qa\_}\\\textbf{1}}} & \rotatebox{90}{\shortstack[l]{\textbf{qa\_}\\\textbf{2}}} & \rotatebox{90}{\textbf{Mean}} \\
\midrule
Uncompressed & 1 & \cellcolor[HTML]{B2C1D1}1.00 & \cellcolor[HTML]{B2C1D1}1.00 & \cellcolor[HTML]{B2C1D1}1.00 & \cellcolor[HTML]{B4C2D1}0.98 & \cellcolor[HTML]{B9C6D5}0.90 & \cellcolor[HTML]{CCD5E0}0.62 & \cellcolor[HTML]{B4C2D2}0.97 & \cellcolor[HTML]{B4C2D1}0.98 & \cellcolor[HTML]{CCD5E0}0.62 & \cellcolor[HTML]{DFE5EC}0.36 & \cellcolor[HTML]{C8D2DE}0.68 & \cellcolor[HTML]{CDD7E1}0.60 & \cellcolor[HTML]{E2E7ED}0.32 & \cellcolor[HTML]{C1CDDA}0.77 \\
\midrule
Transformer-FSQ (2.0$\times$) & 2 & \cellcolor[HTML]{B2C1D1}1.00 & \cellcolor[HTML]{B2C1D1}1.00 & \cellcolor[HTML]{B2C1D1}1.00 & \cellcolor[HTML]{BCC9D7}0.85 & \cellcolor[HTML]{E4E9EF}0.29 & \cellcolor[HTML]{FAFBFC}0.04 & \cellcolor[HTML]{BDC9D7}0.84 & \cellcolor[HTML]{BBC8D6}0.86 & \cellcolor[HTML]{C5D0DC}0.72 & \cellcolor[HTML]{E1E7ED}0.33 & \cellcolor[HTML]{D4DCE5}0.51 & \cellcolor[HTML]{CBD5E0}0.63 & \cellcolor[HTML]{E2E7ED}0.32 & \cellcolor[HTML]{CAD4DF}0.64 \\
Transformer-FSQ (4.0$\times$) & 3 & \cellcolor[HTML]{B2C1D1}1.00 & \cellcolor[HTML]{B3C1D1}0.99 & \cellcolor[HTML]{B3C1D1}0.99 & \cellcolor[HTML]{BBC8D6}0.86 & \cellcolor[HTML]{DAE1E9}0.43 & \cellcolor[HTML]{FAFBFC}0.04 & \cellcolor[HTML]{BCC9D7}0.85 & \cellcolor[HTML]{BBC8D6}0.86 & \cellcolor[HTML]{C0CCD9}0.79 & \cellcolor[HTML]{E1E7ED}0.33 & \cellcolor[HTML]{D7DEE7}0.47 & \cellcolor[HTML]{CCD5E0}0.62 & \cellcolor[HTML]{E1E6ED}0.33 & \cellcolor[HTML]{C9D3DE}0.66 \\
Transformer-FSQ (8.0$\times$) & 3 & \cellcolor[HTML]{B3C1D1}0.99 & \cellcolor[HTML]{BDC9D7}0.84 & \cellcolor[HTML]{B4C2D2}0.97 & \cellcolor[HTML]{CAD4DF}0.64 & \cellcolor[HTML]{F5F7F9}0.09 & \cellcolor[HTML]{FEFEFE}0.01 & \cellcolor[HTML]{C5D0DC}0.72 & \cellcolor[HTML]{CBD5E0}0.63 & \cellcolor[HTML]{D5DDE6}0.49 & \cellcolor[HTML]{DCE3EA}0.40 & \cellcolor[HTML]{E1E7ED}0.33 & \cellcolor[HTML]{D9E0E8}0.43 & \cellcolor[HTML]{EAEEF3}0.21 & \cellcolor[HTML]{D3DBE4}0.52 \\
Transformer-FSQ (12.0$\times$) & 3 & \cellcolor[HTML]{B3C1D1}0.99 & \cellcolor[HTML]{CBD5E0}0.63 & \cellcolor[HTML]{B7C4D3}0.93 & \cellcolor[HTML]{D4DCE5}0.50 & \cellcolor[HTML]{F9FAFB}0.05 & \cellcolor[HTML]{FFFFFF}0.00 & \cellcolor[HTML]{D0D9E2}0.56 & \cellcolor[HTML]{D9E0E8}0.43 & \cellcolor[HTML]{D6DDE6}0.48 & \cellcolor[HTML]{E0E6EC}0.34 & \cellcolor[HTML]{EDF0F4}0.18 & \cellcolor[HTML]{E2E7ED}0.32 & \cellcolor[HTML]{EDF0F4}0.19 & \cellcolor[HTML]{D9E0E8}0.43 \\
Transformer-FSQ (16.0$\times$) & 3 & \cellcolor[HTML]{B2C1D1}1.00 & \cellcolor[HTML]{CCD5E0}0.62 & \cellcolor[HTML]{BBC8D6}0.87 & \cellcolor[HTML]{DDE4EB}0.38 & \cellcolor[HTML]{FAFBFC}0.04 & \cellcolor[HTML]{FFFFFF}0.00 & \cellcolor[HTML]{DAE1E9}0.43 & \cellcolor[HTML]{DFE5EC}0.36 & \cellcolor[HTML]{E7ECF1}0.25 & \cellcolor[HTML]{E6EBF0}0.27 & \cellcolor[HTML]{EEF1F5}0.17 & \cellcolor[HTML]{E6EBF0}0.27 & \cellcolor[HTML]{EDF0F4}0.18 & \cellcolor[HTML]{DEE4EB}0.37 \\
Transformer-FSQ (32.0$\times$) & 3 & \cellcolor[HTML]{B3C1D1}0.99 & \cellcolor[HTML]{D2DAE4}0.54 & \cellcolor[HTML]{CDD7E1}0.60 & \cellcolor[HTML]{E7EBF0}0.26 & \cellcolor[HTML]{FFFFFF}0.00 & \cellcolor[HTML]{FFFFFF}0.00 & \cellcolor[HTML]{EBEFF3}0.20 & \cellcolor[HTML]{EAEEF2}0.22 & \cellcolor[HTML]{FFFFFF}0.00 & \cellcolor[HTML]{F0F3F6}0.15 & \cellcolor[HTML]{F8F9FB}0.06 & \cellcolor[HTML]{EBEFF3}0.20 & \cellcolor[HTML]{F1F3F6}0.14 & \cellcolor[HTML]{E7EBF0}0.26 \\
Transformer-FSQ (97.5$\times$) & 3 & \cellcolor[HTML]{CED7E1}0.59 & \cellcolor[HTML]{F1F3F6}0.14 & \cellcolor[HTML]{FFFFFF}0.00 & \cellcolor[HTML]{F3F5F8}0.11 & \cellcolor[HTML]{FFFFFF}0.00 & \cellcolor[HTML]{FFFFFF}0.00 & \cellcolor[HTML]{F7F8FA}0.07 & \cellcolor[HTML]{F8FAFB}0.05 & \cellcolor[HTML]{FFFFFF}0.00 & \cellcolor[HTML]{FFFFFF}0.00 & \cellcolor[HTML]{FFFFFF}0.00 & \cellcolor[HTML]{F6F8FA}0.08 & \cellcolor[HTML]{FCFDFD}0.02 & \cellcolor[HTML]{F6F8F9}0.08 \\
Transformer-FSQ (409.6$\times$) & 3 & \cellcolor[HTML]{FFFFFF}0.00 & \cellcolor[HTML]{FFFFFF}0.00 & \cellcolor[HTML]{FFFFFF}0.00 & \cellcolor[HTML]{FFFFFF}0.00 & \cellcolor[HTML]{FFFFFF}0.00 & \cellcolor[HTML]{FFFFFF}0.00 & \cellcolor[HTML]{FFFFFF}0.00 & \cellcolor[HTML]{FFFFFF}0.00 & \cellcolor[HTML]{FFFFFF}0.00 & \cellcolor[HTML]{FFFFFF}0.00 & \cellcolor[HTML]{FFFFFF}0.00 & \cellcolor[HTML]{FEFEFE}0.01 & \cellcolor[HTML]{FFFFFF}0.00 & \cellcolor[HTML]{FFFFFF}0.00 \\
FSQ + window (1.8$\times$) & 3 & \cellcolor[HTML]{B2C1D1}1.00 & \cellcolor[HTML]{B3C1D1}0.99 & \cellcolor[HTML]{B2C1D1}1.00 & \cellcolor[HTML]{BDC9D7}0.84 & \cellcolor[HTML]{D5DDE6}0.49 & \cellcolor[HTML]{FDFDFE}0.01 & \cellcolor[HTML]{BCC8D6}0.85 & \cellcolor[HTML]{BAC7D5}0.88 & \cellcolor[HTML]{C8D2DE}0.68 & \cellcolor[HTML]{E2E7ED}0.32 & \cellcolor[HTML]{E2E7ED}0.32 & \cellcolor[HTML]{CCD6E0}0.61 & \cellcolor[HTML]{E2E8EE}0.31 & \cellcolor[HTML]{CAD4DF}0.64 \\
FSQ + window (3.5$\times$) & 3 & \cellcolor[HTML]{B2C1D1}1.00 & \cellcolor[HTML]{B3C1D1}0.99 & \cellcolor[HTML]{B3C1D1}0.99 & \cellcolor[HTML]{BECAD8}0.82 & \cellcolor[HTML]{D6DEE6}0.47 & \cellcolor[HTML]{FDFDFE}0.01 & \cellcolor[HTML]{BDC9D7}0.84 & \cellcolor[HTML]{BECAD8}0.83 & \cellcolor[HTML]{CAD4DF}0.65 & \cellcolor[HTML]{DFE5EC}0.35 & \cellcolor[HTML]{DFE5EC}0.35 & \cellcolor[HTML]{CCD6E0}0.61 & \cellcolor[HTML]{E2E7ED}0.32 & \cellcolor[HTML]{CBD5DF}0.63 \\
FSQ + window (7.1$\times$) & 3 & \cellcolor[HTML]{B2C1D1}1.00 & \cellcolor[HTML]{BDC9D7}0.84 & \cellcolor[HTML]{B2C1D1}1.00 & \cellcolor[HTML]{CBD5E0}0.63 & \cellcolor[HTML]{F8F9FB}0.06 & \cellcolor[HTML]{FFFFFF}0.00 & \cellcolor[HTML]{C6D0DC}0.71 & \cellcolor[HTML]{CCD5E0}0.62 & \cellcolor[HTML]{D6DDE6}0.48 & \cellcolor[HTML]{E0E6EC}0.35 & \cellcolor[HTML]{E9EDF2}0.23 & \cellcolor[HTML]{D8E0E8}0.45 & \cellcolor[HTML]{EAEEF2}0.22 & \cellcolor[HTML]{D4DCE5}0.51 \\
FSQ + window (10.6$\times$) & 3 & \cellcolor[HTML]{B2C1D1}1.00 & \cellcolor[HTML]{C7D1DD}0.69 & \cellcolor[HTML]{B4C2D1}0.98 & \cellcolor[HTML]{D1DAE3}0.55 & \cellcolor[HTML]{F7F9FA}0.07 & \cellcolor[HTML]{FFFFFF}0.00 & \cellcolor[HTML]{D0D8E2}0.57 & \cellcolor[HTML]{D3DCE5}0.52 & \cellcolor[HTML]{D6DEE6}0.48 & \cellcolor[HTML]{E0E6EC}0.34 & \cellcolor[HTML]{EFF2F5}0.16 & \cellcolor[HTML]{DFE5EC}0.35 & \cellcolor[HTML]{EBEFF3}0.20 & \cellcolor[HTML]{D8DFE7}0.45 \\
FSQ + window (14.2$\times$) & 3 & \cellcolor[HTML]{B2C1D1}1.00 & \cellcolor[HTML]{CAD4DF}0.65 & \cellcolor[HTML]{BAC7D5}0.89 & \cellcolor[HTML]{DEE4EB}0.37 & \cellcolor[HTML]{FCFDFD}0.02 & \cellcolor[HTML]{FFFFFF}0.00 & \cellcolor[HTML]{DAE1E9}0.42 & \cellcolor[HTML]{DFE5EC}0.36 & \cellcolor[HTML]{DDE3EA}0.39 & \cellcolor[HTML]{E6EBF0}0.26 & \cellcolor[HTML]{E7ECF1}0.25 & \cellcolor[HTML]{E5EAF0}0.27 & \cellcolor[HTML]{EEF1F5}0.17 & \cellcolor[HTML]{DDE3EA}0.39 \\
FSQ + window (28.4$\times$) & 3 & \cellcolor[HTML]{B3C1D1}0.99 & \cellcolor[HTML]{CAD4DF}0.65 & \cellcolor[HTML]{CFD8E2}0.58 & \cellcolor[HTML]{E5EAEF}0.28 & \cellcolor[HTML]{FFFFFF}0.00 & \cellcolor[HTML]{FFFFFF}0.00 & \cellcolor[HTML]{EAEEF2}0.22 & \cellcolor[HTML]{E8EDF1}0.24 & \cellcolor[HTML]{FFFFFF}0.00 & \cellcolor[HTML]{EFF2F5}0.16 & \cellcolor[HTML]{F9FAFB}0.05 & \cellcolor[HTML]{ECF0F4}0.19 & \cellcolor[HTML]{F1F4F7}0.13 & \cellcolor[HTML]{E6EBF0}0.27 \\
FSQ + window (87.1$\times$) & 3 & \cellcolor[HTML]{C1CDDA}0.77 & \cellcolor[HTML]{EFF2F6}0.15 & \cellcolor[HTML]{FFFFFF}0.00 & \cellcolor[HTML]{F3F5F8}0.11 & \cellcolor[HTML]{FFFFFF}0.00 & \cellcolor[HTML]{FFFFFF}0.00 & \cellcolor[HTML]{F5F7F9}0.09 & \cellcolor[HTML]{F8F9FB}0.06 & \cellcolor[HTML]{FFFFFF}0.00 & \cellcolor[HTML]{FFFFFF}0.00 & \cellcolor[HTML]{FFFFFF}0.00 & \cellcolor[HTML]{F5F7F9}0.09 & \cellcolor[HTML]{FBFCFD}0.03 & \cellcolor[HTML]{F4F6F8}0.10 \\
FSQ + window (372.4$\times$) & 3 & \cellcolor[HTML]{FFFFFF}0.00 & \cellcolor[HTML]{FFFFFF}0.00 & \cellcolor[HTML]{FFFFFF}0.00 & \cellcolor[HTML]{FFFFFF}0.00 & \cellcolor[HTML]{FFFFFF}0.00 & \cellcolor[HTML]{FFFFFF}0.00 & \cellcolor[HTML]{FFFFFF}0.00 & \cellcolor[HTML]{FFFFFF}0.00 & \cellcolor[HTML]{FFFFFF}0.00 & \cellcolor[HTML]{FFFFFF}0.00 & \cellcolor[HTML]{FFFFFF}0.00 & \cellcolor[HTML]{FFFFFF}0.00 & \cellcolor[HTML]{FFFFFF}0.00 & \cellcolor[HTML]{FFFFFF}0.00 \\
Palu (3.9$\times$) & 3 & \cellcolor[HTML]{B2C1D1}1.00 & \cellcolor[HTML]{B2C1D1}1.00 & \cellcolor[HTML]{B2C1D1}1.00 & \cellcolor[HTML]{B6C4D3}0.94 & \cellcolor[HTML]{BAC7D5}0.88 & \cellcolor[HTML]{EBEFF3}0.20 & \cellcolor[HTML]{B5C3D3}0.95 & \cellcolor[HTML]{B5C3D3}0.95 & \cellcolor[HTML]{BDC9D7}0.84 & \cellcolor[HTML]{DEE4EB}0.37 & \cellcolor[HTML]{C8D2DE}0.68 & \cellcolor[HTML]{C8D2DE}0.67 & \cellcolor[HTML]{DFE5EC}0.35 & \cellcolor[HTML]{C2CEDA}0.76 \\
Palu (5.2$\times$) & 3 & \cellcolor[HTML]{B2C1D1}1.00 & \cellcolor[HTML]{B3C1D1}0.99 & \cellcolor[HTML]{B2C1D1}1.00 & \cellcolor[HTML]{B9C6D5}0.90 & \cellcolor[HTML]{BDC9D7}0.84 & \cellcolor[HTML]{FBFCFD}0.03 & \cellcolor[HTML]{B6C4D3}0.94 & \cellcolor[HTML]{B8C5D4}0.92 & \cellcolor[HTML]{C0CCD9}0.79 & \cellcolor[HTML]{DCE3EA}0.40 & \cellcolor[HTML]{C7D2DD}0.69 & \cellcolor[HTML]{CBD5E0}0.63 & \cellcolor[HTML]{E4E9EF}0.29 & \cellcolor[HTML]{C5D0DC}0.72 \\
Palu (10.2$\times$) & 3 & \cellcolor[HTML]{B3C1D1}0.99 & \cellcolor[HTML]{D8DFE7}0.45 & \cellcolor[HTML]{E3E9EE}0.30 & \cellcolor[HTML]{E1E7ED}0.33 & \cellcolor[HTML]{FEFEFE}0.01 & \cellcolor[HTML]{FFFFFF}0.00 & \cellcolor[HTML]{DDE4EB}0.38 & \cellcolor[HTML]{EBEFF3}0.21 & \cellcolor[HTML]{D0D9E2}0.57 & \cellcolor[HTML]{E1E7ED}0.33 & \cellcolor[HTML]{E3E8EE}0.31 & \cellcolor[HTML]{DAE1E9}0.43 & \cellcolor[HTML]{EFF2F5}0.16 & \cellcolor[HTML]{E0E6EC}0.34 \\
Palu (19.7$\times$) & 3 & \cellcolor[HTML]{FEFEFE}0.01 & \cellcolor[HTML]{EDF0F4}0.19 & \cellcolor[HTML]{FFFFFF}0.00 & \cellcolor[HTML]{F8F9FB}0.06 & \cellcolor[HTML]{FFFFFF}0.00 & \cellcolor[HTML]{FFFFFF}0.00 & \cellcolor[HTML]{FBFCFC}0.03 & \cellcolor[HTML]{FCFCFD}0.03 & \cellcolor[HTML]{FFFFFF}0.00 & \cellcolor[HTML]{FFFFFF}0.00 & \cellcolor[HTML]{FCFCFD}0.02 & \cellcolor[HTML]{F3F5F8}0.11 & \cellcolor[HTML]{F4F6F8}0.10 & \cellcolor[HTML]{FAFBFC}0.04 \\
Palu (102.4$\times$) & 3 & \cellcolor[HTML]{FFFFFF}0.00 & \cellcolor[HTML]{FFFFFF}0.00 & \cellcolor[HTML]{FFFFFF}0.00 & \cellcolor[HTML]{FFFFFF}0.00 & \cellcolor[HTML]{FFFFFF}0.00 & \cellcolor[HTML]{FFFFFF}0.00 & \cellcolor[HTML]{FFFFFF}0.00 & \cellcolor[HTML]{FFFFFF}0.00 & \cellcolor[HTML]{FFFFFF}0.00 & \cellcolor[HTML]{FFFFFF}0.00 & \cellcolor[HTML]{FFFFFF}0.00 & \cellcolor[HTML]{FFFFFF}0.00 & \cellcolor[HTML]{FFFFFF}0.00 & \cellcolor[HTML]{FFFFFF}0.00 \\
Palu (512.0$\times$) & 3 & \cellcolor[HTML]{FFFFFF}0.00 & \cellcolor[HTML]{FFFFFF}0.00 & \cellcolor[HTML]{FFFFFF}0.00 & \cellcolor[HTML]{FFFFFF}0.00 & \cellcolor[HTML]{FFFFFF}0.00 & \cellcolor[HTML]{FFFFFF}0.00 & \cellcolor[HTML]{FFFFFF}0.00 & \cellcolor[HTML]{FFFFFF}0.00 & \cellcolor[HTML]{FFFFFF}0.00 & \cellcolor[HTML]{FFFFFF}0.00 & \cellcolor[HTML]{FFFFFF}0.00 & \cellcolor[HTML]{FFFFFF}0.00 & \cellcolor[HTML]{FFFFFF}0.00 & \cellcolor[HTML]{FFFFFF}0.00 \\
TransMLA (2.0$\times$) & 3 & \cellcolor[HTML]{B5C3D2}0.96 & \cellcolor[HTML]{C3CEDB}0.75 & \cellcolor[HTML]{C5D0DC}0.72 & \cellcolor[HTML]{DFE5EC}0.36 & \cellcolor[HTML]{FFFFFF}0.00 & \cellcolor[HTML]{FFFFFF}0.00 & \cellcolor[HTML]{E0E6EC}0.34 & \cellcolor[HTML]{E4E9EF}0.29 & \cellcolor[HTML]{FDFDFE}0.02 & \cellcolor[HTML]{ECF0F4}0.19 & \cellcolor[HTML]{FEFEFE}0.01 & \cellcolor[HTML]{E1E6ED}0.33 & \cellcolor[HTML]{ECF0F4}0.19 & \cellcolor[HTML]{E2E7ED}0.32 \\
TransMLA (4.0$\times$) & 3 & \cellcolor[HTML]{B4C2D2}0.97 & \cellcolor[HTML]{C6D1DC}0.71 & \cellcolor[HTML]{D5DDE6}0.49 & \cellcolor[HTML]{EEF1F5}0.17 & \cellcolor[HTML]{FFFFFF}0.00 & \cellcolor[HTML]{FFFFFF}0.00 & \cellcolor[HTML]{EDF0F4}0.19 & \cellcolor[HTML]{EAEEF2}0.22 & \cellcolor[HTML]{E1E7ED}0.33 & \cellcolor[HTML]{E7ECF1}0.25 & \cellcolor[HTML]{EAEEF2}0.22 & \cellcolor[HTML]{E5EAF0}0.27 & \cellcolor[HTML]{F1F3F6}0.14 & \cellcolor[HTML]{E3E8EE}0.30 \\
TransMLA (8.0$\times$) & 3 & \cellcolor[HTML]{BDCAD7}0.83 & \cellcolor[HTML]{DBE2E9}0.41 & \cellcolor[HTML]{F1F3F6}0.14 & \cellcolor[HTML]{F5F7F9}0.09 & \cellcolor[HTML]{FFFFFF}0.00 & \cellcolor[HTML]{FFFFFF}0.00 & \cellcolor[HTML]{EEF1F5}0.17 & \cellcolor[HTML]{F7F8FA}0.07 & \cellcolor[HTML]{FFFFFF}0.00 & \cellcolor[HTML]{F3F6F8}0.11 & \cellcolor[HTML]{F8F9FA}0.06 & \cellcolor[HTML]{F0F3F6}0.15 & \cellcolor[HTML]{F7F8FA}0.07 & \cellcolor[HTML]{EFF2F5}0.16 \\
TransMLA (16.0$\times$) & 3 & \cellcolor[HTML]{F9FAFB}0.05 & \cellcolor[HTML]{F9FAFC}0.05 & \cellcolor[HTML]{FFFFFF}0.00 & \cellcolor[HTML]{FBFCFC}0.03 & \cellcolor[HTML]{FFFFFF}0.00 & \cellcolor[HTML]{FFFFFF}0.00 & \cellcolor[HTML]{FBFCFD}0.03 & \cellcolor[HTML]{FCFDFD}0.02 & \cellcolor[HTML]{FFFFFF}0.00 & \cellcolor[HTML]{FFFFFF}0.00 & \cellcolor[HTML]{F6F8FA}0.08 & \cellcolor[HTML]{F8F9FB}0.06 & \cellcolor[HTML]{FFFFFF}0.00 & \cellcolor[HTML]{FCFCFD}0.02 \\
TransMLA (32.0$\times$) & 3 & \cellcolor[HTML]{FFFFFF}0.00 & \cellcolor[HTML]{FFFFFF}0.00 & \cellcolor[HTML]{FFFFFF}0.00 & \cellcolor[HTML]{FFFFFF}0.00 & \cellcolor[HTML]{FFFFFF}0.00 & \cellcolor[HTML]{FFFFFF}0.00 & \cellcolor[HTML]{FFFFFF}0.00 & \cellcolor[HTML]{FFFFFF}0.00 & \cellcolor[HTML]{FFFFFF}0.00 & \cellcolor[HTML]{FFFFFF}0.00 & \cellcolor[HTML]{FFFFFF}0.00 & \cellcolor[HTML]{FFFFFF}0.00 & \cellcolor[HTML]{FFFFFF}0.00 & \cellcolor[HTML]{FFFFFF}0.00 \\
TransMLA (97.5$\times$) & 3 & \cellcolor[HTML]{FFFFFF}0.00 & \cellcolor[HTML]{FFFFFF}0.00 & \cellcolor[HTML]{FFFFFF}0.00 & \cellcolor[HTML]{FFFFFF}0.00 & \cellcolor[HTML]{FFFFFF}0.00 & \cellcolor[HTML]{FFFFFF}0.00 & \cellcolor[HTML]{FFFFFF}0.00 & \cellcolor[HTML]{FFFFFF}0.00 & \cellcolor[HTML]{FFFFFF}0.00 & \cellcolor[HTML]{FFFFFF}0.00 & \cellcolor[HTML]{FFFFFF}0.00 & \cellcolor[HTML]{FFFFFF}0.00 & \cellcolor[HTML]{FFFFFF}0.00 & \cellcolor[HTML]{FFFFFF}0.00 \\
TransMLA (409.6$\times$) & 3 & \cellcolor[HTML]{FFFFFF}0.00 & \cellcolor[HTML]{FFFFFF}0.00 & \cellcolor[HTML]{FFFFFF}0.00 & \cellcolor[HTML]{FFFFFF}0.00 & \cellcolor[HTML]{FFFFFF}0.00 & \cellcolor[HTML]{FFFFFF}0.00 & \cellcolor[HTML]{FFFFFF}0.00 & \cellcolor[HTML]{FFFFFF}0.00 & \cellcolor[HTML]{FFFFFF}0.00 & \cellcolor[HTML]{FFFFFF}0.00 & \cellcolor[HTML]{FFFFFF}0.00 & \cellcolor[HTML]{FFFFFF}0.00 & \cellcolor[HTML]{FFFFFF}0.00 & \cellcolor[HTML]{FFFFFF}0.00 \\
KIVI (4.9$\times$) & 1 & \cellcolor[HTML]{B5C3D2}0.96 & \cellcolor[HTML]{B6C4D3}0.94 & \cellcolor[HTML]{C6D1DD}0.70 & \cellcolor[HTML]{D2DAE4}0.54 & \cellcolor[HTML]{F1F3F6}0.14 & \cellcolor[HTML]{FFFFFF}0.00 & \cellcolor[HTML]{CDD7E1}0.60 & \cellcolor[HTML]{D2DAE4}0.54 & \cellcolor[HTML]{D9E0E8}0.44 & \cellcolor[HTML]{D6DEE6}0.48 & \cellcolor[HTML]{CDD6E1}0.61 & \cellcolor[HTML]{D3DBE4}0.52 & \cellcolor[HTML]{E7EBF0}0.26 & \cellcolor[HTML]{D3DBE5}0.52 \\
PyramidKV (4.0$\times$) & 1 & \cellcolor[HTML]{B2C1D1}1.00 & \cellcolor[HTML]{B6C4D3}0.94 & \cellcolor[HTML]{FCFDFD}0.02 & \cellcolor[HTML]{C6D1DD}0.70 & \cellcolor[HTML]{E7EBF0}0.26 & \cellcolor[HTML]{FFFFFF}0.00 & \cellcolor[HTML]{CFD8E2}0.58 & \cellcolor[HTML]{D4DCE5}0.51 & \cellcolor[HTML]{CFD8E2}0.58 & \cellcolor[HTML]{E1E7ED}0.33 & \cellcolor[HTML]{DAE1E9}0.43 & \cellcolor[HTML]{CDD7E1}0.60 & \cellcolor[HTML]{E3E9EE}0.30 & \cellcolor[HTML]{D6DEE6}0.48 \\
PyramidKV (12.0$\times$) & 1 & \cellcolor[HTML]{B2C1D1}1.00 & \cellcolor[HTML]{DAE1E9}0.42 & \cellcolor[HTML]{FFFFFF}0.00 & \cellcolor[HTML]{E8ECF1}0.24 & \cellcolor[HTML]{FFFFFF}0.00 & \cellcolor[HTML]{FFFFFF}0.00 & \cellcolor[HTML]{F6F7F9}0.09 & \cellcolor[HTML]{FDFDFE}0.01 & \cellcolor[HTML]{D3DBE5}0.52 & \cellcolor[HTML]{E9EDF2}0.23 & \cellcolor[HTML]{DBE2E9}0.41 & \cellcolor[HTML]{D0D9E3}0.56 & \cellcolor[HTML]{E5EAEF}0.28 & \cellcolor[HTML]{E4E9EF}0.29 \\
PyramidKV (32.0$\times$) & 1 & \cellcolor[HTML]{D3DBE4}0.52 & \cellcolor[HTML]{F2F5F7}0.12 & \cellcolor[HTML]{FFFFFF}0.00 & \cellcolor[HTML]{FCFDFD}0.02 & \cellcolor[HTML]{FFFFFF}0.00 & \cellcolor[HTML]{FFFFFF}0.00 & \cellcolor[HTML]{FFFFFF}0.00 & \cellcolor[HTML]{FFFFFF}0.00 & \cellcolor[HTML]{DDE4EB}0.38 & \cellcolor[HTML]{FBFCFD}0.03 & \cellcolor[HTML]{F9FAFC}0.05 & \cellcolor[HTML]{D9E0E8}0.44 & \cellcolor[HTML]{EBEFF3}0.20 & \cellcolor[HTML]{F1F4F7}0.14 \\
VecInfer (4.3$\times$) & 1 & \cellcolor[HTML]{B2C1D1}1.00 & \cellcolor[HTML]{B4C2D1}0.98 & \cellcolor[HTML]{B2C1D1}1.00 & \cellcolor[HTML]{B2C1D1}1.00 & \cellcolor[HTML]{C5D0DC}0.72 & \cellcolor[HTML]{D0D9E3}0.56 & \cellcolor[HTML]{B6C4D3}0.94 & \cellcolor[HTML]{B5C3D3}0.95 & \cellcolor[HTML]{D2DAE4}0.54 & \cellcolor[HTML]{DBE2E9}0.41 & \cellcolor[HTML]{C7D2DD}0.69 & \cellcolor[HTML]{CDD7E1}0.60 & \cellcolor[HTML]{E2E7ED}0.32 & \cellcolor[HTML]{C3CEDB}0.75 \\
VecInfer (12.7$\times$) & 1 & \cellcolor[HTML]{FCFDFD}0.02 & \cellcolor[HTML]{FAFBFC}0.04 & \cellcolor[HTML]{FAFBFC}0.04 & \cellcolor[HTML]{F2F5F7}0.12 & \cellcolor[HTML]{FCFDFD}0.02 & \cellcolor[HTML]{FFFFFF}0.00 & \cellcolor[HTML]{FEFEFE}0.01 & \cellcolor[HTML]{F9FBFC}0.04 & \cellcolor[HTML]{FFFFFF}0.00 & \cellcolor[HTML]{FEFEFE}0.01 & \cellcolor[HTML]{E2E8EE}0.31 & \cellcolor[HTML]{EDF0F4}0.18 & \cellcolor[HTML]{F6F8FA}0.08 & \cellcolor[HTML]{F7F9FA}0.07 \\
VecInfer (35.9$\times$) & 1 & \cellcolor[HTML]{FCFDFD}0.02 & \cellcolor[HTML]{FCFDFD}0.02 & \cellcolor[HTML]{FFFFFF}0.00 & \cellcolor[HTML]{FAFBFC}0.04 & \cellcolor[HTML]{FFFFFF}0.00 & \cellcolor[HTML]{FFFFFF}0.00 & \cellcolor[HTML]{FFFFFF}0.00 & \cellcolor[HTML]{FCFCFD}0.03 & \cellcolor[HTML]{FFFFFF}0.00 & \cellcolor[HTML]{FDFDFE}0.02 & \cellcolor[HTML]{FDFDFE}0.01 & \cellcolor[HTML]{FCFDFD}0.02 & \cellcolor[HTML]{FAFBFC}0.04 & \cellcolor[HTML]{FDFDFE}0.01 \\
\bottomrule
\end{tabular}
\par\vspace{3pt}\parbox{\textwidth}{\small\textbf{Notes:} \textbf{seeds} states how many complete finetuning seeds the score above averages; where \textbf{seeds} ${}>1$, the score is the mean of that many complete seeds' own per-task means. $\pm$ is omitted for compactness. Ratio: uncompressed bytes per token per layer over the cache actually held (\S\ref{sec:budget}), not the nominal budget. FSQ + window: Transformer-FSQ that also keeps the newest 128 tokens and the first 4 tokens as unrounded latents (Appendix~\ref{app:ablations}), whose bytes the ratio includes. Omitted: PyramidKV and VecInfer at 42 and 10\,B, where no configuration of the method fits the budget.}
\end{table}

\clearpage
\begin{table}[H]
\centering
\caption{Per-task RULER on Qwen3-8B-Base (13 tasks of 50 items, $8{,}192$ tokens, bfloat16). Finetuned methods use $0.25$ tokens per parameter; KIVI, PyramidKV and VecInfer are training-free. Rows give achieved compression ratios; scores average complete seeds. Shading encodes the task score.}
\label{tab:qwen8b-ruler-breakdown}
\small
\setlength{\tabcolsep}{1.2pt}
\renewcommand{\arraystretch}{0.98}
\begin{tabular}{lccccccccccccccc}
\toprule
\textbf{Method} & \rotatebox{90}{\textbf{seeds}} & \rotatebox{90}{\shortstack[l]{\textbf{niah\_}\\\textbf{single\_1}}} & \rotatebox{90}{\shortstack[l]{\textbf{niah\_}\\\textbf{single\_2}}} & \rotatebox{90}{\shortstack[l]{\textbf{niah\_}\\\textbf{single\_3}}} & \rotatebox{90}{\shortstack[l]{\textbf{niah\_}\\\textbf{multikey\_1}}} & \rotatebox{90}{\shortstack[l]{\textbf{niah\_}\\\textbf{multikey\_2}}} & \rotatebox{90}{\shortstack[l]{\textbf{niah\_}\\\textbf{multikey\_3}}} & \rotatebox{90}{\shortstack[l]{\textbf{niah\_}\\\textbf{multivalue}}} & \rotatebox{90}{\shortstack[l]{\textbf{niah\_}\\\textbf{multiquery}}} & \rotatebox{90}{\textbf{vt}} & \rotatebox{90}{\textbf{cwe}} & \rotatebox{90}{\textbf{fwe}} & \rotatebox{90}{\shortstack[l]{\textbf{qa\_}\\\textbf{1}}} & \rotatebox{90}{\shortstack[l]{\textbf{qa\_}\\\textbf{2}}} & \rotatebox{90}{\textbf{Mean}} \\
\midrule
Uncompressed & 1 & \cellcolor[HTML]{B2C1D1}1.00 & \cellcolor[HTML]{B4C2D1}0.98 & \cellcolor[HTML]{B2C1D1}1.00 & \cellcolor[HTML]{B2C1D1}1.00 & \cellcolor[HTML]{B5C3D2}0.96 & \cellcolor[HTML]{B2C1D1}1.00 & \cellcolor[HTML]{B3C1D1}0.99 & \cellcolor[HTML]{B3C1D1}0.99 & \cellcolor[HTML]{B9C6D5}0.90 & \cellcolor[HTML]{B7C5D4}0.92 & \cellcolor[HTML]{B9C6D5}0.90 & \cellcolor[HTML]{B9C6D5}0.90 & \cellcolor[HTML]{D9E0E8}0.44 & \cellcolor[HTML]{B7C5D4}0.92 \\
\midrule
Transformer-FSQ (8.0$\times$) & 1 & \cellcolor[HTML]{B2C1D1}1.00 & \cellcolor[HTML]{B5C3D2}0.96 & \cellcolor[HTML]{B5C3D2}0.96 & \cellcolor[HTML]{BECAD8}0.82 & \cellcolor[HTML]{E8ECF1}0.24 & \cellcolor[HTML]{FAFBFC}0.04 & \cellcolor[HTML]{BCC9D7}0.84 & \cellcolor[HTML]{C1CDDA}0.78 & \cellcolor[HTML]{CED7E1}0.60 & \cellcolor[HTML]{C6D1DD}0.70 & \cellcolor[HTML]{BCC9D7}0.85 & \cellcolor[HTML]{CCD5E0}0.62 & \cellcolor[HTML]{EAEEF2}0.22 & \cellcolor[HTML]{C9D3DE}0.66 \\
Transformer-FSQ (12.0$\times$) & 1 & \cellcolor[HTML]{B2C1D1}1.00 & \cellcolor[HTML]{B2C1D1}1.00 & \cellcolor[HTML]{B2C1D1}1.00 & \cellcolor[HTML]{C9D3DE}0.66 & \cellcolor[HTML]{F1F3F6}0.14 & \cellcolor[HTML]{FCFDFD}0.02 & \cellcolor[HTML]{C0CCD9}0.79 & \cellcolor[HTML]{C2CDDA}0.77 & \cellcolor[HTML]{CFD8E2}0.58 & \cellcolor[HTML]{CFD8E2}0.58 & \cellcolor[HTML]{C4CFDB}0.73 & \cellcolor[HTML]{D3DBE4}0.52 & \cellcolor[HTML]{EBEFF3}0.20 & \cellcolor[HTML]{CCD6E0}0.61 \\
Transformer-FSQ (32.0$\times$) & 1 & \cellcolor[HTML]{B2C1D1}1.00 & \cellcolor[HTML]{B2C1D1}1.00 & \cellcolor[HTML]{B9C6D5}0.90 & \cellcolor[HTML]{CCD5E0}0.62 & \cellcolor[HTML]{FAFBFC}0.04 & \cellcolor[HTML]{FFFFFF}0.00 & \cellcolor[HTML]{D0D9E3}0.56 & \cellcolor[HTML]{D3DBE4}0.52 & \cellcolor[HTML]{DBE2EA}0.40 & \cellcolor[HTML]{D8DFE7}0.46 & \cellcolor[HTML]{DDE4EB}0.38 & \cellcolor[HTML]{DFE5EC}0.36 & \cellcolor[HTML]{EDF0F4}0.18 & \cellcolor[HTML]{D5DDE6}0.49 \\
Transformer-FSQ (97.5$\times$) & 1 & \cellcolor[HTML]{B5C3D2}0.96 & \cellcolor[HTML]{C4CFDB}0.74 & \cellcolor[HTML]{E3E9EE}0.30 & \cellcolor[HTML]{EBEFF3}0.20 & \cellcolor[HTML]{FFFFFF}0.00 & \cellcolor[HTML]{FFFFFF}0.00 & \cellcolor[HTML]{EEF1F4}0.17 & \cellcolor[HTML]{F1F4F7}0.14 & \cellcolor[HTML]{FFFFFF}0.00 & \cellcolor[HTML]{F7F8FA}0.07 & \cellcolor[HTML]{EFF2F6}0.15 & \cellcolor[HTML]{F1F3F6}0.14 & \cellcolor[HTML]{F1F3F6}0.14 & \cellcolor[HTML]{E9EDF2}0.23 \\
Transformer-FSQ (409.6$\times$) & 1 & \cellcolor[HTML]{FFFFFF}0.00 & \cellcolor[HTML]{FFFFFF}0.00 & \cellcolor[HTML]{FFFFFF}0.00 & \cellcolor[HTML]{FFFFFF}0.00 & \cellcolor[HTML]{FFFFFF}0.00 & \cellcolor[HTML]{FFFFFF}0.00 & \cellcolor[HTML]{FFFFFF}0.00 & \cellcolor[HTML]{FFFFFF}0.00 & \cellcolor[HTML]{FFFFFF}0.00 & \cellcolor[HTML]{FFFFFF}0.00 & \cellcolor[HTML]{FFFFFF}0.00 & \cellcolor[HTML]{F8F9FB}0.06 & \cellcolor[HTML]{FFFFFF}0.00 & \cellcolor[HTML]{FEFEFF}0.00 \\
FSQ + window (28.4$\times$) & 1 & \cellcolor[HTML]{B2C1D1}1.00 & \cellcolor[HTML]{B2C1D1}1.00 & \cellcolor[HTML]{B9C6D5}0.90 & \cellcolor[HTML]{CDD7E1}0.60 & \cellcolor[HTML]{FAFBFC}0.04 & \cellcolor[HTML]{FFFFFF}0.00 & \cellcolor[HTML]{CED7E1}0.59 & \cellcolor[HTML]{D0D9E2}0.56 & \cellcolor[HTML]{CFD8E2}0.58 & \cellcolor[HTML]{D3DBE4}0.52 & \cellcolor[HTML]{D7DEE7}0.47 & \cellcolor[HTML]{DCE2EA}0.40 & \cellcolor[HTML]{EAEEF2}0.22 & \cellcolor[HTML]{D2DBE4}0.53 \\
FSQ + window (87.1$\times$) & 1 & \cellcolor[HTML]{B5C3D2}0.96 & \cellcolor[HTML]{C1CCD9}0.78 & \cellcolor[HTML]{E8ECF1}0.24 & \cellcolor[HTML]{E5EAEF}0.28 & \cellcolor[HTML]{FFFFFF}0.00 & \cellcolor[HTML]{FFFFFF}0.00 & \cellcolor[HTML]{EBEFF3}0.20 & \cellcolor[HTML]{F0F3F6}0.15 & \cellcolor[HTML]{FFFFFF}0.00 & \cellcolor[HTML]{F7F8FA}0.07 & \cellcolor[HTML]{F2F4F7}0.13 & \cellcolor[HTML]{E8ECF1}0.24 & \cellcolor[HTML]{F4F6F8}0.10 & \cellcolor[HTML]{E8ECF1}0.24 \\
FSQ + window (372.4$\times$) & 1 & \cellcolor[HTML]{FFFFFF}0.00 & \cellcolor[HTML]{FFFFFF}0.00 & \cellcolor[HTML]{FFFFFF}0.00 & \cellcolor[HTML]{FFFFFF}0.00 & \cellcolor[HTML]{FFFFFF}0.00 & \cellcolor[HTML]{FFFFFF}0.00 & \cellcolor[HTML]{FFFFFF}0.00 & \cellcolor[HTML]{FFFFFF}0.00 & \cellcolor[HTML]{FFFFFF}0.00 & \cellcolor[HTML]{FFFFFF}0.00 & \cellcolor[HTML]{FFFFFF}0.00 & \cellcolor[HTML]{F8F9FB}0.06 & \cellcolor[HTML]{FFFFFF}0.00 & \cellcolor[HTML]{FEFEFF}0.00 \\
Palu (10.2$\times$) & 1 & \cellcolor[HTML]{B4C2D1}0.98 & \cellcolor[HTML]{CCD5E0}0.62 & \cellcolor[HTML]{F1F3F6}0.14 & \cellcolor[HTML]{E3E9EE}0.30 & \cellcolor[HTML]{FFFFFF}0.00 & \cellcolor[HTML]{FFFFFF}0.00 & \cellcolor[HTML]{EFF2F5}0.16 & \cellcolor[HTML]{EEF1F4}0.17 & \cellcolor[HTML]{E4E9EF}0.29 & \cellcolor[HTML]{DCE2EA}0.40 & \cellcolor[HTML]{D9E0E8}0.43 & \cellcolor[HTML]{E8ECF1}0.24 & \cellcolor[HTML]{F4F6F8}0.10 & \cellcolor[HTML]{E4E9EF}0.30 \\
Palu (15.1$\times$) & 1 & \cellcolor[HTML]{FFFFFF}0.00 & \cellcolor[HTML]{FFFFFF}0.00 & \cellcolor[HTML]{FFFFFF}0.00 & \cellcolor[HTML]{FFFFFF}0.00 & \cellcolor[HTML]{FFFFFF}0.00 & \cellcolor[HTML]{FFFFFF}0.00 & \cellcolor[HTML]{FFFFFF}0.00 & \cellcolor[HTML]{FFFFFF}0.00 & \cellcolor[HTML]{FFFFFF}0.00 & \cellcolor[HTML]{FFFFFF}0.00 & \cellcolor[HTML]{FFFFFF}0.00 & \cellcolor[HTML]{F8F9FB}0.06 & \cellcolor[HTML]{FFFFFF}0.00 & \cellcolor[HTML]{FEFEFF}0.00 \\
Palu (36.6$\times$) & 1 & \cellcolor[HTML]{FFFFFF}0.00 & \cellcolor[HTML]{FFFFFF}0.00 & \cellcolor[HTML]{FFFFFF}0.00 & \cellcolor[HTML]{FFFFFF}0.00 & \cellcolor[HTML]{FFFFFF}0.00 & \cellcolor[HTML]{FFFFFF}0.00 & \cellcolor[HTML]{FFFFFF}0.00 & \cellcolor[HTML]{FFFFFF}0.00 & \cellcolor[HTML]{FFFFFF}0.00 & \cellcolor[HTML]{FFFFFF}0.00 & \cellcolor[HTML]{FFFFFF}0.00 & \cellcolor[HTML]{F6F8FA}0.08 & \cellcolor[HTML]{FFFFFF}0.00 & \cellcolor[HTML]{FEFEFE}0.01 \\
Palu (102.4$\times$) & 1 & \cellcolor[HTML]{FFFFFF}0.00 & \cellcolor[HTML]{FFFFFF}0.00 & \cellcolor[HTML]{FFFFFF}0.00 & \cellcolor[HTML]{FFFFFF}0.00 & \cellcolor[HTML]{FFFFFF}0.00 & \cellcolor[HTML]{FFFFFF}0.00 & \cellcolor[HTML]{FFFFFF}0.00 & \cellcolor[HTML]{FFFFFF}0.00 & \cellcolor[HTML]{FFFFFF}0.00 & \cellcolor[HTML]{FFFFFF}0.00 & \cellcolor[HTML]{FFFFFF}0.00 & \cellcolor[HTML]{FAFBFC}0.04 & \cellcolor[HTML]{FFFFFF}0.00 & \cellcolor[HTML]{FEFFFF}0.00 \\
Palu (512.0$\times$) & 1 & \cellcolor[HTML]{FFFFFF}0.00 & \cellcolor[HTML]{FFFFFF}0.00 & \cellcolor[HTML]{FFFFFF}0.00 & \cellcolor[HTML]{FFFFFF}0.00 & \cellcolor[HTML]{FFFFFF}0.00 & \cellcolor[HTML]{FFFFFF}0.00 & \cellcolor[HTML]{FFFFFF}0.00 & \cellcolor[HTML]{FFFFFF}0.00 & \cellcolor[HTML]{FFFFFF}0.00 & \cellcolor[HTML]{FFFFFF}0.00 & \cellcolor[HTML]{FFFFFF}0.00 & \cellcolor[HTML]{FCFDFD}0.02 & \cellcolor[HTML]{FFFFFF}0.00 & \cellcolor[HTML]{FFFFFF}0.00 \\
TransMLA (8.0$\times$) & 1 & \cellcolor[HTML]{F8F9FB}0.06 & \cellcolor[HTML]{FFFFFF}0.00 & \cellcolor[HTML]{FFFFFF}0.00 & \cellcolor[HTML]{FFFFFF}0.00 & \cellcolor[HTML]{FFFFFF}0.00 & \cellcolor[HTML]{FFFFFF}0.00 & \cellcolor[HTML]{FFFFFF}0.00 & \cellcolor[HTML]{FFFFFF}0.00 & \cellcolor[HTML]{FFFFFF}0.00 & \cellcolor[HTML]{F4F6F8}0.11 & \cellcolor[HTML]{F4F6F8}0.11 & \cellcolor[HTML]{FAFBFC}0.04 & \cellcolor[HTML]{FFFFFF}0.00 & \cellcolor[HTML]{FCFCFD}0.02 \\
TransMLA (12.0$\times$) & 1 & \cellcolor[HTML]{FFFFFF}0.00 & \cellcolor[HTML]{FFFFFF}0.00 & \cellcolor[HTML]{FFFFFF}0.00 & \cellcolor[HTML]{FFFFFF}0.00 & \cellcolor[HTML]{FFFFFF}0.00 & \cellcolor[HTML]{FFFFFF}0.00 & \cellcolor[HTML]{FFFFFF}0.00 & \cellcolor[HTML]{FFFFFF}0.00 & \cellcolor[HTML]{FFFFFF}0.00 & \cellcolor[HTML]{FFFFFF}0.00 & \cellcolor[HTML]{FBFCFD}0.03 & \cellcolor[HTML]{FAFBFC}0.04 & \cellcolor[HTML]{FFFFFF}0.00 & \cellcolor[HTML]{FEFEFE}0.01 \\
TransMLA (32.0$\times$) & 1 & \cellcolor[HTML]{FFFFFF}0.00 & \cellcolor[HTML]{FFFFFF}0.00 & \cellcolor[HTML]{FFFFFF}0.00 & \cellcolor[HTML]{FFFFFF}0.00 & \cellcolor[HTML]{FFFFFF}0.00 & \cellcolor[HTML]{FFFFFF}0.00 & \cellcolor[HTML]{FFFFFF}0.00 & \cellcolor[HTML]{FFFFFF}0.00 & \cellcolor[HTML]{FFFFFF}0.00 & \cellcolor[HTML]{FFFFFF}0.00 & \cellcolor[HTML]{FFFFFF}0.00 & \cellcolor[HTML]{FFFFFF}0.00 & \cellcolor[HTML]{FFFFFF}0.00 & \cellcolor[HTML]{FFFFFF}0.00 \\
TransMLA (97.5$\times$) & 1 & \cellcolor[HTML]{FFFFFF}0.00 & \cellcolor[HTML]{FFFFFF}0.00 & \cellcolor[HTML]{FFFFFF}0.00 & \cellcolor[HTML]{FFFFFF}0.00 & \cellcolor[HTML]{FFFFFF}0.00 & \cellcolor[HTML]{FFFFFF}0.00 & \cellcolor[HTML]{FFFFFF}0.00 & \cellcolor[HTML]{FFFFFF}0.00 & \cellcolor[HTML]{FFFFFF}0.00 & \cellcolor[HTML]{FFFFFF}0.00 & \cellcolor[HTML]{FFFFFF}0.00 & \cellcolor[HTML]{FCFDFD}0.02 & \cellcolor[HTML]{FFFFFF}0.00 & \cellcolor[HTML]{FFFFFF}0.00 \\
TransMLA (409.6$\times$) & 1 & \cellcolor[HTML]{FFFFFF}0.00 & \cellcolor[HTML]{FFFFFF}0.00 & \cellcolor[HTML]{FFFFFF}0.00 & \cellcolor[HTML]{FFFFFF}0.00 & \cellcolor[HTML]{FFFFFF}0.00 & \cellcolor[HTML]{FFFFFF}0.00 & \cellcolor[HTML]{FFFFFF}0.00 & \cellcolor[HTML]{FFFFFF}0.00 & \cellcolor[HTML]{FFFFFF}0.00 & \cellcolor[HTML]{FFFFFF}0.00 & \cellcolor[HTML]{FFFFFF}0.00 & \cellcolor[HTML]{FFFFFF}0.00 & \cellcolor[HTML]{FFFFFF}0.00 & \cellcolor[HTML]{FFFFFF}0.00 \\
KIVI (4.9$\times$) & 1 & \cellcolor[HTML]{B2C1D1}1.00 & \cellcolor[HTML]{B4C2D1}0.98 & \cellcolor[HTML]{B4C2D1}0.98 & \cellcolor[HTML]{B6C4D3}0.94 & \cellcolor[HTML]{D3DBE4}0.52 & \cellcolor[HTML]{C5D0DC}0.72 & \cellcolor[HTML]{B3C1D1}0.99 & \cellcolor[HTML]{B7C5D4}0.93 & \cellcolor[HTML]{BFCBD8}0.81 & \cellcolor[HTML]{B8C5D4}0.92 & \cellcolor[HTML]{BBC8D6}0.87 & \cellcolor[HTML]{C4CFDB}0.74 & \cellcolor[HTML]{E3E9EE}0.30 & \cellcolor[HTML]{BECAD8}0.82 \\
PyramidKV (4.0$\times$) & 1 & \cellcolor[HTML]{B2C1D1}1.00 & \cellcolor[HTML]{B2C1D1}1.00 & \cellcolor[HTML]{DDE4EB}0.38 & \cellcolor[HTML]{B4C2D1}0.98 & \cellcolor[HTML]{BDC9D7}0.84 & \cellcolor[HTML]{FFFFFF}0.00 & \cellcolor[HTML]{B3C1D1}0.99 & \cellcolor[HTML]{B4C2D2}0.97 & \cellcolor[HTML]{BDCAD7}0.83 & \cellcolor[HTML]{D5DCE5}0.50 & \cellcolor[HTML]{C4CFDB}0.74 & \cellcolor[HTML]{BAC7D5}0.88 & \cellcolor[HTML]{D9E0E8}0.44 & \cellcolor[HTML]{C4CFDB}0.74 \\
PyramidKV (12.0$\times$) & 1 & \cellcolor[HTML]{B2C1D1}1.00 & \cellcolor[HTML]{B5C3D2}0.96 & \cellcolor[HTML]{FFFFFF}0.00 & \cellcolor[HTML]{B2C1D1}1.00 & \cellcolor[HTML]{DAE1E9}0.42 & \cellcolor[HTML]{FFFFFF}0.00 & \cellcolor[HTML]{D0D9E3}0.56 & \cellcolor[HTML]{CDD7E1}0.60 & \cellcolor[HTML]{BBC8D6}0.87 & \cellcolor[HTML]{EEF1F5}0.17 & \cellcolor[HTML]{CDD6E1}0.61 & \cellcolor[HTML]{BDC9D7}0.84 & \cellcolor[HTML]{DAE1E9}0.42 & \cellcolor[HTML]{CFD8E2}0.57 \\
PyramidKV (32.0$\times$) & 1 & \cellcolor[HTML]{B5C3D2}0.96 & \cellcolor[HTML]{DAE1E9}0.42 & \cellcolor[HTML]{FFFFFF}0.00 & \cellcolor[HTML]{E3E9EE}0.30 & \cellcolor[HTML]{F6F8FA}0.08 & \cellcolor[HTML]{FFFFFF}0.00 & \cellcolor[HTML]{FCFDFD}0.02 & \cellcolor[HTML]{FDFEFE}0.01 & \cellcolor[HTML]{EBEFF3}0.21 & \cellcolor[HTML]{FEFEFE}0.01 & \cellcolor[HTML]{F9FAFB}0.05 & \cellcolor[HTML]{C6D1DD}0.70 & \cellcolor[HTML]{DCE2EA}0.40 & \cellcolor[HTML]{E8ECF1}0.24 \\
VecInfer (4.3$\times$) & 1 & \cellcolor[HTML]{B2C1D1}1.00 & \cellcolor[HTML]{B2C1D1}1.00 & \cellcolor[HTML]{B2C1D1}1.00 & \cellcolor[HTML]{B5C3D2}0.96 & \cellcolor[HTML]{BAC7D5}0.88 & \cellcolor[HTML]{B4C2D1}0.98 & \cellcolor[HTML]{B4C2D2}0.97 & \cellcolor[HTML]{B3C1D1}0.98 & \cellcolor[HTML]{BDCAD7}0.83 & \cellcolor[HTML]{B7C5D4}0.92 & \cellcolor[HTML]{B8C6D4}0.91 & \cellcolor[HTML]{BFCBD9}0.80 & \cellcolor[HTML]{E2E7ED}0.32 & \cellcolor[HTML]{BAC7D5}0.89 \\
VecInfer (12.7$\times$) & 1 & \cellcolor[HTML]{FCFDFD}0.02 & \cellcolor[HTML]{FAFBFC}0.04 & \cellcolor[HTML]{FAFBFC}0.04 & \cellcolor[HTML]{F2F5F7}0.12 & \cellcolor[HTML]{FCFDFD}0.02 & \cellcolor[HTML]{FFFFFF}0.00 & \cellcolor[HTML]{FEFEFE}0.01 & \cellcolor[HTML]{FAFBFC}0.04 & \cellcolor[HTML]{FEFEFE}0.01 & \cellcolor[HTML]{FAFBFC}0.04 & \cellcolor[HTML]{DAE1E9}0.42 & \cellcolor[HTML]{E7EBF0}0.26 & \cellcolor[HTML]{F2F5F7}0.12 & \cellcolor[HTML]{F5F7F9}0.09 \\
VecInfer (35.9$\times$) & 1 & \cellcolor[HTML]{FCFDFD}0.02 & \cellcolor[HTML]{FAFBFC}0.04 & \cellcolor[HTML]{FAFBFC}0.04 & \cellcolor[HTML]{F2F5F7}0.12 & \cellcolor[HTML]{FCFDFD}0.02 & \cellcolor[HTML]{FFFFFF}0.00 & \cellcolor[HTML]{FEFEFE}0.01 & \cellcolor[HTML]{F9FBFC}0.04 & \cellcolor[HTML]{FEFEFE}0.01 & \cellcolor[HTML]{F4F6F8}0.10 & \cellcolor[HTML]{D1DAE3}0.55 & \cellcolor[HTML]{EAEEF2}0.22 & \cellcolor[HTML]{F6F8FA}0.08 & \cellcolor[HTML]{F5F7F9}0.10 \\
\bottomrule
\end{tabular}
\par\vspace{3pt}\parbox{\textwidth}{\small\textbf{Notes:} \textbf{seeds} states how many complete finetuning seeds the score above averages; $\pm$ is omitted for compactness. Ratio: uncompressed bytes per token per layer over the cache actually held (\S\ref{sec:budget}), not the nominal budget. FSQ + window: Transformer-FSQ that also keeps the newest 128 tokens and the first 4 tokens as unrounded latents (Appendix~\ref{app:ablations}), whose bytes the ratio includes. Omitted: PyramidKV and VecInfer at 42 and 10\,B, where no configuration of the method fits the budget.}
\end{table}

\subsection{Per-task long-context results: LongBench}

Tables~\ref{tab:qwen-longbench-breakdown} and~\ref{tab:qwen17b-longbench-breakdown} give LongBench on Qwen3-0.6B-Base and Qwen3-1.7B-Base, each task in its own metric (\texttt{narrativeqa} F1, \texttt{gov\_report} ROUGE-L, \texttt{trec} accuracy, all from $0$ to $100$) and never averaged across tasks; each task is averaged only over the documents that every configuration compared in its table scored.

\begin{table}[H]
\centering
\caption{Per-task LongBench on Qwen3-0.6B-Base, each task in its own metric (over-length prompts truncated), over the 20 documents per task that every configuration scored. Finetuned methods use $0.25$ tokens per parameter; KIVI, PyramidKV and VecInfer are training-free. Transformer-FSQ and uncompressed rows run in bfloat16, all other rows in float32.}
\label{tab:qwen-longbench-breakdown}
\vspace{1pt}
\small
\setlength{\tabcolsep}{4pt}
\renewcommand{\arraystretch}{0.88}
\begin{tabular}{lrrcccccc}
\toprule
\textbf{Method} & \textbf{B/tok} & \textbf{Ratio} & \textbf{seeds} & \textbf{$n$} & \textbf{skip} & \shortstack{\textbf{gov\_}\\\textbf{report}} & \shortstack{\textbf{narrative-}\\\textbf{qa}} & \shortstack{\textbf{trec}} \\
\midrule
Uncompressed & -- & -- & 1 & 60 & 0 & 18.2 & 13.0 & 10.0 \\
\midrule
\multicolumn{9}{l}{\textit{Compression 1.4$\times$--2.8$\times$}} \\
Transformer-FSQ & 2048 & 2.0 & 3 & 180 & 0 & 17.1${\pm}$1.3 & 5.3${\pm}$1.2 & 25.0${\pm}$5.0 \\
TransMLA & 2048 & 2.0 & 3 & 180 & 0 & 12.7${\pm}$0.0 & 2.7${\pm}$0.6 & 16.7${\pm}$5.8 \\
VecInfer & 1601$^{\S}$ & 2.6 & 1 & 60 & 0 & 18.1 & 8.0 & 10.0 \\
\midrule
\multicolumn{9}{l}{\textit{Compression 2.8$\times$--5.7$\times$}} \\
Transformer-FSQ & 1024 & 4.0 & 3 & 180 & 0 & 17.4${\pm}$0.9 & 6.0${\pm}$1.1 & 21.7${\pm}$5.8 \\
Palu & 1040 & 3.9 & 3 & 180 & 0 & 17.5${\pm}$0.4 & 10.7${\pm}$2.5 & 16.7${\pm}$2.9 \\
Palu & 784 & 5.2 & 3 & 180 & 0 & 16.0${\pm}$1.5 & 10.5${\pm}$0.5 & 20.0${\pm}$5.0 \\
TransMLA & 1024 & 4.0 & 3 & 180 & 0 & 12.9${\pm}$0.1 & 2.7${\pm}$0.2 & 8.3${\pm}$5.8 \\
KIVI & 832$^{\S}$ & 4.9 & 1 & 60 & 0 & 15.1 & 7.8 & 35.0 \\
\midrule
\multicolumn{9}{l}{\textit{Compression 5.7$\times$--11.3$\times$}} \\
Transformer-FSQ & 512 & 8.0 & 3 & 180 & 0 & 16.7${\pm}$1.1 & 5.7${\pm}$1.1 & 23.3${\pm}$7.6 \\
FSQ-small (2048\,B) & 682 & 6.0 & 1 & 60 & 0 & 14.4 & 6.6 & 30.0 \\
FSQ-small (1024\,B) & 682 & 6.0 & 3 & 180 & 0 & 14.9${\pm}$0.5 & 6.7${\pm}$0.4 & 30.0${\pm}$5.0 \\
FSQ-small & 512 & 8.0 & 3 & 180 & 0 & 14.0${\pm}$1.1 & 4.1${\pm}$0.6 & 35.0${\pm}$8.7 \\
Palu & 400 & 10.2 & 3 & 180 & 0 & 14.5${\pm}$0.3 & 6.1${\pm}$0.5 & 5.0${\pm}$5.0 \\
TransMLA & 512 & 8.0 & 3 & 180 & 0 & 9.3${\pm}$1.0 & 1.6${\pm}$0.4 & 10.0${\pm}$8.7 \\
PyramidKV & 509$^{\S}$ & 8.0 & 1 & 60 & 0 & 12.7 & 10.4 & 10.0 \\
VecInfer & 705$^{\S}$ & 5.8 & 1 & 60 & 0 & 13.1 & 2.9 & 15.0 \\
\midrule
\multicolumn{9}{l}{\textit{Compression 11.3$\times$--22.6$\times$}} \\
Transformer-FSQ & 256 & 16.0 & 3 & 180 & 0 & 14.2${\pm}$0.4 & 3.9${\pm}$1.0 & 31.7${\pm}$7.6 \\
FSQ-small & 256 & 16.0 & 3 & 180 & 0 & 13.7${\pm}$0.5 & 4.5${\pm}$0.5 & 31.7${\pm}$5.8 \\
Palu & 208 & 19.7 & 3 & 180 & 0 & 12.1${\pm}$0.4 & 2.2${\pm}$0.5 & 15.0${\pm}$5.0 \\
TransMLA & 256 & 16.0 & 3 & 180 & 0 & 8.5${\pm}$0.7 & 1.5${\pm}$0.6 & 6.7${\pm}$7.6 \\
PyramidKV & 238$^{\S}$ & 17.2 & 1 & 60 & 0 & 11.3 & 9.4 & 15.0 \\
PyramidKV@8k & 226$^{\S}$ & 18.1 & 1 & 60 & 0 & 11.2 & 14.3 & 15.0 \\
VecInfer & 322$^{\S}$ & 12.7 & 1 & 60 & 0 & 6.1 & 2.9 & 5.0 \\
VecInfer@8k & 225$^{\S}$ & 18.2 & 1 & 60 & 0 & 6.1 & 0.7 & 0.0 \\
\midrule
\multicolumn{9}{l}{\textit{Compression 22.6$\times$--45.3$\times$}} \\
Transformer-FSQ & 128 & 32.0 & 3 & 180 & 0 & 11.7${\pm}$0.7 & 5.3${\pm}$1.5 & 16.7${\pm}$7.6 \\
FSQ-small & 128 & 32.0 & 3 & 180 & 0 & 11.3${\pm}$1.1 & 4.5${\pm}$1.3 & 18.3${\pm}$5.8 \\
TransMLA & 128 & 32.0 & 3 & 180 & 0 & 3.6${\pm}$0.3 & 1.1${\pm}$1.1 & 0.0 \\
PyramidKV & 103$^{\S}$ & 39.6 & 1 & 60 & 0 & 8.2 & 8.4 & 20.0 \\
PyramidKV@8k & 96$^{\S}$ & 42.5 & 1 & 60 & 0 & 8.4 & 8.8 & 20.0 \\
VecInfer@8k & 113$^{\S}$ & 36.1 & 1 & 60 & 0 & 3.9 & 0.4 & 15.0 \\
\midrule
\multicolumn{9}{l}{\textit{Compression 90.5$\times$--181.0$\times$}} \\
Transformer-FSQ & 42 & 97.5 & 3 & 180 & 0 & 6.8${\pm}$0.6 & 3.9${\pm}$0.5 & 3.3${\pm}$5.8 \\
FSQ-small & 42 & 97.5 & 3 & 180 & 0 & 7.9${\pm}$1.0 & 4.0${\pm}$1.0 & 5.0${\pm}$5.0 \\
Palu & 40 & 102.4 & 3 & 180 & 0 & 0.3${\pm}$0.5 & 1.6${\pm}$0.7 & 0.0 \\
TransMLA & 42 & 97.5 & 3 & 180 & 0 & 2.3${\pm}$1.2 & 0.7${\pm}$0.6 & 0.0 \\
\midrule
\multicolumn{9}{l}{\textit{Compression 362.0$\times$--724.1$\times$}} \\
Transformer-FSQ & 10 & 409.6 & 3 & 180 & 0 & 1.4${\pm}$0.2 & 2.2${\pm}$1.4 & 0.0 \\
FSQ-small & 10 & 409.6 & 3 & 180 & 0 & 1.4${\pm}$0.5 & 2.3${\pm}$0.4 & 0.0 \\
Palu & 8 & 512.0 & 3 & 180 & 0 & 7.3${\pm}$1.5 & 1.5${\pm}$0.6 & 0.0 \\
TransMLA & 10 & 409.6 & 3 & 180 & 0 & 0.9${\pm}$0.8 & 2.5${\pm}$2.0 & 0.0 \\
\bottomrule
\end{tabular}
\par\vspace{1pt}\parbox{\textwidth}{\small\textbf{Notes:} $\pm$: one sample standard deviation over \textbf{seeds}. \textbf{B/tok}: cache bytes per token per layer; \textbf{ratio}: uncompressed size over \textbf{B/tok}; \textbf{$n$}/\textbf{skip}: examples the row's own runs scored/did not score (prefill out of memory), over seeds and tasks; task columns use only the caption's documents. Task columns: raw scores (0--100), averaged over documents, then seeds; never averaged across tasks. \texttt{@} suffix (e.g.\ PyramidKV@8k): cache sized at that context, not 2048 tokens; its own row. FSQ-small: Transformer-FSQ with its code capped at two thirds of the key-value width, where the bottleneck stops adding parameters over the K and V projections it replaces, with 8-bit levels where they fit more bytes; a parenthesised size is the budget the run asked for, which the cap holds to \textbf{B/tok}. $^{\S}$: \textbf{B/tok}, \textbf{ratio} depend on tokens held (KIVI, PyramidKV, VecInfer); averaged over the row's scored examples.}
\end{table}

\begin{table}[H]
\centering
\caption{Per-task LongBench on Qwen3-1.7B-Base, each task in its own metric (over-length prompts truncated), over the 20 documents per task that every configuration scored. Finetuned methods use $0.25$ tokens per parameter; KIVI, PyramidKV and VecInfer are training-free. Transformer-FSQ and uncompressed rows run in bfloat16, all other rows in float32.}
\label{tab:qwen17b-longbench-breakdown}
\vspace{1pt}
\small
\setlength{\tabcolsep}{4pt}
\renewcommand{\arraystretch}{0.88}
\begin{tabular}{lrrcccccc}
\toprule
\textbf{Method} & \textbf{B/tok} & \textbf{Ratio} & \textbf{seeds} & \textbf{$n$} & \textbf{skip} & \shortstack{\textbf{gov\_}\\\textbf{report}} & \shortstack{\textbf{narrative-}\\\textbf{qa}} & \shortstack{\textbf{trec}} \\
\midrule
Uncompressed & -- & -- & 1 & 60 & 0 & 17.9 & 12.0 & 40.0 \\
\midrule
\multicolumn{9}{l}{\textit{Compression 1.4$\times$--2.8$\times$}} \\
Transformer-FSQ & 2048 & 2.0 & 3 & 180 & 0 & 15.9${\pm}$0.6 & 8.9${\pm}$0.9 & 30.0${\pm}$13.2 \\
TransMLA & 2048 & 2.0 & 3 & 180 & 0 & 14.6${\pm}$0.1 & 4.1${\pm}$1.5 & 20.0${\pm}$5.0 \\
VecInfer & 1601$^{\S}$ & 2.6 & 3 & 180 & 0 & 17.1${\pm}$0.5 & 10.1${\pm}$1.6 & 38.3${\pm}$2.9 \\
\midrule
\multicolumn{9}{l}{\textit{Compression 2.8$\times$--5.7$\times$}} \\
Transformer-FSQ & 1024 & 4.0 & 3 & 180 & 0 & 16.4${\pm}$0.3 & 9.0${\pm}$0.6 & 30.0${\pm}$8.7 \\
Palu & 1040 & 3.9 & 3 & 180 & 0 & 17.6${\pm}$0.3 & 11.5${\pm}$1.1 & 45.0${\pm}$5.0 \\
Palu & 784 & 5.2 & 3 & 180 & 0 & 16.4${\pm}$0.7 & 11.4${\pm}$1.2 & 36.7${\pm}$2.9 \\
TransMLA & 1024 & 4.0 & 3 & 180 & 0 & 13.2${\pm}$0.4 & 3.8${\pm}$0.4 & 15.0${\pm}$8.7 \\
KIVI & 832$^{\S}$ & 4.9 & 3 & 180 & 0 & 16.5 & 14.1 & 50.0 \\
\midrule
\multicolumn{9}{l}{\textit{Compression 5.7$\times$--11.3$\times$}} \\
Transformer-FSQ & 512 & 8.0 & 3 & 180 & 0 & 14.6${\pm}$0.7 & 5.2${\pm}$0.9 & 28.3${\pm}$5.8 \\
Palu & 400 & 10.2 & 3 & 180 & 0 & 13.1${\pm}$0.2 & 7.0${\pm}$1.4 & 56.7${\pm}$7.6 \\
TransMLA & 512 & 8.0 & 3 & 180 & 0 & 8.9${\pm}$0.5 & 2.3${\pm}$1.4 & 6.7${\pm}$5.8 \\
PyramidKV & 504$^{\S}$ & 8.1 & 3 & 180 & 0 & 11.7 & 12.1 & 35.0 \\
VecInfer & 705$^{\S}$ & 5.8 & 3 & 180 & 0 & 13.9${\pm}$0.6 & 5.0${\pm}$2.3 & 35.0${\pm}$5.0 \\
\midrule
\multicolumn{9}{l}{\textit{Compression 11.3$\times$--22.6$\times$}} \\
Transformer-FSQ & 341 & 12.0 & 3 & 180 & 0 & 13.4${\pm}$1.3 & 4.3${\pm}$1.5 & 25.0${\pm}$5.0 \\
Transformer-FSQ & 256 & 16.0 & 3 & 180 & 0 & 13.1${\pm}$0.8 & 6.8${\pm}$0.5 & 31.7${\pm}$2.9 \\
Palu & 208 & 19.7 & 3 & 180 & 0 & 10.6${\pm}$0.7 & 3.8${\pm}$1.1 & 11.7${\pm}$7.6 \\
TransMLA & 256 & 16.0 & 3 & 180 & 0 & 5.8${\pm}$0.8 & 1.7${\pm}$0.9 & 5.0${\pm}$5.0 \\
PyramidKV & 229$^{\S}$ & 17.9 & 3 & 180 & 0 & 9.9 & 14.6 & 30.0 \\
PyramidKV@8k & 216$^{\S}$ & 18.9 & 1 & 60 & 0 & 9.1 & 12.8 & 30.0 \\
VecInfer & 322$^{\S}$ & 12.7 & 3 & 180 & 0 & 6.0${\pm}$0.4 & 3.8${\pm}$1.5 & 25.0${\pm}$5.0 \\
VecInfer@8k & 226$^{\S}$ & 18.2 & 1 & 60 & 0 & 6.7 & 2.4 & 30.0 \\
\midrule
\multicolumn{9}{l}{\textit{Compression 22.6$\times$--45.3$\times$}} \\
Transformer-FSQ & 128 & 32.0 & 3 & 180 & 0 & 11.7${\pm}$1.2 & 6.1${\pm}$1.4 & 35.0${\pm}$5.0 \\
TransMLA & 128 & 32.0 & 3 & 180 & 0 & 4.4${\pm}$1.1 & 1.5${\pm}$0.3 & 0.0 \\
PyramidKV & 93$^{\S}$ & 44.0 & 3 & 180 & 0 & 8.2 & 12.3 & 25.0 \\
VecInfer & 129$^{\S}$ & 31.6 & 3 & 180 & 0 & 6.0${\pm}$0.9 & 3.1${\pm}$1.1 & 18.3${\pm}$5.8 \\
VecInfer@8k & 113$^{\S}$ & 36.1 & 1 & 60 & 0 & 6.8 & 1.3 & 25.0 \\
\midrule
\multicolumn{9}{l}{\textit{Compression 45.3$\times$--90.5$\times$}} \\
PyramidKV & 47$^{\S}$ & 87.4 & 3 & 180 & 0 & 7.1 & 14.8 & 20.0 \\
PyramidKV@8k & 86$^{\S}$ & 47.8 & 1 & 60 & 0 & 8.5 & 18.5 & 25.0 \\
\midrule
\multicolumn{9}{l}{\textit{Compression 90.5$\times$--181.0$\times$}} \\
Transformer-FSQ & 42 & 97.5 & 3 & 180 & 0 & 9.6${\pm}$0.4 & 2.0${\pm}$0.3 & 20.0 \\
Palu & 40 & 102.4 & 3 & 180 & 0 & 0.1${\pm}$0.1 & 0.5${\pm}$0.4 & 0.0 \\
TransMLA & 42 & 97.5 & 3 & 180 & 0 & 3.3${\pm}$0.5 & 0.3${\pm}$0.6 & 0.0 \\
\midrule
\multicolumn{9}{l}{\textit{Compression 362.0$\times$--724.1$\times$}} \\
Transformer-FSQ & 10 & 409.6 & 3 & 180 & 0 & 1.7${\pm}$1.8 & 3.5${\pm}$1.3 & 0.0 \\
Palu & 8 & 512.0 & 3 & 180 & 0 & 1.9${\pm}$1.5 & 0.6${\pm}$0.5 & 0.0 \\
TransMLA & 10 & 409.6 & 3 & 180 & 0 & 0.3${\pm}$0.4 & 0.8${\pm}$0.8 & 0.0 \\
\bottomrule
\end{tabular}
\par\vspace{1pt}\parbox{\textwidth}{\small\textbf{Notes:} $\pm$: one sample standard deviation over \textbf{seeds}. \textbf{B/tok}: cache bytes per token per layer; \textbf{ratio}: uncompressed size over \textbf{B/tok}; \textbf{$n$}/\textbf{skip}: examples the row's own runs scored/did not score (prefill out of memory), over seeds and tasks; task columns use only the caption's documents. Task columns: raw scores (0--100), averaged over documents, then seeds; never averaged across tasks. \texttt{@} suffix (e.g.\ PyramidKV@8k): cache sized at that context, not 2048 tokens; its own row. $^{\S}$: \textbf{B/tok}, \textbf{ratio} depend on tokens held (KIVI, PyramidKV, VecInfer); averaged over the row's scored examples.}
\end{table}

Table~\ref{tab:qwen8b-longbench-breakdown} gives the same breakdown for Qwen3-8B-Base, from the same $200$-document-per-task campaign as Figure~\ref{fig:longbench}.

\begin{table}[H]
\centering
\caption{Per-task LongBench on Qwen3-8B-Base, each task in its own metric (over-length prompts truncated), over the 180 gov\_report, 70 narrativeqa and 200 trec documents that every configuration scored. Finetuned methods use $0.25$ tokens per parameter; KIVI, PyramidKV and VecInfer are training-free. Transformer-FSQ and uncompressed rows run in bfloat16, all other rows in float32.}
\label{tab:qwen8b-longbench-breakdown}
\vspace{1pt}
\small
\setlength{\tabcolsep}{4pt}
\renewcommand{\arraystretch}{0.88}
\begin{tabular}{lrrcccccc}
\toprule
\textbf{Method} & \textbf{B/tok} & \textbf{Ratio} & \textbf{seeds} & \textbf{$n$} & \textbf{skip} & \shortstack{\textbf{gov\_}\\\textbf{report}} & \shortstack{\textbf{narrative-}\\\textbf{qa}} & \shortstack{\textbf{trec}} \\
\midrule
Uncompressed & -- & -- & 1 & 462 & 138 & 16.9 & 17.4 & 63.0 \\
\midrule
\multicolumn{9}{l}{\textit{Compression 2.8$\times$--5.7$\times$}} \\
KIVI & 847$^{\S}$ & 4.8 & 1 & 450 & 150 & 15.4 & 23.5 & 58.0 \\
\midrule
\multicolumn{9}{l}{\textit{Compression 5.7$\times$--11.3$\times$}} \\
Transformer-FSQ & 512 & 8.0 & 1 & 600 & 0 & 15.0 & 10.8 & 43.5 \\
VecInfer & 721$^{\S}$ & 5.7 & 1 & 450 & 150 & 15.6 & 16.3 & 59.0 \\
\midrule
\multicolumn{9}{l}{\textit{Compression 11.3$\times$--22.6$\times$}} \\
Transformer-FSQ & 341 & 12.0 & 1 & 600 & 0 & 14.5 & 10.1 & 46.0 \\
PyramidKV & 287$^{\S}$ & 14.3 & 1 & 478 & 122 & 10.5 & 18.8 & 60.5 \\
\midrule
\multicolumn{9}{l}{\textit{Compression 22.6$\times$--45.3$\times$}} \\
Transformer-FSQ & 128 & 32.0 & 1 & 600 & 0 & 12.5 & 8.5 & 38.0 \\
Palu & 112 & 36.6 & 1 & 462 & 138 & 8.0 & 3.0 & 2.0 \\
VecInfer & 145$^{\S}$ & 28.2 & 1 & 462 & 138 & 9.2 & 7.4 & 20.0 \\
\midrule
\multicolumn{9}{l}{\textit{Compression 45.3$\times$--90.5$\times$}} \\
PyramidKV & 56$^{\S}$ & 73.1 & 1 & 480 & 120 & 5.4 & 11.5 & 27.5 \\
\midrule
\multicolumn{9}{l}{\textit{Compression 90.5$\times$--181.0$\times$}} \\
Transformer-FSQ & 42 & 97.5 & 1 & 600 & 0 & 11.4 & 6.5 & 3.5 \\
Palu & 40 & 102.4 & 1 & 462 & 138 & 7.0 & 2.2 & 2.0 \\
\midrule
\multicolumn{9}{l}{\textit{Compression 362.0$\times$--724.1$\times$}} \\
Transformer-FSQ & 10 & 409.6 & 1 & 600 & 0 & 8.8 & 4.7 & 3.5 \\
Palu & 8 & 512.0 & 1 & 462 & 138 & 3.6 & 2.2 & 0.0 \\
\bottomrule
\end{tabular}
\par\vspace{1pt}\parbox{\textwidth}{\small\textbf{Notes:} $\pm$: one sample standard deviation over \textbf{seeds}. \textbf{B/tok}: cache bytes per token per layer; \textbf{ratio}: uncompressed size over \textbf{B/tok}; \textbf{$n$}/\textbf{skip}: examples the row's own runs scored/did not score (prefill out of memory), over seeds and tasks; task columns use only the caption's documents. Task columns: raw scores (0--100), averaged over documents, then seeds; never averaged across tasks. $^{\S}$: \textbf{B/tok}, \textbf{ratio} depend on tokens held (KIVI, PyramidKV, VecInfer); averaged over the row's scored examples.}
\end{table}

\end{document}